\documentclass[10pt,twocolumn,letterpaper]{article}

\usepackage{iccv}
\usepackage{times}
\usepackage{epsfig}
\usepackage{graphicx}
\usepackage{amsmath, amsfonts, amssymb}
\usepackage{algorithm, algpseudocode} %
\usepackage{enumitem}
\usepackage{wrapfig}
\usepackage{pifont} %
\usepackage{stmaryrd} %
\usepackage{booktabs, multirow} %
\usepackage{tablefootnote}
\usepackage{comment}
\usepackage{colortbl} %
\usepackage{xcolor}
\usepackage{sidecap} %
\usepackage{dblfloatfix} %

\def\1{\mathbf{1}}

\newcommand{\cmark}{\ding{52}}

\newcommand{\xmark}{\ding{55}}

\definecolor{apricot}{rgb}{0.98, 0.81, 0.69}

\usepackage[pagebackref=true,breaklinks=true,colorlinks,bookmarks=false]{hyperref}

\iccvfinalcopy %

\ificcvfinal\fi

\begin{document}

\title{The Stable Signature: Rooting Watermarks in Latent Diffusion Models}
\author{
    Pierre Fernandez$^{1,2}$ \quad
    Guillaume Couairon$^{1,3}$ \quad
    Herv\'e J\'egou$^{1}$  \quad
    Matthijs Douze$^{1}$ \quad
    Teddy Furon$^{2}$\thanks{Work supported by ANR / AID under Chaire SAIDA ANR-20-CHIA-0011. Correspondance to pfz@meta.com} 
    \\[0.5cm]
    \normalsize{
        $^1$Meta AI \qquad $^2$Centre Inria de l'Université de Rennes \qquad $^3$Sorbonne University
    }
}
\maketitle
\ificcvfinal\thispagestyle{empty}\fi

\vspace*{-0.8cm}
\begin{abstract}
Generative image modeling enables a wide range of applications but raises ethical concerns about responsible deployment.   
This paper introduces an active strategy combining image watermarking and Latent Diffusion Models. 
The goal is for all generated images to conceal an invisible watermark allowing for future detection and/or identification.
The method quickly fine-tunes the latent decoder of the image generator, conditioned on a binary signature.
A pre-trained watermark extractor recovers the hidden signature from any generated image and a statistical test then determines whether it comes from the generative model. 
We evaluate the invisibility and robustness of the watermarks on a variety of generation tasks, showing that Stable Signature works even after the images are modified.
For instance, it detects the origin of an image generated from a text prompt, then cropped to keep 10\% of the content, with 90+\% accuracy at a false positive rate below 10$^{-6}$.
\end{abstract}

\vspace*{-0.4cm}
\section{Introduction}

Recent progress in generative modeling and natural language processing enable easy creation and manipulation of photo-realistic images, such as with DALL·E~2~\cite{ramesh2022dalle2} or Stable Diffusion~\cite{rombach2022ldm}.
They have given birth to many image edition tools like ControlNet~\cite{zhang2023adding}, Instruct-Pix2Pix~\cite{brooks2022instructpix2pix}, and others~\cite{couairon2022diffedit, gal2022image, ruiz2022dreambooth}, that are establishing themselves as creative tools for artists, designers, and the general public.

While this is a great step forward for generative AI, it also renews concerns about undermining confidence in the authenticity or veracity of photo-realistic images. 
Indeed, methods to convincingly augment photo-realistic images have existed for a while, but generative AI significantly lowers the barriers to convincing synthetic image generation and edition (\eg a generated picture recently won an art competition~\cite{gault2022vice}).
Not being able to identify that images are generated by AI makes it difficult to remove them from certain platforms and to ensure their compliance with ethical standards. 
It opens the doors to new risks like deep fakes, impersonation or copyright usurpation~\cite{brundage2018malicious, denton2021ethical}.
    
A baseline solution to identify generated images is forensics, \ie~passive methods to detect generated/manipulated images.
On the other hand, existing watermarking methods can be added on top of image generation.
They are based on the idea of invisibly embedding a secret message into the image, which can then be extracted and used to identify the image.
This has several drawbacks. 
If the model leaks or is open-sourced, the post-generation watermarking is easy to remove.
The open source Stable Diffusion~\cite{2022stablediffusion} is a case in point, since removing the watermark amounts to commenting out a single line in the source code.

\begin{figure}
    \centering 
    \includegraphics[width=0.92\linewidth, trim={0cm 0cm 0cm 0cm}, clip]{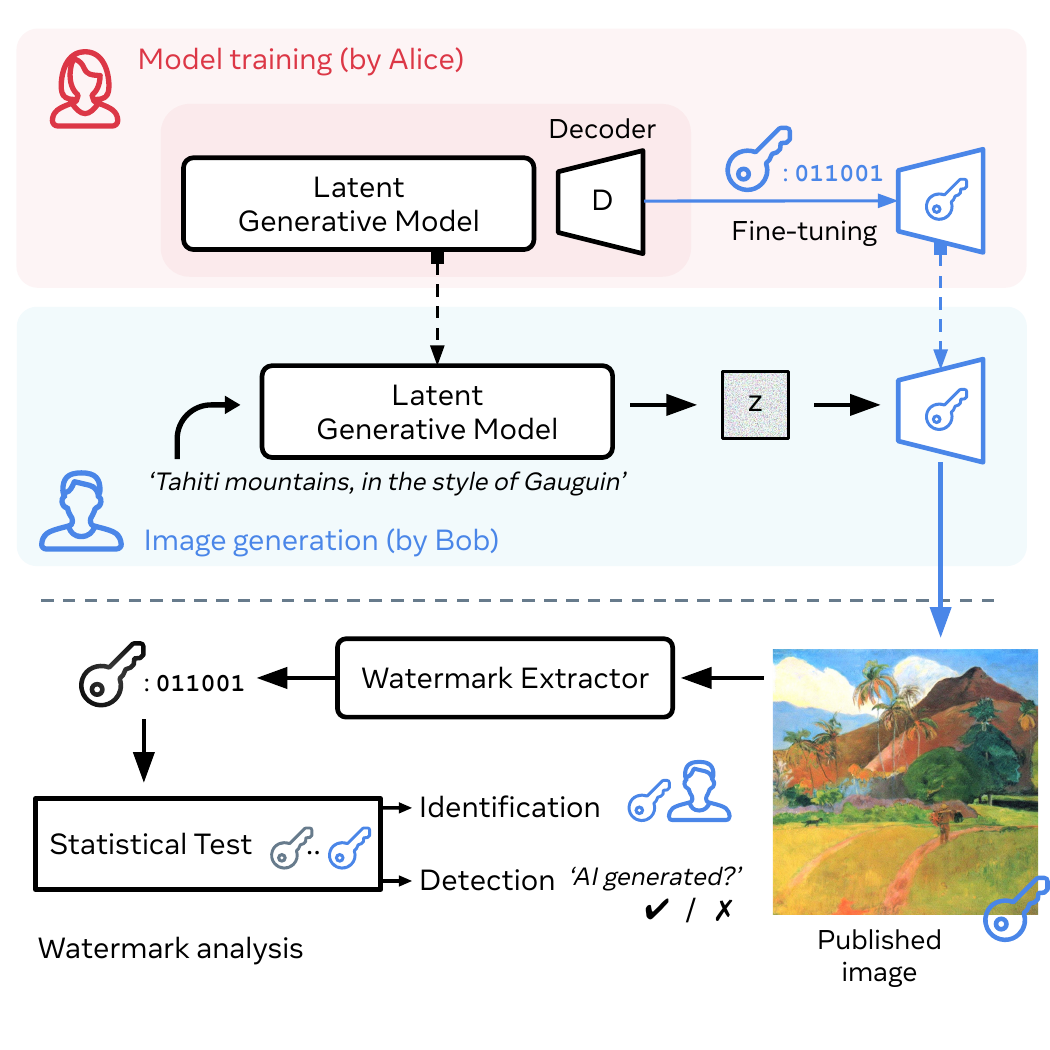}
    \vspace{-0.5cm}
    \caption{
        Overview. 
        The latent decoder can be fine-tuned to preemptively embed a signature into all generated images.
    }
    \label{fig:fig1} \vspace*{-0.5cm}
\end{figure}

Our Stable Signature method merges watermarking into the generation process itself, without any architectural changes.
It adjusts the pre-trained generative model such that all the images it produces conceal a given watermark.
There are several advantages to this approach~\cite{lin2022cycleganwm, yu2022responsible}.
It protects both the generator and its productions. 
Besides, it does not require additional processing of the generated image, which makes the watermarking computationally lighter, straightforward, and secure.
Model providers would then be able to deploy their models to different user groups with a unique watermark, and monitor that they are used in a responsible manner.
They could also give art platforms, news outlets and other sharing platforms %
the ability to detect when an image has been generated by their AI.

We focus on Latent Diffusion Models (LDM)~\cite{rombach2022ldm} since they can perform a wide range of generative tasks.
This work shows that simply fine-tuning a small part of the generative model -- the decoder that generates images from the latent vectors -- is enough to natively embed a watermark into all generated images.
Stable Signature does not require any architectural change and does not modify the diffusion process. Hence it is  compatible with most of the LDM-based generative methods~\cite{brooks2022instructpix2pix, couairon2022diffedit, peebles2022dit, ruiz2022dreambooth, zhang2023adding}.
The fine-tuning stage is performed by back-propagating a combination of a perceptual image loss and the hidden message loss from a watermark extractor back to the LDM decoder.
We pre-train the extractor with a simplified version of the deep watermarking method HiDDeN~\cite{zhu2018hidden}.

We create an evaluation benchmark close to real world situations where images may be edited.
The tasks are: detection of AI generated images, tracing models from their generations.
For instance, we detect $90\%$ of images generated with the generative model, even if they are cropped to $10\%$ of their original size, while flagging only one false positive every $10^6$ images. 
To ensure that the model's utility is not weakened, we show that the FID~\cite{heusel2017gans} score of the generation is not affected and that the generated images are perceptually indistinguishable from the ones produced by the original model. 
This is done over several tasks involving LDM (text-to-image, inpainting, edition, etc.).

As a summary, 
(1) we efficiently merge watermarking into the generation process of LDMs, in a way that is compatible with most of the LDM-based generative methods;
(2) we demonstrate how it can be used to detect and trace generated images, through a real-world evaluation benchmark;
(3) we compare to post-hoc watermarking methods, showing that it is competitive while being more secure and efficient, and (4) evaluate robustness to intentional attacks.

\section{Related Work}

\paragraph{Image generation}
has long been dominated by GANs, still state-of-the-art on many datasets~\cite{karras2020training, karras2019style, karras2020analyzing, sauer2022stylegan, walton2022stylenat}. 
Transformers have also been successfully used for modeling image~\cite{ramesh2021zero, ding2021cogview} or video~\cite{singer2022makeavideo} distributions, providing higher diversity at the expense of increased inference time. 
Images are typically converted to token lists using vector-quantized architectures~\cite{esser2021taming, razavi2019generating, yu2021vector}, relying on an image decoder.

Diffusion models~\cite{dhariwal2021diffusion, ho2020denoising, nichol2021improved, song2020denoising} have brought huge improvements in text-conditional image generation, 
being now able to synthesize high-resolution photo-realistic images for a wide variety of text prompts~\cite{balaji2022ediffi, ho2022imagenvideo, ramesh2022hierarchical, rombach2022high, saharia2022photorealistic}. 
They can also perform conditional image generation tasks -- like inpainting or text-guided image editing -- by fine-tuning the diffusion model with additional conditioning, \eg masked input image, segmentation map, etc.~\cite{lugmayr2022repaint, saharia2022palette}. 
Because of their iterative denoising algorithm, diffusion models can also be adapted for image editing in a zero-shot fashion by guiding the generative process~\cite{couairon2022diffedit, hertz2022prompt, kawar2022imagic, mokady2022null, valevski2022unitune,  wu2022unifying}.
All these methods, when applied on top of Stable Diffusion, operate in the latent space of images, requiring a latent decoder to produce an RGB image.

\vspace*{-0.5cm}
\paragraph{Detection of AI-generated/manipulated images}
is notably active in the context of deep-fakes~\cite{guarnera2020deepfake, zhao2021multi}. 
Many works focus on the detection of GAN-generated images~\cite{chai2020makes, gragnaniello2021gan, wang2020cnn, zhang2019detecting}.
One way is to detect inconsistencies in the generated images, via lights, perspective or physical objects~\cite{farid2022lighting, farid2022perspective, li2018exposing, ma2022totems, wang2019detecting}. 
These approaches are restricted to photo-realistic images or faces but do not cover artworks where objects are not necessarily physically correct.

Other approaches use traces left by the generators in the spatial~\cite{marra2019gans, yu2019attributing} or frequency~\cite{frank2020leveraging, zhang2019detecting} domains.
They have extended to diffusion models in recent works~\cite{corvi2022detection, sha2022fake}, and showed encouraging results.
However purely relying on forensics and passive detection is limiting. 
As an example, the best performing method to our knowledge~\cite{corvi2022detection} is able to detect $50\%$ of generated images for an FPR around $1$/$100$.
Put differently, if a user-generated content platform were to receive $1$ billion images every day, it would need to wrongly flag $10$ million images to detect only half of the generated images.
Besides, passive techniques cannot trace images from different versions of the same model, conversely to active ones like watermarking.

\vspace*{-0.5cm}
\paragraph{Image watermarking} 
has long been studied in the context of tracing and intellectual property protection~\cite{cox2007digital}.
More recently, deep learning encoder/extractor alternatives like HiDDeN~\cite{ahmadi2020redmark, lee2020resolution, luo2020distortion, zhang2020udh, zhu2018hidden} or iterative methods by Vukoti\'c~\etal~\cite{fernandez2022sslwatermarking,kishore2021fixed, vukotic2018deep} showed competitive results in terms of robustness to a wide range of transformations, namely geometric ones.

In the specific case of \textbf{generative models}, some works deal with watemarking the training set on which the generative model is learned~\cite{yu2021artificial}.
It is highly inefficient since every new message to be embedded requires a new training pipeline.
Merging the watermarking and the generative process is a recent idea~\cite{fei2022supervised, lin2022cycleganwm, nie2023attributing, qiao2023novel, wu2020watermarking, yu2022responsible, zhang2020model}, that is closer to the model watermarking litterature~\cite{uchida2017embedding}.
They suffer from two strong limitations.
First, these methods only apply to GAN, while LDM are beginning to replace them in most applications. 
Second, watermarking is incorporated in the training process of the GAN from the start. 
This strategy is very risky because the generative model training is more and more costly\footnote{Stable Diffusion training costs $\sim$\$600k of cloud compute (\href{https://en.wikipedia.org/wiki/Stable_Diffusion}{Wikipedia}).}.
Our work shows that a quick fine-tuning of the latent decoder part of the generative model is enough to achieve a good watermarking performance, provided that the watermark extractor is well chosen.

\begin{figure*}[b]
    \centering
    \includegraphics[width=1.0\textwidth, trim={0cm 0.7cm 0cm 0cm}, clip]{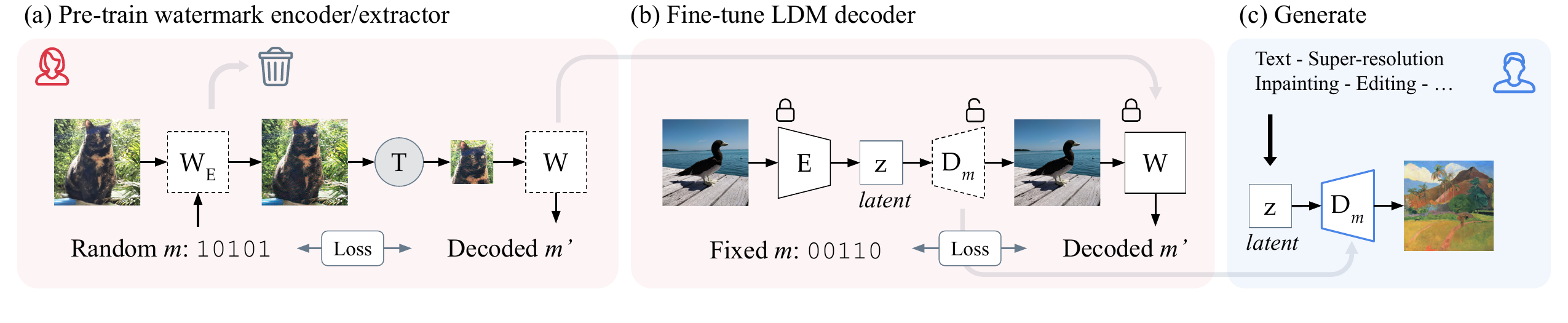}
    \caption{
        Steps of the method.
        (a) We pre-train a watermark encoder $\mathcal{W}_E$ and extractor $\mathcal{W}$, to extract binary messages.
        (b) We fine-tune the decoder $\mathcal{D}$ of the LDM's auto-encoder with a fixed signature $m$ such that all the generated images (c) lead to $m$ through $\mathcal{W}$.
        }
    \label{fig:method}
\end{figure*}

\section{Problem Statement \& Background}

\autoref{fig:fig1} shows a model provider \emph{Alice} who deploys a latent diffusion model to users \emph{Bobs}.
Stable Signature embeds a binary signature into the generated images. 
This section derives how Alice can use this signature for two scenarios:
\vspace{-0.6em}
\begin{itemize}[leftmargin=*, labelindent=0pt, label=\textbullet, itemsep=-4pt]
    \item 
    \emph{Detection: ``Is it generated by my model?''.} 
    Alice detects if an image was generated by her model.
    As many generations as possible should be flagged, while controlling the probability of flagging a natural image. 
    \item 
    \emph{Identification: ``Who generated this image?''.} 
    Alice monitors who created each image, while avoiding to mistakenly identifying a Bob who did not generate the image.
\end{itemize}

\subsection{Image watermarking for detection}
Alice embeds a $k$-bit binary signature into the generated images.
The watermark extractor then decodes messages from the images it receives and detects when the message is close to Alice's signature.
An example application is to block AI-generated images on a content sharing platform.

\vspace{-0.3cm}
\paragraph{Statistical test}\label{subsec:statistical-test}
Let $m\in \{ 0,1 \}^{k}$ be Alice's signature. 
We extract the message $m'$ from an image $x$ and compare it to $m$.
As done in previous works~\cite{lin2022cycleganwm, yu2021artificial},
the detection test relies on the number of matching bits $M(m,m')$: if
\vspace{-0.3cm}
\begin{equation} 
M\left(m,m'\right) \geq \tau \,\,\textrm{ where }\,\, \tau\in 
\{0,\ldots,k\}, \vspace{-0.3cm}
\label{eq:detectiontest}
\end{equation}
then the image is flagged.
This provides a level of robustness to imperfections of the watermarking. 

Formally, we test the statistical hypothesis $H_1$: ``$x$ was generated by Alice's model'' 
against the null hypothesis $H_0$: ``$x$ was not generated by Alice's model''.
Under $H_0$ (\ie for vanilla images), we assume that bits $m'_1,\ldots, m'_k$ are (i.i.d.) Bernoulli random variables with parameter $0.5$.
Then $M(m, m')$ follows a binomial distribution with parameters ($k$, $0.5$).
We verify this assumption experimentally in App.~\ref{app:assumption}.
The False Positive Rate (FPR) is the probability that $M(m, m')$ takes a value bigger than the threshold $\tau$.
It is obtained from the CDF of the binomial distribution, and a closed-form can be written with the regularized incomplete beta function $I_x(a;b)$:
\vspace{-0.2cm}
\begin{align}\label{eq:p-value}
    \text{FPR}(\tau) & = \mathbb{P}\left(M > \tau | H_0\right) = I_{1/2}(\tau+1, k - \tau). 
\end{align}

\subsection{Image watermarking for identification}
Alice now embeds a signature $m^{(i)}$ drawn randomly from $\{0,1\}^k$ into %
the model distributed to
Bob$^{(i)}$ (for $i=1\cdots N$, with $N$ the number of Bobs).
Alice can trace any misuse of her model: generated images violating her policy (gore content, deepfakes) are linked back to the specific Bob by comparing the extracted message to Bobs' signatures.

\vspace{-0.2cm}
\paragraph{Statistical test}\label{subsec:statistical-test-identification}
We compare the message $m'$ from the watermark extractor to $ \left( m^{(1)},\dots, m^{(N)} \right)$. %
There are now $N$ detection hypotheses to test.
If the $N$ hypotheses are rejected, we conclude that the image was not generated by any of the models.
Otherwise, we attribute the image to $\textrm{argmax}_{i=1..N} M\left(m', m^{(i)}\right)$.
With regards to the detection task, false positives are more likely since there are $N$ tests. 
The global FPR at a given threshold $\tau$ is:
\vspace{-0.2cm}
\begin{equation}\label{eq:globalFPR}
    \text{FPR}(\tau,N) = 1-(1-\text{FPR}(\tau))^N\approx N.\text{FPR}(\tau).
\vspace{-0.2cm}
\end{equation}

Equation~\eqref{eq:globalFPR} (resp.~\eqref{eq:p-value}), is used reversely: we find threshold $\tau$ to achieve a required FPR for identification (resp.~detection). 
Note that these formulae hold only under the assumption of i.i.d. Bernoulli bits extracted from vanilla images. 
This crucial point is enforced in the next section.

\section{Method}

Stable Signature modifies the generative network so that the generated images have a given signature through a fixed watermark extractor.
It is trained in two phases.
First, we create the watermark extractor network $\mathcal{W}$. 
Then, we fine-tune the Latent Diffusion Model (LDM) decoder $\mathcal{D}$, such that all generated images have a given signature through $\mathcal{W}$.

\def\payload{k}
\def\im{x}
\def\IM{\mathcal{X}}

\subsection{Pre-training the watermark extractor}\label{subsec:pre-training}

We use HiDDeN~\cite{zhu2018hidden}, a classical method in the deep watermarking literature.
It jointly optimizes the parameters of watermark encoder $\mathcal{W}_E$ and extractor network $\mathcal{W}$ to embed $\payload$-bit messages into images, robustly to transformations that are applied during training. 
We discard $\mathcal{W}_E$ after training, since only $\mathcal{W}$ serves our purpose.

Formally, $\mathcal{W}_E$ takes as inputs a cover image $\im_o \in \mathbb{R}^{W\times H\times 3}$ and a $\payload$-bit message $m\in\{0,1\}^\payload$.
Similar to ReDMark~\cite{ahmadi2020redmark}, $\mathcal{W}_E$ outputs a residual image $\delta$ of the same size as $\im_o$, that is multiplied by a factor $\alpha$ to produce watermarked image $\im_w = \im_o + \alpha \delta$.
At each optimization step an image transformation $T$ is sampled from a set $\mathcal{T}$ that includes common image processing operations such as cropping and JPEG compression\footnote{The transformation needs to be differentiable in pixel space. This is not the case for JPEG compression so we use the forward attack simulation layer introduced by Zhang~\etal~\cite{zhang2021asl}.}. 
A ``soft'' message is extracted from the transformed image: $m' = \mathcal{W}(T(\im_w))$ (at inference time, the decoded message is given by the signs of the components of $m'$).
The \emph{message loss} is the Binary Cross Entropy (BCE) between $m$ and the sigmoid $\sigma (m')$:
\vspace{-0.3cm}
\begin{align*}\label{eq:loss1}
    \mathcal{L}_{m} = - \sum_{i=1}^\payload m_i \cdot \log \sigma (m'_i) + (1-m_i) \cdot \log ( 1 - \sigma (m'_i)).
\vspace{-0.3cm}
\end{align*}

The network architectures are kept simple to ease the LDM fine-tuning in the second phase.
They are the same as HiDDeN~\cite{zhu2018hidden} (see App.~\ref{app:archi-hidden}) with two changes. 

First, since $\mathcal{W}_E$ is discarded, its perceptual quality is not as important, so the perceptual loss or the adversarial network are not needed. 
Instead, the distortion is constrained by a $\mathrm{tanh}$ function on output of $\mathcal{W}_E$ and by the scaling factor $\alpha$.
This improves the bit accuracy of the recovered message and makes it possible to increase its size $k$.

Second, we observed that $\mathcal{W}$'s output bits for vanilla images are correlated and highly biased, which violates the assumptions of Sec.~\ref{subsec:statistical-test}. 
Therefore we remove the bias and decorrelate the outputs of $\mathcal{W}$ by applying a PCA whitening transformation (more details in App.~\ref{app:hidden-centering}).

\subsection{Fine-tuning the generative model}\label{subsec:finetuning}

In LDM, the diffusion happens in the latent space of an auto-encoder.
The latent vector $z$ obtained at the end of the diffusion is input to decoder $\mathcal{D}$ to produce an image.
Here we fine-tune $\mathcal{D}$ such that the image contains a given message $m$ that can be extracted by $\mathcal{W}$.
Stable Signature is compatible with many generative tasks, since modifying only $\mathcal{D}$ does not affect the diffusion process.

First, we fix the signature $m=(m_1,\ldots, m_\payload) \in \{0,1\}^k$. 
The fine-tuning of $\mathcal{D}$ into $\mathcal{D}_m$ is inspired by the original training of the auto-encoder in LDM~\cite{rombach2022ldm}.

Training image $\im \in \mathbb{R}^{H\times W\times 3}$ is fed to the LDM encoder $\mathcal{E}$ 
that outputs activation map $z = \mathcal{E}(\im) \in \mathbb{R}^{h\times w\times c}$, downsampled by a power-of-two factor $f = H/h = W/w $.
The decoder reconstructs an image $\im' = \mathcal{D}_m(z)$ and the extractor recovers $m' = \mathcal{W} (\im')$.
The \emph{message loss} is the BCE between $m'$ and the original $m$: $\mathcal{L}_m = \mathrm{BCE}(\sigma \left( m' \right), m)$.

In addition, the original decoder $\mathcal{D}$ reconstructs the image without watermark: $\im'_o = \mathcal{D}(z)$. 
The \emph{image perceptual loss} $\mathcal{L}_\mathrm{i}$ between $\im'$ and $\im'_o$, controls the distortion.
We use the Watson-VGG perceptual loss introduced by Czolbe~\etal\cite{czolbe2020loss}, an improved version of LPIPS~\cite{zhang2018unreasonable}. %
It is essential that the decoder learns luminance and contrast masking to add less perceivable watermarks. %

The weights of $\mathcal{D}_m$ are optimized in a few backpropagation steps to minimize
\vspace{-0.3cm}
\begin{equation}\label{eq:loss2}
    \mathcal{L} = \mathcal{L}_\mathrm{m} + \lambda_\mathrm{i}~ \mathcal{L}_\mathrm{i}.
    \vspace{-0.3cm}
\end{equation}

This is done over $100$ iterations with the AdamW optimizer~\cite{loshchilov2017decoupled} and batch of size $4$, \ie the fine-tuning sees \emph{less than 500 images} and takes \emph{one minute on a single GPU}.
The learning rate follows a cosine annealing schedule with $20$ iterations of linear warmup to $10^{-4}$ and decays to $10^{-6}$.
The factor $\lambda_\mathrm{i}$ in~\eqref{eq:loss2} is set to $2.0$ by default.

\begin{figure*}[b]
    \begin{minipage}{0.33\textwidth}
        \centering
        \includegraphics[width=1.0\textwidth, trim={0cm 0cm 0cm 0cm}, clip]{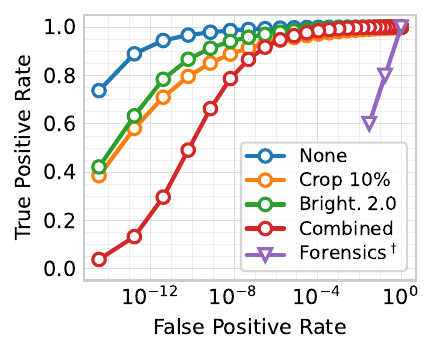}
        \caption{
            \textbf{Detection results}. TPR/FPR curve of the detection under different transformations.
            Forensics$^\dagger$ indicates passive detection (without watermark)~\cite{corvi2022detection}.
        }\label{fig:tpr-fpr}
    \end{minipage}\hspace{0.4cm}
    \begin{minipage}{0.33\textwidth}
        \centering
        \includegraphics[width=1.0\textwidth, trim={0cm 0cm 0cm 0cm}, clip]{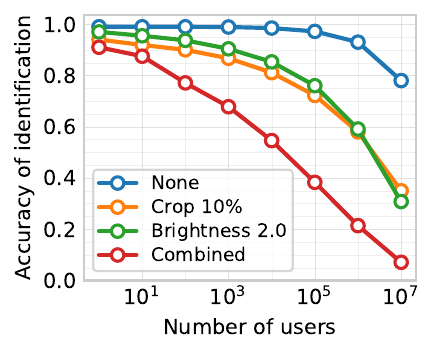}
        \caption{
            \textbf{Identification results}. 
            Proportion of well-identified users.
            Detection with FPR=$10^{-6}$ is run beforehand, and we consider it an error if the image is not flagged.
        }\label{fig:identification}
    \end{minipage}\hfill
    \begin{minipage}{0.28\textwidth}
        \centering
        \includegraphics[width=1.0\textwidth, trim={0cm 0cm 0cm 0cm}, clip]{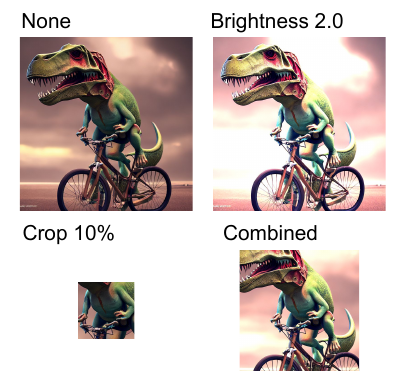}
        \caption{
            Transformations evaluated in Sec.~\ref{sec:application} \&~\ref{sec:experiments}.
            `Combined' is made of crop $50\%$, brightness adjustment $1.5$ and JPEG $80$ compression.
        }\label{fig:transformations}
    \end{minipage}
\end{figure*}

\section{Text-to-Image Watermarking Performance }\label{sec:application}
This section shows the potential of our method for detection and identification or images generated by a Stable-Diffusion-like model~\cite{rombach2022ldm}\footnote{
    We refrain from experimenting with pre-existing third-party generative models, such as Stable Diffusion or LDMs, and instead use a large diffusion model (2.2B parameters) trained on an internal dataset of 330M licensed image-text pairs.
}.
We apply generative models watermarked with $48$-bit signatures on prompts of the MS-COCO~\cite{lin2014microsoft} validation set.
We evaluate detection and identification on the outputs, as illustrated in~\autoref{fig:fig1}.

We evaluate their robustness to different transformations applied to generated images:
strong cropping ($10\%$ of the image remaining), 
brightness shift (strength factor $2.0$), 
as well as a combination of crop $50\%$, brightness shift $1.5$ and JPEG $80$. 
This covers typical geometric and photometric edits (see Fig.~\ref{fig:transformations} for visual examples).

The performance is partly obtained from experiments and partly by extrapolating small-scale measurements.

\subsection{Detection results}
For detection, we fine-tune the decoder of the LDM with a random key $m$, generate $1000$ images and use the test of Eq.~\eqref{eq:detectiontest}.
We report the tradeoff between True Positive Rate (TPR), \ie the probability of flagging a generated image and the FPR, while varying $\tau\in \{0, .. ,48\}$.
For instance, for $\tau=0$, we flag all images so $\textrm{FPR}=1$, and $\textrm{TPR}=1$.
The TPR is measured directly. 
In contrast the FPR is inferred from Eq.~\eqref{eq:p-value}, because it would otherwise be too small to be measured on reasonably sized problems 
(this approximation is validated experimentally in App.~\ref{app:fpr-check}).
The experiment is run on $10$ random signatures and we report averaged results.

\autoref{fig:tpr-fpr} shows the tradeoff under image transformations.
For example, when the generated images are not modified, Stable Signature detects $99\%$ of them, while only $1$ vanilla image out of $10^9$ is flagged.
At the same $\textrm{FPR}=10^{-9}$, Stable Signature detects $84\%$ of generated images for a crop that keeps $10\%$ of the image, 
and $65\%$ for a transformation that combines a crop, a color shift, and a JPEG compression.
For comparison, we report results of a state-of-the-art passive method~\cite{corvi2022detection}, applied on resized and compressed images.
As to be expected, we observe that these baseline results have orders of magnitudes larger FPR than Stable Signature, which actively marks the content.

\subsection{Identification results}
Each Bob has its own copy of the generative model. 
Given an image, the goal is to find if any of the $N$ Bobs created it (detection) and if so, which one (identification).
There are $3$ types of error: 
\emph{false positive}: flag a vanilla image; 
\emph{false negative}: miss a generated image; 
\emph{false accusation}: flag a generated image but identify the wrong user.

For evaluation, we fine-tune $N'=1000$ models with random signatures.
Each model generates $100$ images.
For each of these $100$k watermarked images, we extract the Stable Signature message, compute the matching score with all $N$ signatures and select the user with the highest score. 
The image is predicted to be generated by that user if this score is above threshold $\tau$.
We determined $\tau$ such that $\textrm{FPR}=10^{-6}$, see Eq.~\eqref{eq:globalFPR}. 
For example, for $N=1$, $\tau=41$ and for $N=1000$, $\tau=44$.
Accuracy is extrapolated beyond the $N'$ users by adding additional signatures and having $N > N'$ (\eg users that have not generated any images).

\autoref{fig:identification} reports the per-transformation identification accuracy.
For example, we identify a user among $N$=$10^{5}$ with $98\%$ accuracy when the image is not modified.
Note that for the combined edit, this becomes $40\%$.
This may still be dissuasive:
if a user generates $3$ images, he will be identified $80\%$ of the time.
We observe that at this scale, the false accusation rate is zero, \ie we never identify the wrong user.
This is because $\tau$ is set high to avoid FPs, which also makes false accusations unlikely.
We observe that the identification accuracy decreases when $N$ increases, because the threshold $\tau$ required to avoid false positives is higher when $N$ increases, as pointed out by the approximation in~\eqref{eq:globalFPR}.
In a nutshell, by distributing more models, Alice trades some accuracy of detection against the ability to identify users.

\section{Experimental Results}\label{sec:experiments}

We presented in the previous section how to leverage watermarks for detection and identification of images generated from text prompts.
We now present more general results on robustness and image quality for different generative tasks.
We also compare Stable Signature to other watermarking algorithms applied post-generation.

\begin{SCtable*}
    \centering
    \footnotesize
    \setlength{\tabcolsep}{4pt}
        \begin{tabular}{ c l @{\hspace{2pt}} l  *{2}{c} *{4}{p{15pt}}}
        \toprule
       & & \multirow{2}{*}{}          & \multirow{2}{*}{PSNR / SSIM $\uparrow$} & \multirow{2}{*}{FID $\downarrow$} &\multicolumn{4}{c}{Bit accuracy $\uparrow$ on:} \\ 

    &                                         &                           &           &                                            &  None & Crop & Brigh. & Comb.  \\ \midrule
\multirow{6}{*}{\rotatebox[origin=c]{90}{Tasks}}  
        & Text-to-Image                      & LDM~\cite{rombach2022ldm}             & $30.0$ / $0.89$   & $19.6$ \color{orange}{($-0.3$)} & $0.99$ & $0.95$ & $0.97$ & $0.92$ \\ \cmidrule{2-9}
       & Image Edition                       & DiffEdit~\cite{couairon2022diffedit}                  & $31.2$ / $0.92$ & $15.0$ \color{orange}{($-0.3$)}       & $0.99$ & $0.95$ & $0.98$ & $0.94$ \\ \cmidrule{2-9}
       & Inpainting - Full          & \multirow{2}{*}{Glide~\cite{nichol2021glide}}   & $31.1$ / $0.91$  & $16.8$ \color{orange}{($+0.6$)} & $0.99$ & $0.97$ & $0.98$ & $0.93$ \\ 
       & {\color{white}Inpa} - Mask only          &                                   & $37.8$ / $0.98$  & $9.0$~~ \color{orange}{($+0.1$)} & $0.89$ & $0.76$ & $0.84$ & $0.78$\\ \cmidrule{2-9}
       & Super-Resolution & LDM~\cite{rombach2022ldm}  & $34.0$ / $0.94$ & $11.6$ \color{orange}{($+0.0$)}      & $0.98$ & $0.93$ & $0.96$ & $0.92$ \\ 
    \midrule \rule{0pt}{8pt} \rule{0pt}{8pt} 
\multirow{7}{*}{\rotatebox[origin=c]{90}{WM Methods}} 
           & \emph{Post generation} \\
           & Dct-Dwt \cite{cox2007digital}                      & $0.14$ (s/img)      &  $39.5$ / $0.97$  & $19.5$ \color{orange}{($-0.4$)} & $0.86$ & $0.52$ & $0.51$ & $0.51$ \\ 
           & SSL Watermark \cite{fernandez2022sslwatermarking}  & $0.45$ (s/img)      &  $31.1$ / $0.86$  & $20.6$ \color{orange}{($+0.7$)} & $1.00$ & $0.73$ & $0.93$ & $0.66$ \\ 
           & FNNS \cite{kishore2021fixed}                       & $0.28$ (s/img)      &  $32.1$ / $0.90$  & $19.0$ \color{orange}{($-0.9$)} & $0.93$ & $0.93$ & $0.91$ & $0.93$ \\ 
           & HiDDeN \cite{zhu2018hidden}                        & $0.11$ (s/img)      &  $32.0$ / $0.88$  & $19.7$ \color{orange}{($-0.2$)} & $0.99$ & $0.97$ & $0.99$ & $0.98$ \\ \cmidrule{2-9}
           & \emph{Merged in generation} \\
           & Stable Signature                            & $0.00$ (s/img)             &  $30.0$ / $0.89$ & $19.6$ \color{orange}{($-0.3$)}     & $0.99$ & $0.95$ & $0.97$ & $0.92$ \\ 
        \bottomrule \vspace*{-0.2cm}
    \end{tabular}
    \caption{
        Generation quality and comparison to post-hoc watermarking on 512$\times$512 images and $48$-bit signatures.
        PSNR and SSIM are computed between generations of the original and watermarked generators.
        For FID, we show in {\color{orange} (color)} the difference with regards to original.
        Post-hoc watermarking is evaluated on text-generated images.
        (App.~\ref{sec:supp-robustness} gives results on more transformations, and App.~\ref{app:implementation-details} gives more details on the evaluations.)
        Overall, Stable Signature has minimal impact on generation quality. It has comparable robustness to post-hoc methods while being rooted in the generation itself.
        \vspace*{-0.2cm}
    }\label{tab:quality-watermarking} 
\end{SCtable*}

\subsection{Tasks \& evaluation metrics}
Since our method only involves the LDM decoder, it makes it compatible with many generative tasks. 
We evaluate text-to-image generation and image edition on the validation set of MS-COCO~\cite{lin2014microsoft}, super-resolution and inpainting on the validation set of ImageNet~\cite{deng2009imagenet} 
(all evaluation details are available in App.~\ref{app:evaluation}).

We evaluate the image distortion with the Peak Signal-to-Noise Ratio (PSNR), which is defined as $\mathrm{PSNR}(x,x') = -10\cdot \log_{10} (\mathrm{MSE}(x,x'))$, for $x,x'\in [0,1]^{c\times h\times w}$, as well as Structural Similarity score (SSIM)~\cite{wang2004image}.
They compare images generated with and without watermark. 
On the other hand, we evaluate the diversity and quality of the generated images with the Fr\'echet Inception Distance (FID)~\cite{heusel2017gans}.
The bit accuracy -- the percentage of bits correctly decoded -- evaluates the watermarks' robustness.

\subsection{Image generation quality}
\begin{figure}[b]
    \centering
    \scriptsize
    \setlength{\tabcolsep}{0pt}
    \resizebox{0.99\linewidth}{!}{
    \begin{tabular}{c @{\hspace{0.1cm}} c @{\hspace{0.1cm}} c}
        \toprule
        Generated with original & Generated with watermark & Pixel-wise difference ($\times 10$) \\
        \midrule
        \includegraphics[width=0.33\linewidth]{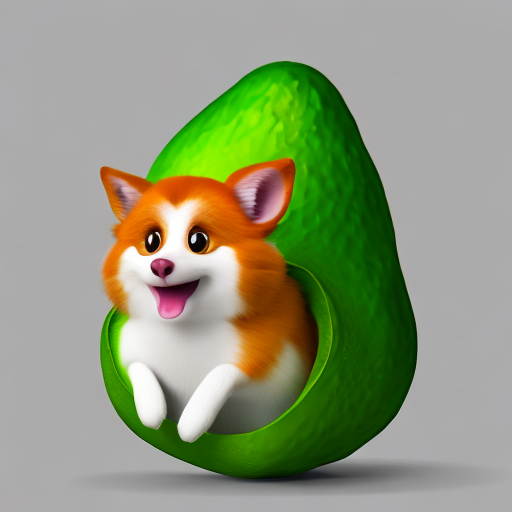} &
        \includegraphics[width=0.33\linewidth]{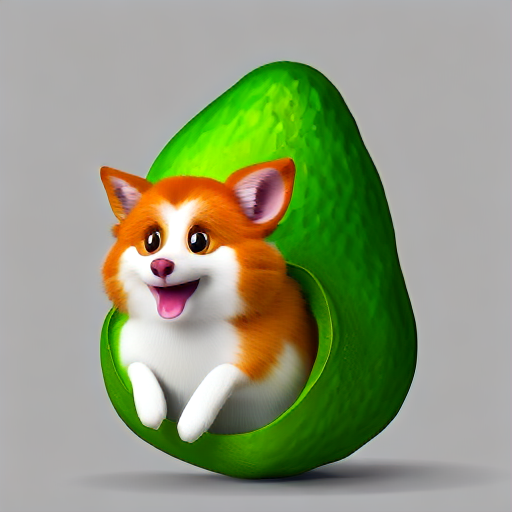} &
        \includegraphics[width=0.33\linewidth]{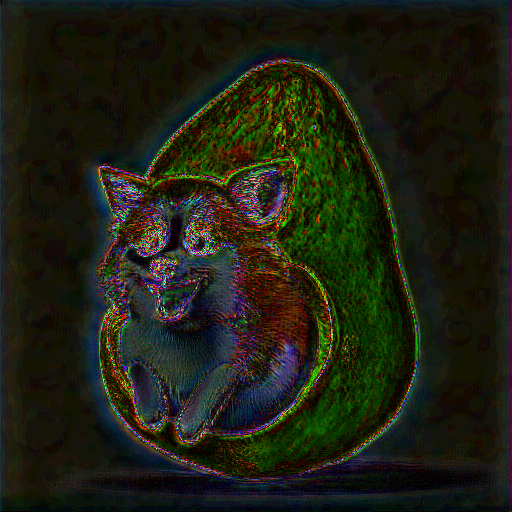} \\ 
        \rule{0pt}{0.2cm}
        \includegraphics[width=0.33\linewidth]{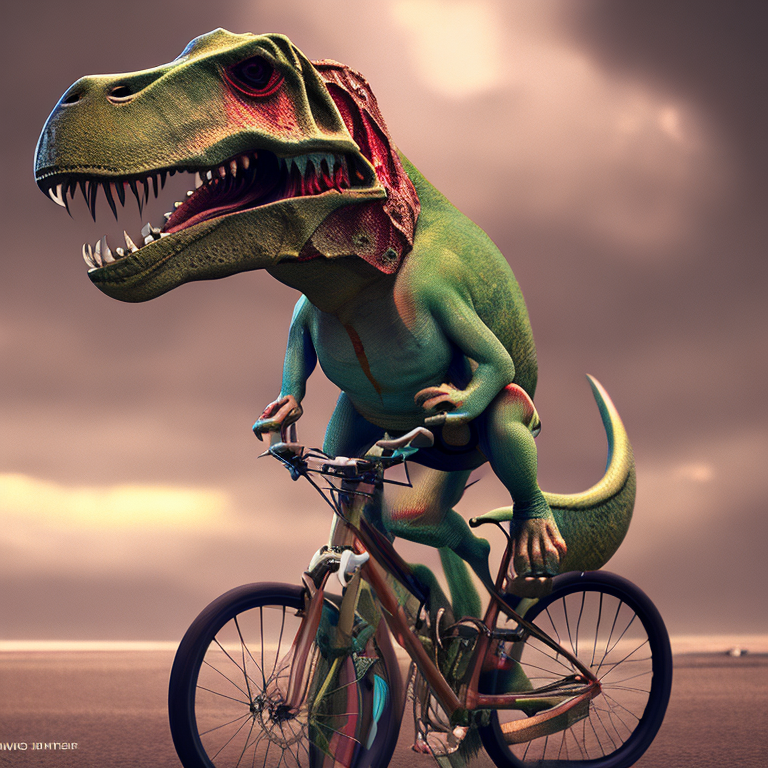} &
        \includegraphics[width=0.33\linewidth]{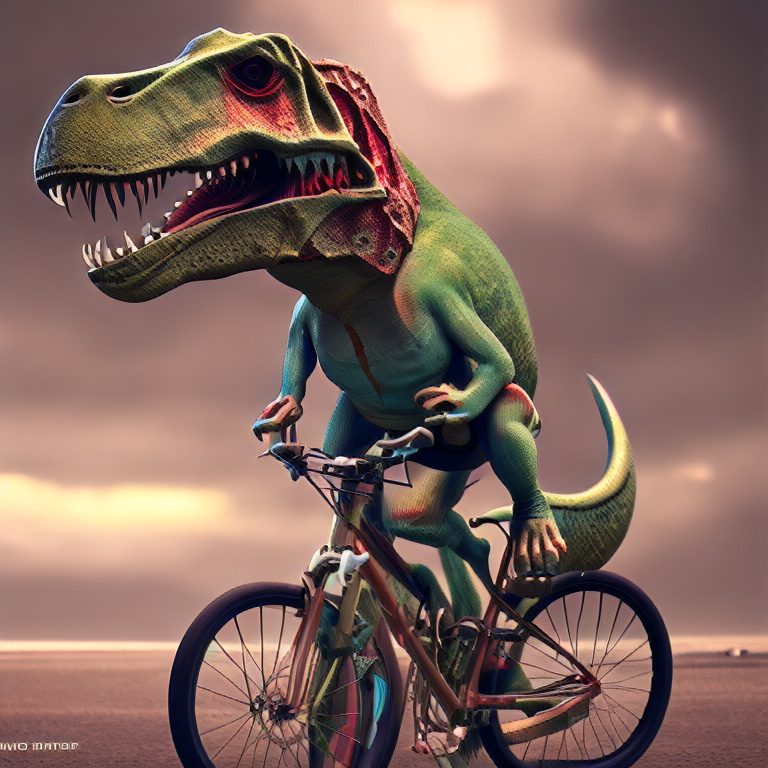} &
        \includegraphics[width=0.33\linewidth]{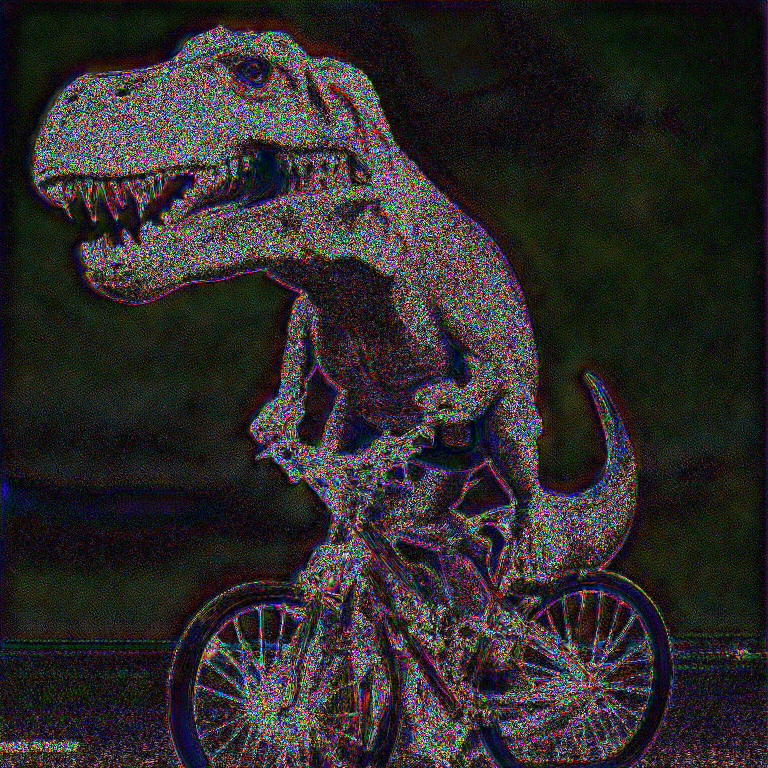} \\       
        \bottomrule\\
    \end{tabular}
    }
    \caption{
    Images generated with Stable Diffusion. 
    The PSNR is $35.4$\,dB in the first row and $28.6$\,dB in the second.
    Images generated with Stable Signature look natural because modified areas are located where the eye is not sensitive.
    More examples in App.~\ref{app:qualitative}.
    }
    \label{fig:qualitative}
\end{figure}

\autoref{fig:qualitative} shows qualitative examples of how the image generation is altered by the latent decoder's fine-tuning. 
The difference is very hard to perceive even for a trained eye. 
This is surprising for such a low PSNR, especially since the watermark embedding is not constrained by any Human Visual System like in professional watermarking techniques. 
Most interestingly, the LDM decoder has indeed learned to add the watermark signal only over textured areas where the human eyes are not sensitive, while the uniform backgrounds are kept intact (see the pixel-wise difference).

\autoref{tab:quality-watermarking} presents a quantitative evaluation of image generation quality on the different tasks.
We report the FID, and the average PSNR and SSIM that are computed between the images generated by the fine-tuned LDM and the original one.
The results show that no matter the task, the watermarking has very small impact on the FID of the generation.

The average PSNR is around $30$~dB and SSIM around $0.9$ between images generated by the original and a  watermarked model.
They are a bit low from a watermarking perspective because we do not explicitly optimize for them.
Indeed, in a real world scenario, one would only have the watermarked version of the image. 
Therefore we don't need to be as close as possible to the original image but only want to generate artifacts-free images. Without access to the image generated by the original LDM, it is very hard to tell whether a watermark is present or not.

\begin{table}[t]
    \centering
    \caption{
        Watermark robustness on image transformations applied before decoding, details of which are available in App.~\ref{app:evaluation}.
        We report the bit accuracy, averaged over $10\times1$k images generated from COCO prompts with $10$ different keys.
        }\label{tab:robustness}
        \footnotesize
        \vspace*{0.2cm}
        \setlength{\tabcolsep}{4pt}
        \resizebox{0.96\linewidth}{!}{
            \begin{tabular}{ll|ll|ll}
                \toprule
                \bf{Attack}             & \bf{Bit acc.}     & Comb.         & $0.92$    & Sharpness $2.0$ & $0.99$ \\
                None & $0.99$                               & Bright. $2.0$  & $0.97$    & Med. Filter $k$=7  & $0.94$ \\
                Crop $0.1$ & $0.95$                         & Cont. $2.0$    & $0.98$    & Resize $0.7$  & $0.91$  \\
                JPEG $50$ & $0.88$                          & Sat. $2.0$  & $0.99$      & Text overlay    & $0.99$ \\
                \bottomrule
            \end{tabular}
        }
        \vspace*{-0.3cm}
    \end{table}

\subsection{Watermark robustness}\label{subsec:robustness}
We evaluate the robustness of the watermark to different image transformations applied before extraction.
For each task, we generate $1$k images with $10$ models fine-tuned for different messages, and report the average bit accuracy in \autoref{tab:quality-watermarking}.
Additionally, \autoref{tab:robustness} reports results on more image transformations for images generated from COCO prompts.
The main evaluated transformations are presented in Fig.~\ref{fig:transformations} (more evaluation details are available in App.~\ref{app:evaluation}).

We see that the watermark is indeed robust for several tasks and across transformations.
The bit accuracy is always above $0.9$, except for inpainting, when replacing only the masked region of the image (between $1-50$\% of the image, with an average of $27\%$ across masks).
Besides, the bit accuracy is not perfect even without edition, mainly because there are images that are harder to watermark (\eg the ones that are very uniform, like the background in Fig.~\ref{fig:qualitative}) and for which the accuracy is lower.

Note that the robustness comes even without any transformation during the LDM fine-tuning phase:
it is due to the watermark extractor.
If the watermark embedding pipeline is learned to be robust against an augmentation, then the LDM will learn how to produce watermarks that are robust against it during fine-tuning.

\subsection{Comparison to post-hoc watermarking}\label{subsec:watermarking}

An alternative way to watermark generated images is to process them after the generation (post-hoc). 
This may be simpler, but less secure and efficient than Stable Signature.
We compare our method to a frequency based method, DCT-DWT~\cite{cox2007digital},
iterative approaches (SSL Watermark~\cite{fernandez2022sslwatermarking} and FNNS~\cite{kishore2021fixed}), and an encoder/decoder one like HiDDeN~\cite{zhu2018hidden}.
We choose DCT-DWT since it is employed by the original open source release of Stable Diffusion~\cite{2022stablediffusion}, and the other methods because of their performance and their ability to handle arbitrary image sizes and number of bits.
We use our implementations (see details in App.~\ref{app:watermarking}).

\autoref{tab:quality-watermarking} compares the generation quality and the robustness over $5$k generated images.
Overall, Stable Signature achieves comparable results in terms of robustness. 
HiDDeN's performance is a bit higher but its output bits are not i.i.d. meaning that it cannot be used with the same guarantees as the other methods.
We also observe that post-hoc generation gives worse qualitative results, images tend to present artifacts (see Fig.~\ref{fig:supp-watermark} in the supplement).
One explanation is that Stable Signature is merged into the high-quality generation process with the LDM auto-encoder model, which is able to modify images in a more subtle way.

\subsection{Can we trade image quality for robustness?}\label{subsec:quality-tradeoff}

We can choose to maximize the image quality or the robustness of the watermark thanks to the weight $\lambda_i$ of the perceptual loss in~\eqref{eq:loss2}.
We report the average PSNR of $1$k generated images, as well as the bit accuracy obtained on the extracted message for the `Combined' editing applied before detection (qualitative results are in App.~\ref{sec:supp-percep-loss}).
A higher $\lambda_i$ leads to an image closer to the original one, but to lower bit accuracies on the extracted message, see \autoref{tab:tradeoff}.

\begin{table}[t]
        \centering
        \caption{Quality-robustness trade-off during fine-tuning.}\label{tab:tradeoff}
        \resizebox{0.9\linewidth}{!}{
        \begin{tabular}{l *{7}{c@{\hspace*{8pt}}}}
            \toprule
            $\lambda_i$ for fine-tuning     & $0.8$ & $0.4$ & $0.2$ & $0.1$ & $0.05$ & $0.025$ \\ \midrule
            \rule{0pt}{2ex}
            PSNR $\uparrow$ & $31.4$ & $30.6$ & $29.7$ & $28.5$ & $26.8$ & $24.6$ \\ 
            \rule{0pt}{2ex}
            Bit acc. $\uparrow$ on `comb.' & $0.85$ & $0.88$ & $0.90$ & $0.92$ & $0.94$ & $0.95$ \\ 
            \bottomrule 
            \vspace*{-0.4cm}
        \end{tabular}
        }
\end{table}

\subsection{Attack simulation layer}\label{subsec:message-decoder}

\begin{table}[t]
    \centering
    \caption{Role of the attack simulation layer at pre-training.}\label{tab:asl}
    \resizebox{0.95\linewidth}{!}{
    \begin{tabular}{c@{\hspace*{4pt}} *{5}{c@{\hspace*{8pt}}}}
        \toprule
        \multirow{2}{*}{ \shortstack{ Seen at  \\ $\mathcal{W}$ training \vspace*{-4pt}} } & \multicolumn{5}{c}{Bit accuracy $\uparrow$ at test time:} \\ \cmidrule{2-6}
            & Crop $0.1$ & Rot. $90$ &JPEG $50$ & Bright. $2.0$ & Res. $0.7$  \\
        \midrule
        \xmark     & 1.00 & 0.56 & 0.50 & 0.99 & 0.48 \\
        \cmark     & 1.00 & 0.99 & 0.90 & 0.99 & 0.91 \\
        \bottomrule
        \vspace*{-0.8cm} 
    \end{tabular} 
    }
\end{table}

Watermark robustness against image transformations depends solely on the watermark extractor.
here, we pre-train them with or without specific transformations in the simulation layer, on a shorter schedule of $50$ epochs, with $128\times 128$ images and $16$-bits messages.
From there, we plug them in the LDM fine-tuning stage and we generate $1$k images from text prompts.
We report the bit accuracy of the extracted watermarks in \autoref{tab:asl}.
The extractor is naturally robust to some transformations, such as crops or brightness, without being trained with them, while others, like rotations or JPEG, require simulation during training for the watermark to be recovered at test time.
Empirically we observed that adding a transformation improves results for the latter, but makes training more challenging.

\section{Attacks on Stable Signature's Watermarks}\label{sec:attacks}

We examine the watermark's resistance to intentional tampering, 
as opposed to distortions that happen without bad intentions like crops or compression (discussed in Sec.~\ref{sec:application}). 
We consider two threat models: one  is typical for many image watermarking methods~\cite{cox2007digital} and operates at the image level, and another targets the generative model level. 
For image-level attacks, we evaluate on $5$k images generated from COCO prompts.
Full details on the following experiments can be found in Appendix~\ref{app:attacks}. 

\subsection{Image-level attacks}

\begin{figure}[b]
    \centering
    \hspace{-0.4cm}
    \includegraphics[width=\linewidth, trim={0 0 0 0}, clip]{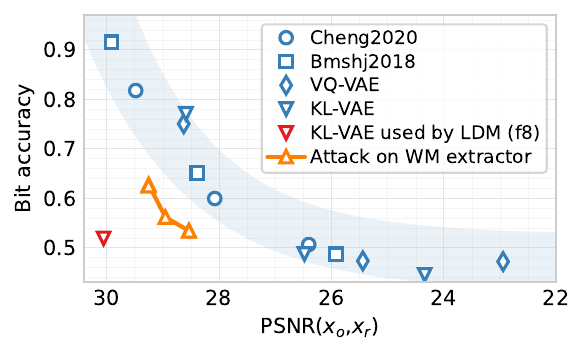}
    \caption{\textbf{Removal attacks.}
    $x_o$ is the image produced by the original generator, 
    $x_r$ is the version produced by the watermarked generator and then attacked.
    Bit accuracy is on the watermark extracted from $x_r$.
    Neural auto-encoders follow the \colorbox[HTML]{ebf2f8}{same trend}, except with the one used by the LDM (`KL-f8' for our LDM).
    When access to the watermark extractor is granted, adversarial attacks also remove the watermark at lower PSNR budget.
    }
    \label{fig:purification}
\end{figure}

\paragraph{Watermark removal.}
Bob alters the image to remove the watermark with deep learning techniques, like methods used for adversarial purification~\cite{shi2021online, yoon2021adversarial} or neural auto-encoders~\cite{abdelnabi2021adversarial, liu2020defending}.
Note that this kind of attacks has not been explored in the image watermarking literature to our knowledge.
\autoref{fig:purification} evaluates the robustness of the watermark against neural auto-encoders~\cite{balle2018variational, cheng2020learned, esser2021taming, rombach2022ldm} at different compression rates.
To reduce the bit accuracy closer to random (50\%), the image distortion needs to be strong (PSNR$<$26).
However, assuming the attack is \emph{informed on the generative model}, \ie the auto-encoder is the same as the one used to generate the images, the attack becomes much more  effective.
It erases the watermark while achieving high quality (PSNR$>$29).
This is because the image is modified precisely in the bandwidth where the watermark is embedded.
Note that this assumption is strong, because Alice does not need to distribute the original generator.

\vspace*{-0.2cm}
\paragraph{Watermark removal \& embedding (white-box).}
To go further, we assume that the attack is \emph{informed on the watermark extractor} -- \eg because it has leaked.
Bob can use an adversarial attack to remove the watermark by optimizing the image under a PSNR constraint.
The objective is to minimize the $\ell_2$ distance between a random binary message sampled beforehand and the extractor's output, effectively replacing the original signature with a random one.
It makes it possible to erase the watermark with a lower distortion budget, as seen in Fig.~\ref{fig:purification}.

Instead of removing the watermark, an attacker could embed a signature into vanilla images (unauthorized embedding~\cite{cox2007digital}) to impersonate another Bob of whom they have a generated image. 
It highlights the importance of keeping the watermark extractor private.

\subsection{Network-level attacks}

\paragraph{Model purification.} 
Bob gets Alice's generative model and uses a fine-tuning process akin to Sec.~\ref{subsec:finetuning} to eliminate the watermark embedding -- that we coin \emph{model purification}. 
This involves removing the message loss $\mathcal L_m$, and shifting the focus to the perceptual loss $\mathcal L_i$ between the original image and the one reconstructed by the LDM auto-encoder.

\begin{figure}[b]
    \centering
    \includegraphics[width=0.9\linewidth, trim={0 0 0 0}, clip]{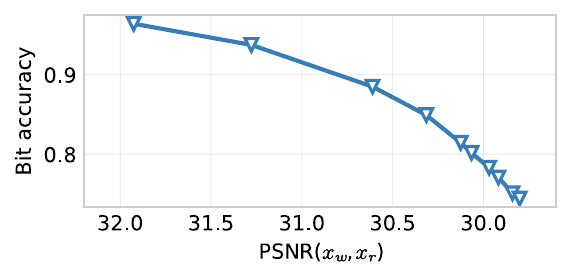}
    \caption{
        \textbf{Robustness to model purification}, \ie fine-tuning the model to remove watermarks. 
         $x_w$ is the watermarked image, $x_{r}$ is generated with the purified model at different steps of the process.
    }\label{fig:purification-model}
\end{figure}

\autoref{fig:purification-model} shows the results of this attack for the MSE loss.
The PSNR between the watermarked and purified images is plotted at various stages of fine-tuning.
Empirically, it is difficult to significantly reduce the bit accuracy without compromising the image quality: artifacts start to appear during the purification.

\vspace*{-0.2cm}
\paragraph{Model collusion.}
Users may collude by aggregating their models.
For instance, Bob$^{(i)}$ and Bob$^{(j)}$ can average the weights of their models (like Model soups~\cite{wortsman2022model}) creating a new model to deceive identification.
We found that the bit at position $\ell$ output by the extractor will be $0$ (resp. $1$) when the $\ell$-th bits of Bob$^{(i)}$ and Bob$^{(j)}$ are both $0$ (resp. $1$), and that the extracted bit is random when their bits disagree.
We show the distributions of the soft bits (before thresholding) output by the watermark extractor on images generated by the average model. 
The $\ell$-th output is labeled by bits of Bob$^{(i)}$ and Bob$^{(j)}$ (\texttt{00} means both have \texttt{0} at position $\ell$):
\\[4pt]
{ \centering
    \includegraphics[width=1.0\linewidth, trim={0 0.7cm 0 0cm}, clip]{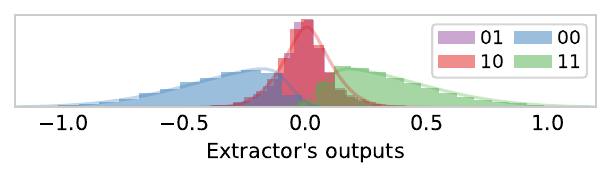}
}
\vspace{-0.1cm}

This so-called \emph{marking assumption} plays a crucial role in traitor tracing literature~\cite{furon:hal-00757152,meerwald:hal-00740964, tardos2008optimal}.
Surprisingly, it holds even though our watermarking process is not explicitly designed for it.
The study has room for improvement, such as creating user identifiers with more powerful traitor tracing codes~\cite{tardos2008optimal} and using more powerful traitor accusation algorithms~\cite{furon:hal-00757152,meerwald:hal-00740964}.
Importantly, we found the precedent remarks also hold if the colluders operate at the image level.

\section{Conclusion \& Discussion}\label{sec:conclusion}

By a quick fine-tuning of the decoder of Latent Diffusion Models, we can embed watermarks in all the images they generate.
This does not alter the diffusion process, making it compatible with most of LDM-based generative models.
These watermarks are robust, invisible to the human eye and can be employed to \emph{detect} generated images and \emph{identify} the user that generated it, with very high performance.

The public release of image generative models has an important societal impact.
With this work, we put to light the usefulness of using watermarking instead of relying on passive detection methods.
We hope it will encourage researchers and practitioners to employ similar approaches before making their models publicly available.

\vspace{-0.3cm}
\paragraph{Reproducibility Statement.}

Although the diffusion-based generative model has been trained on an internal dataset of licensed images, we use the KL auto-encoder from LDM~\cite{rombach2022ldm} with compression factor $f=8$.
This is the one used by open-source alternatives.
Code is available at \href{https://github.com/facebookresearch/stable_signature}{github.com/facebookresearch/stable\_signature}.

\vspace{-0.3cm}
\paragraph{Environmental Impact.}
We do not expect any environmental impact specific from this work.
The cost of the experiments and the method is high, though order of magnitudes less than other computer vision fields. 
We roughly estimated that the total GPU-days used for running all our experiments to $2000$, or $\approx 50000$ GPU-hours.
This amounts to total emissions in the order of 10 tons of CO$_2$eq.
This is excluding the training of the generative model itself, since we did not perform that training. 
Estimations are conducted using the \href{https://mlco2.github.io/impact#compute}{Machine Learning Impact calculator} presented by Lacoste~\etal\cite{lacoste2019quantifying}.
We do not consider in this approximation: memory storage, CPU-hours, production cost of GPUs/ CPUs, etc.

{\small
    \bibliographystyle{ieee_fullname}
    \bibliography{references}
}

\clearpage
\appendix
\twocolumn[{%
 \noindent
 \LARGE Supplementary Material\\[0.5em]
 \large The Stable Signature: Rooting Watermarks in Latent Diffusion Models\\[1em]
}]

\section{Implementation Details \& Parameters}\label{app:implementation-details}

\subsection{Details on the watermark encoder/extractor}\label{app:hidden}

\paragraph*{Architectures of the watermark encoder/extractor.}\label{app:archi-hidden}
We keep the same architecture as in HiDDeN~\cite{zhu2018hidden}, which is a simple convolutional encoder and extractor.
The encoder consist of $4$ Conv-BN-ReLU blocks, with $64$ output filters, $3\times 3$ kernels, stride $1$ and padding $1$.
The extractor has $7$ blocks, followed by a block with $k$ output filters ($k$ being the number of bits to hide), an average pooling layer, and a $k\times k$ linear layer.
For more details, we refer the reader to the original paper~\cite{zhu2018hidden}.

\paragraph{Optimization.}
We train on the MS-COCO dataset~\cite{lin2014microsoft}, with $256 \times 256$ images.
The number of bits is $k=48$, and the scaling factor is $\alpha=0.3$.
The optimization is carried out for $300$ epochs on $8$ GPUs, with the Lamb optimizer~\cite{you2019lamb} (it takes around a day). 
The learning rate follows a cosine annealing schedule with $5$ epochs of linear warmup to $10^{-2}$, and decays to $10^{-6}$.
The batch size per GPU is $64$.

\paragraph*{Attack simulation layer.}\label{app:hidden-attack}
The attack layer produces edited versions of the watermarked image to improve robustness to image processing.
It takes as input the image output by the watermark encoder $x_w = \mathcal{W}_E(x_o)$ and outputs a new image $x'$ that is fed to the decoder $\mathcal{W}$.
This layer is made of cropping, resizing, or identity chosen at random in our experiments, unless otherwise stated.
The parameter for the crop or resize is set to $0.3$ or $0.7$ with equal probability.
This is followed by a JPEG compression with probability $0.5$.
The parameter for the compression is set to $50$ or $80$ with equal probability.
This last layer is not differentiable, therefore we back-propagate only through the difference between the uncompressed and compressed images:
$x'= x_{\mathrm{aug}} + \mathrm{nograd}(x_{\mathrm{aug}, \mathrm{JPEG}} - x_{\mathrm{aug}})$~\cite{zhang2021asl}.

\paragraph*{Whitening.}\label{app:hidden-centering}
At the end of the training, we whiten the output of the watermark extractor to make the hard thresholded bits independently and identically Bernoulli distributed on vanilla images (so that the assumption of~\ref{subsec:statistical-test} holds better, see App.~\ref{app:assumption}).
We perform the PCA of the output of the watermark extractor on a set of $10$k vanilla images, and get the mean $\mu$ and eigendecomposition of the covariance matrix $\Sigma= U\Lambda U^T$. 
The whitening is applied with a linear layer with bias $-\Lambda^{-1/2}U^T\mu$ and weight $\Lambda^{-1/2}U^T$, appended to the extractor.

\begin{figure*}
    \centering
    \resizebox{0.85\linewidth}{!}{
    \begin{tabular}{*{5}{l}}
        Crop 0.1 & JPEG 50 & Resize 0.7 & Brightness 2.0 & Contrast 2.0 \\ 
        \begin{minipage}{.16\linewidth}\centering \includegraphics[width=0.3\linewidth]{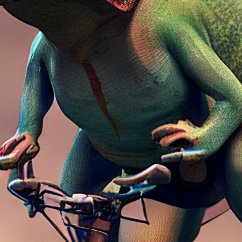}\end{minipage} &  
        \begin{minipage}{.16\linewidth}\includegraphics[width=\linewidth]{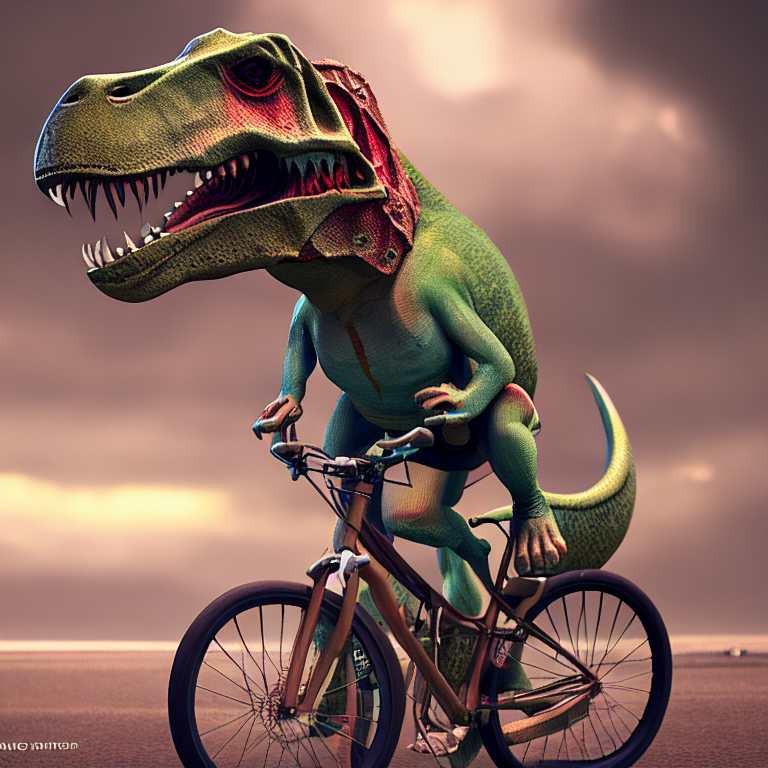}\end{minipage} &  
        \begin{minipage}{.16\linewidth}\centering\includegraphics[width=0.8\linewidth]{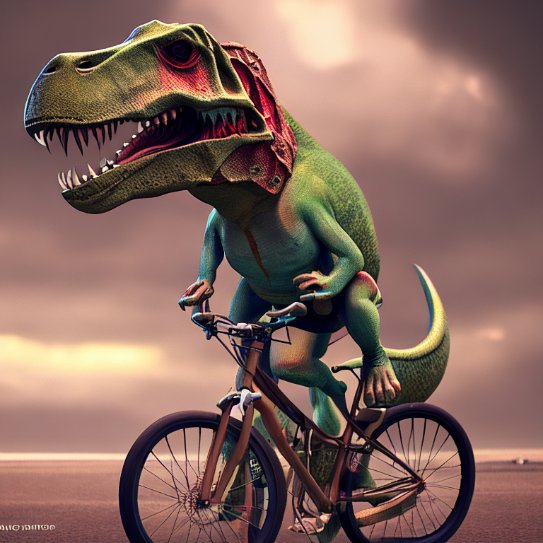}\end{minipage} &  
        \begin{minipage}{.16\linewidth}\includegraphics[width=\linewidth]{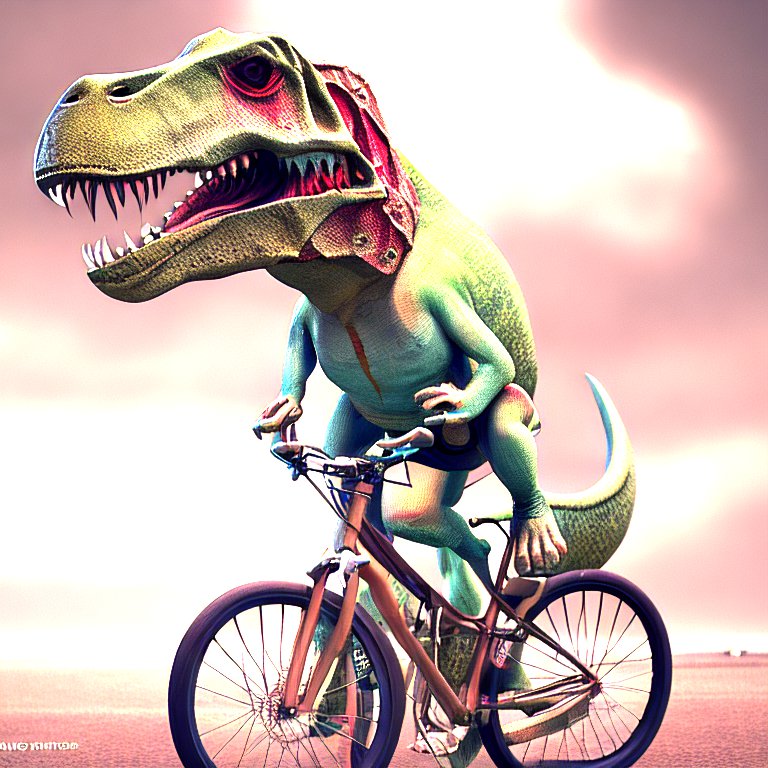}\end{minipage} &
        \begin{minipage}{.16\linewidth}\includegraphics[width=\linewidth]{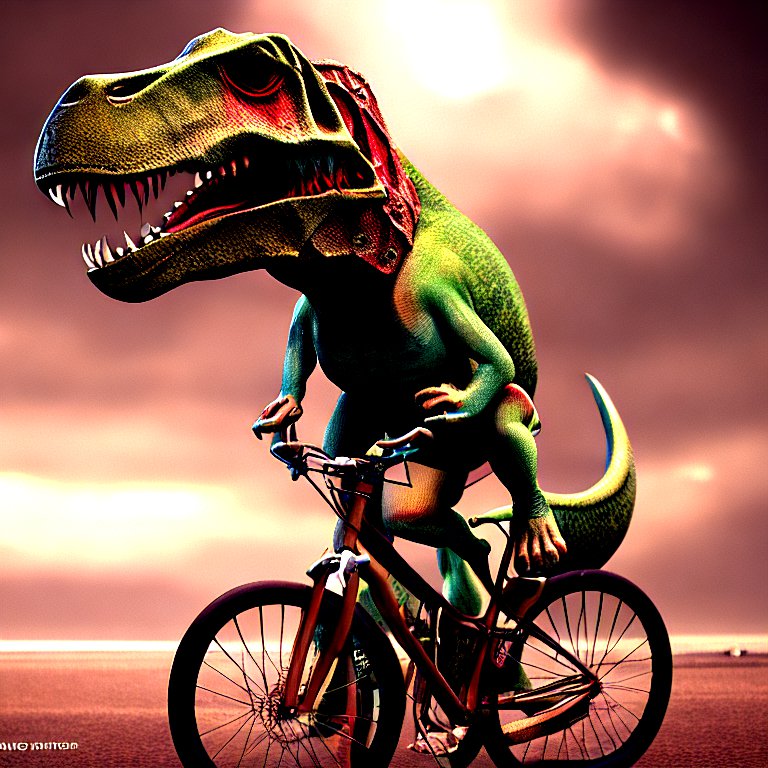}\end{minipage} 
        \\ \\
        Saturation 2.0 & Sharpness 2.0 & Rotation $90$ & Text overlay & Combined \\
        \begin{minipage}{.16\linewidth}\includegraphics[width=\linewidth]{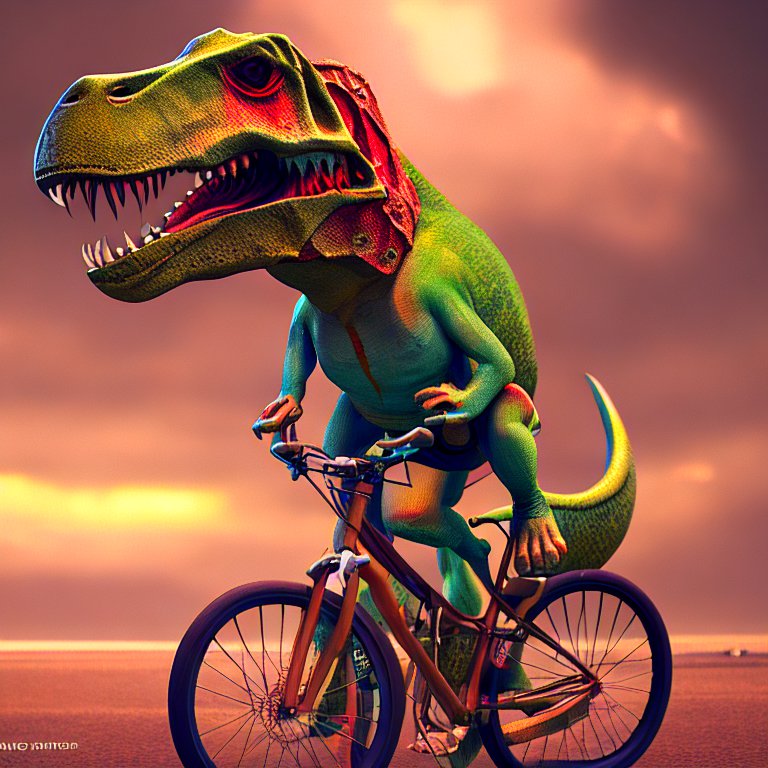}\end{minipage} &  
        \begin{minipage}{.16\linewidth}\includegraphics[width=\linewidth]{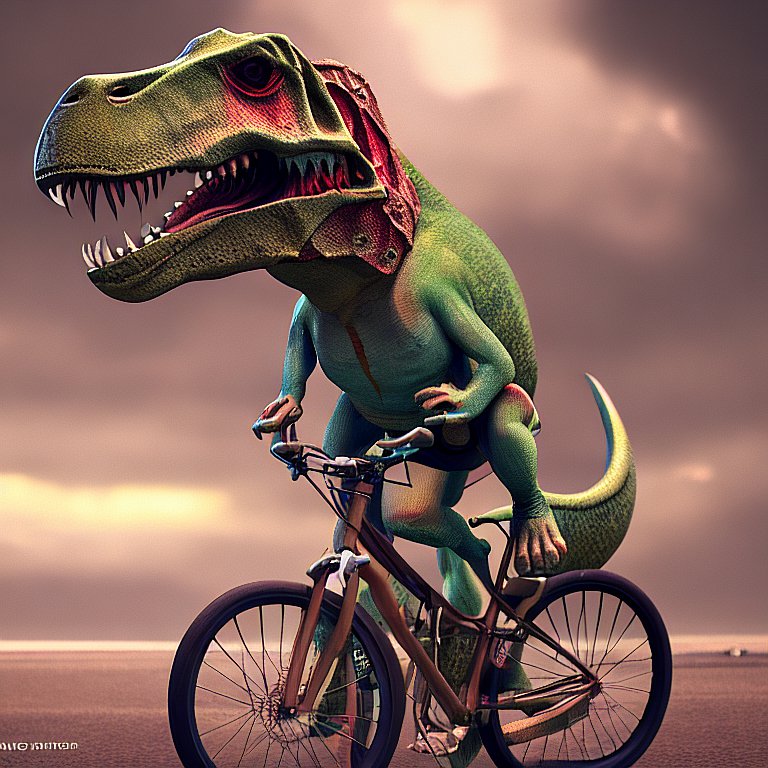}\end{minipage} &  
        \begin{minipage}{.16\linewidth}\includegraphics[width=\linewidth]{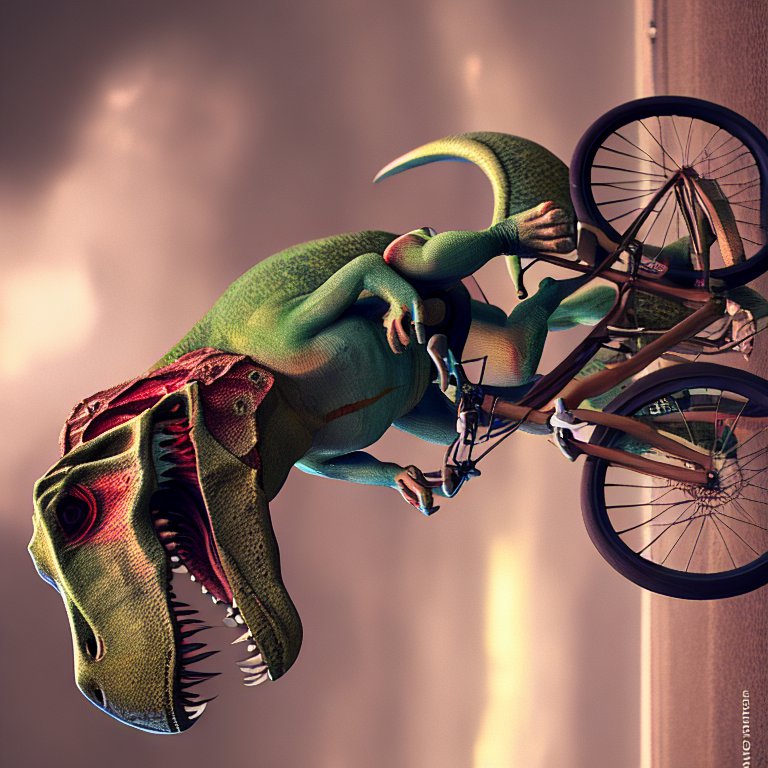}\end{minipage} &  
        \begin{minipage}{.16\linewidth}\includegraphics[width=\linewidth]{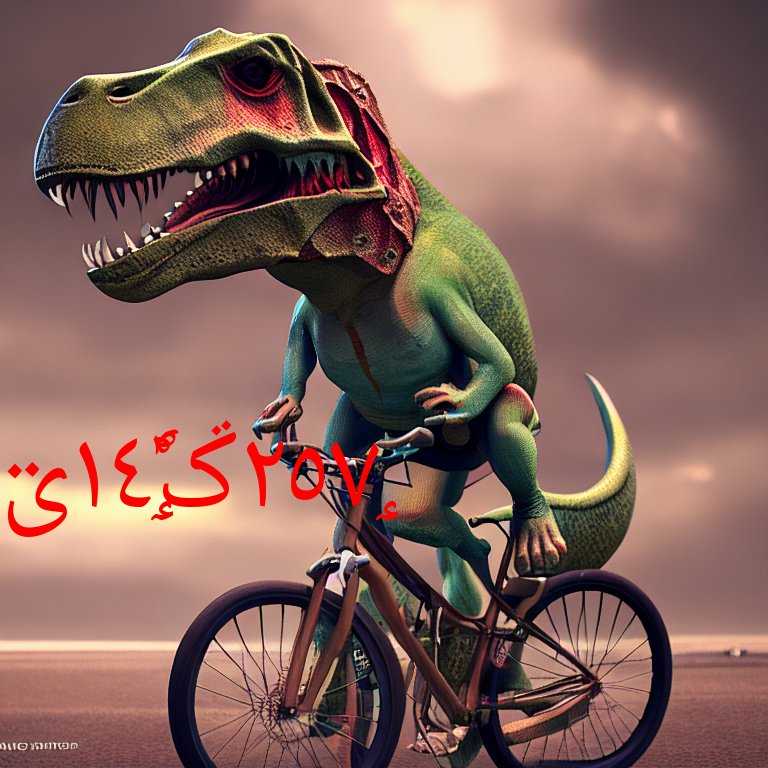}\end{minipage} &  
        \begin{minipage}{.16\linewidth}\includegraphics[width=\linewidth]{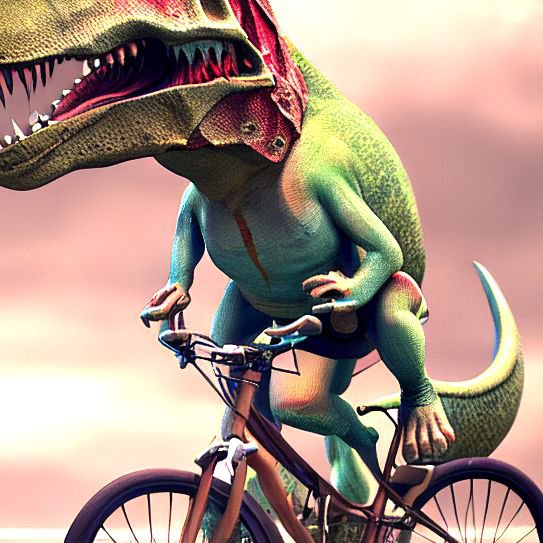}\end{minipage} 
        \\ \\
    \end{tabular}
    }
\caption{Illustration of all transformations evaluated in sections~\ref{sec:application} and \ref{sec:experiments}.}
\vspace{-0.5cm}
\label{fig:all-transformations}
\end{figure*}

\subsection{Image transformations}\label{app:transformations}
We evaluate the robustness of the watermark to a set of transformations in sections~\ref{sec:application}, \ref{sec:experiments} and \ref{sec:supp-robustness}.
They simulate image processing steps that are commonly used in image editing software.
We illustrate them in \autoref{fig:all-transformations}.
For crop and resize, the parameter is the ratio of the new area to the original area.
For rotation, the parameter is the angle in degrees.
For JPEG compression, the parameter is the quality factor (in general 90\% or higher is considered high quality, 80\%-90\% is medium, and 70\%-80\% is low).
For brightness, contrast, saturation, and sharpness, the parameter is the default factor used in the PIL and Torchvision~\cite{marcel2010torchvision} libraries.
The text overlay is made through the AugLy library~\cite{papakipos2022augly}, and adds a text at a random position in the image.
The combined transformation is a combination of a crop $0.5$, a brightness change $1.5$, and a JPEG $80$ compression.

\subsection{Generative tasks}\label{app:evaluation}

\paragraph*{Text-to-image.}
In text-to-image generation, the diffusion process is guided by a text prompt. 
We follow the standard protocol in the literature~\cite{ramesh2022hierarchical, ramesh2021zero, rombach2022high, saharia2022photorealistic} and evaluate the generation on prompts from the validation set of MS-COCO~\cite{lin2014microsoft}.
To do so, we first retrieve all the captions from the validation set, keep only the first one for each image, and select the first $1000$ or $5000$ captions depending on the evaluation protocol.
We use guidance scale $3.0$ and $50$ diffusion steps.
If not specified, the generation is done for $5000$ images.
The FID is computed over the validation set of MS-COCO, resized to $512\times 512$.

\vspace{-0.2cm}
\paragraph*{Image edition.}
DiffEdit~\cite{couairon2022diffedit} takes as input an image, a text describing the image and a novel description that the edited image should match. 
First, a mask is computed to identify which regions of the image should be edited. 
Then, mask-based generation is performed in the latent space, before converting the output back to RGB space with the image decoder. 
We use the default parameters used in the original paper, with an encoding ratio of 90\%, and compute a set of $5000$ images from the COCO dataset, edited with the same prompts as the paper~\cite{couairon2022diffedit}.
The FID is computed over the validation set of MS-COCO, resized to $512\times 512$.

\vspace{-0.2cm}
\paragraph*{Inpainting.}
We follow the protocol of LaMa~\cite{suvorov2022resolution}, and generate $5000$ masks with the ``thick'' setting, at resolution $512\times 512$, each mask covering $1-50\%$ of the initial image (with an average of $27\%$).
For the diffusion-based inpainting, we use the inference-time algorithm presented in \cite{song2020score}, also used in Glide~\cite{nichol2021glide}, which corrects intermediate estimations of the final generated image with the ground truth pixel values outside the inpainting mask. 
For latent diffusion models, the same algorithm can be applied in latent space, by encoding the image to be inpainted and downsampling the inpainting mask. 
In this case, we consider $2$ different variations: (1) inpainting is performed in the latent space and the final image is obtained by simply decoding the latent image; and (2) the same procedure is applied, but after decoding, ground truth pixel values from outside the inpainting mask are copy-pasted from the original image. 
The latter allows to keep the rest of the image perfectly identical to the original one, at the cost of introducing copy-paste artifacts, visible in the borders. 
Image quality is measured with an FID score, computed over the validation set of ImageNet~\cite{deng2009imagenet}, resized to $512\times 512$.

\vspace{-0.4cm}
\paragraph*{Super-resolution.}
We follow the protocol suggested by Saharia~\etal~\cite{saharia2022image}.
We first resize $5000$ random images from the validation set of ImageNet to $128\times 128$ using bicubic interpolation, and upscale them to $512\times 512$.
The FID is computed over the validation set of ImageNet, cropped and resized to $512\times 512$.

\begin{figure*}[b]
    \centering
    \scriptsize
    \newcommand{\imwidth}{0.165\textwidth}
    \setlength{\tabcolsep}{0pt}
    \begin{tabular}{cc@{\hskip 2pt}cc@{\hskip 2pt}cc}
        \toprule
        $\lambda_i = 0.025$ &  & $\lambda_i = 0.05$ &  & $\lambda_i = 0.1$ &  \\
        \midrule
        \includegraphics[width=\imwidth]{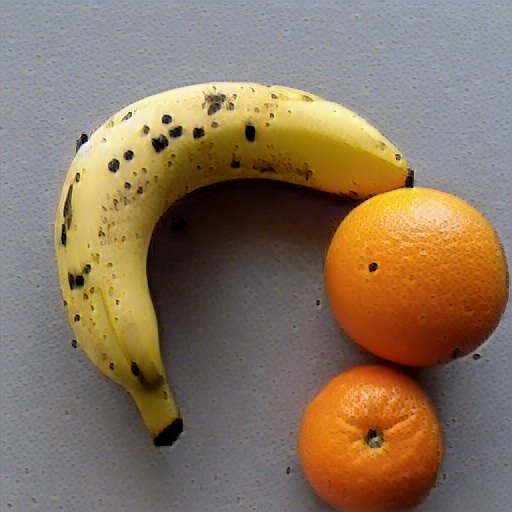} &
        \includegraphics[width=\imwidth]{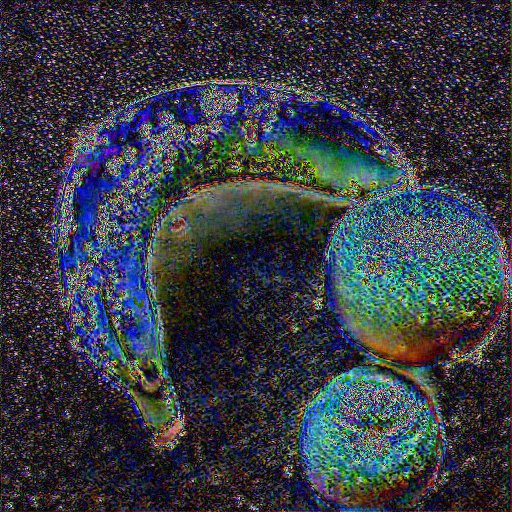} &
        \includegraphics[width=\imwidth]{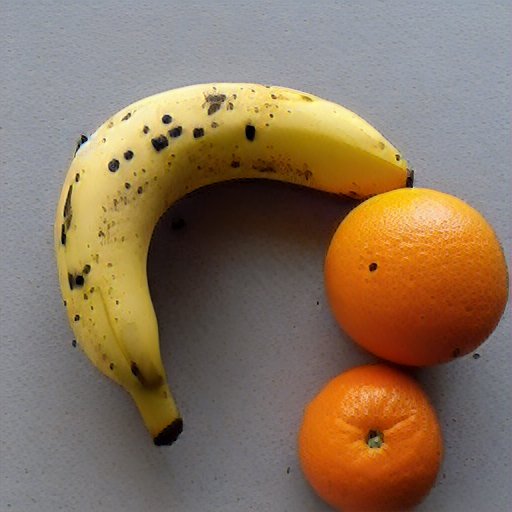} &
        \includegraphics[width=\imwidth]{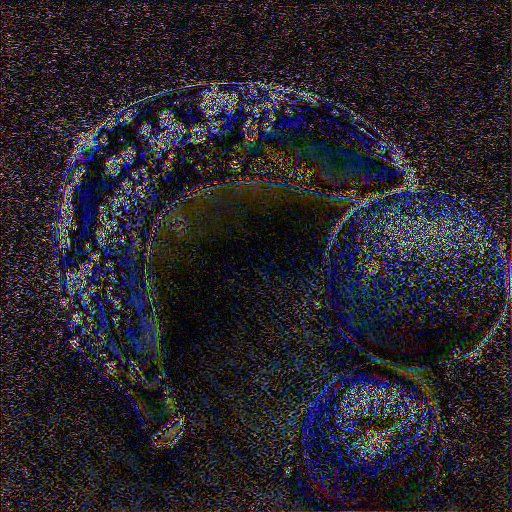} &
        \includegraphics[width=\imwidth]{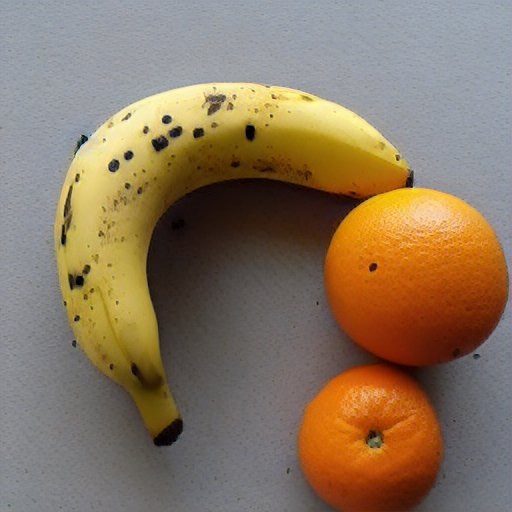} &
        \includegraphics[width=\imwidth]{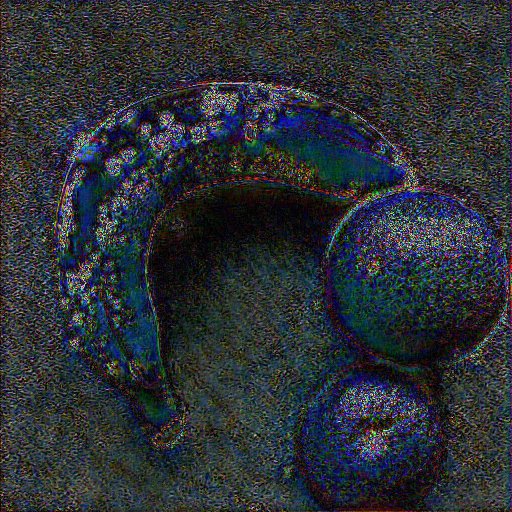} \\
    \end{tabular}
    \begin{tabular}{c@{\hskip 2pt}ccccc}
        \toprule
        Original & Watson-VGG & Watson-DFT & LPIPS & MSE & LPIPS + $0.1\cdot$MSE \\
        \midrule
        \includegraphics[width=\imwidth]{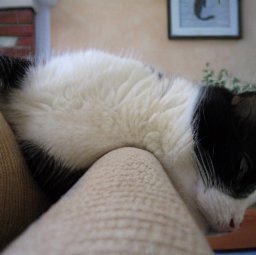} &
        \includegraphics[width=\imwidth]{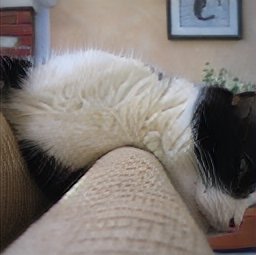} &
        \includegraphics[width=\imwidth]{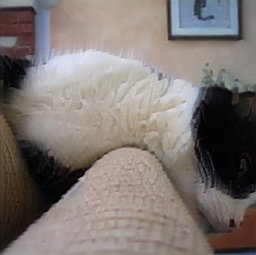} &
        \includegraphics[width=\imwidth]{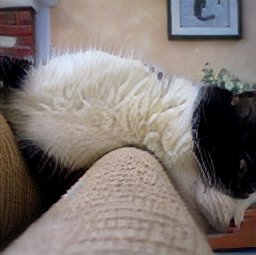} &
        \includegraphics[width=\imwidth]{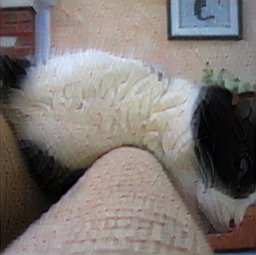} &
        \includegraphics[width=\imwidth]{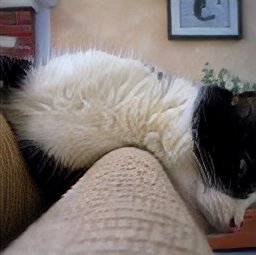} \\
        \bottomrule \\
    \end{tabular}
    \caption{
        Qualitative influence of the perceptual loss during LDM fine-tuning. 
        (Top): we show images generated with the LDM auto-encoder fine-tuned with different $\lambda_i$, and the pixel-wise difference ($\times 10$) with regards to the image obtained with the original model.
        PSNR are $24$dB, $26$dB, $28$dB from left to right. 
        (Bottom): we change the perceptual loss and fix $\lambda_i$ to have approximately the same bit accuracy of $0.95$ on the ``combined'' augmentation. 
    }
    \label{fig:supp-lossi}
\end{figure*}

\subsection{Watermarking methods}\label{app:watermarking}
For Dct-Dwt, we use the implementation of \href{https://github.com/ShieldMnt/invisible-watermark}{https://github.com/ShieldMnt/invisible-watermark} (the one used in Stable Diffusion).
For SSL Watermark~\cite{fernandez2022sslwatermarking} and FNNS~\cite{kishore2021fixed} the watermark is embedded by optimizing the image, such that the output of a pre-trained model is close to the given key (like in adversarial examples~\cite{goodfellow2014adversarial}).
The difference between the two is that in SSL Watermark we use a model pre-trained with DINO~\cite{caron2021dino}, while FNNS uses a watermark or stenography model.
For SSL Watermark we use the default pre-trained model of the original paper. 
For FNNS we use the HiDDeN extractor used in all our experiments, and not SteganoGan~\cite{zhang2019steganogan} as in the original paper, because we want to extract watermarks from images of different sizes.
We use the image optimization scheme of Active Indexing~\cite{fernandez2022active}, \ie we optimize the distortion image for $10$ iterations, and modulate it with a perceptual just noticeable difference (JND) mask.
This avoids visible artifacts and gives a PSNR comparable with our method ($\approx 30$dB).
For HiDDeN, we use the watermark encoder and extractor from our pre-training phase, but the extractor is not whitened and we modulate the encoder output with the same JND mask.
Note that in all cases we watermark images one by one for simplicity. 
In practice the watermarking could be done by batch, which would be more efficient.

\subsection{Attacks}\label{app:attacks}

\paragraph*{Watermark removal.} 
The perceptual auto-encoders aim to create compressed latent representations of images.
We select $2$ state-of-the-art auto-encoders from the CompressAI library zoo~\cite{begaint2020compressai}: the factorized prior model~\cite{balle2018variational} and the anchor model variant~\cite{cheng2020learned}.
We also select the auto-encoders from Esser~\etal~\cite{esser2021taming} and Rombach~\etal~\cite{rombach2022ldm}.
For all models, we use different compression factors to observe the trade-off between quality degradation and removal robustness.
For \texttt{bmshj2018}: $1$, $4$ and $8$, for \texttt{cheng2020}: $1$, $3$ and $6$, for \texttt{esser2021}: VQ-$4$, $8$ and $16$, for \texttt{rombach2022} KL-$4$, $8$, $16$ and $32$ (KL-$8$ being the one used by SD v1.4).
We generate $1$k images from text prompts with our LDM watermarked with a $48$-bits key.
We then try to remove the watermark using the auto-encoders, and compute the bit accuracy on the extracted watermark.
The PSNR is computed between the original image and the reconstructed one, which explains why the PSNR does not exceed $30$dB (since the watermarked image already has a PNSR of $30$dB).
If we compared between the watermarked image and the image reconstructed by the auto-encoder instead, the curves would show the same trend but the PSNR would be $2$-$3$ points higher.

\paragraph*{Watermark removal (white-box).} 
In the white-box case, we assume have access to the extractor model.
The adversarial attack is performed by optimizing the image in the same manner as~\cite{fernandez2022sslwatermarking}.
The objective is a MSE loss between the output of the extractor and a random binary message fixed beforehand. 
The attack is performed for $10$ iterations with the Adam optimizer~\cite{kingma2014adam} with learning rate $0.1$.

\paragraph*{Watermark removal (network-level).} 
We use the same fine-tuning procedure as in Sec.~\ref{subsec:finetuning}.
This is done for different numbers of steps, namely $100$, $200$, and every multiple of $200$ up to $1600$.
The bit accuracy and the reported PSNR are computed on $1$k images of the validation set of COCO, for the auto-encoding task.

\paragraph*{Model collusion.} 
The goal is to observe the decoded watermarks on the generation when $2$ models are averaged together.
We fine-tune the LDM decoder for $10$ different $48$-bits keys (representing $10$ Bobs).
We then randomly sample a pair of Bobs and average the $2$ models, with which we generate $100$ images.
We then extract the watermark from the generated images and compare them to the $2$ original keys.
We repeat this experiment $10$ times, meaning that we observe $10\times 100 \times 48=48000$ decoded bits.

In the inline figure, the rightmost skewed normal is fitted with the Scipy library and the corresponding parameters are $a:6.96, e:0.06, w:0.38$. 
This done over all bits where Bobs both have a $1$.
The same observation holds when there is no collusion, with approximately the same parameters.
When the bit is not the same between Bobs, we denote by $m_1^{(i)}$ the random variable representing the output of the extractor in the case where the generative model only comes from Bob$^{(i)}$,
and by $m_2$ the random variable representing the output of the extractor in the case where the generative model comes from the average of the two Bobs.
Then in our model $m_2 = 0.5 \cdot ( m_1^{(i)} + m_1^{(j)})$, and the pdf of $m_2$ is the convolution of the pdf of $m_1^{(i)}$ and the pdf of $m_1^{(j)}$, rescaled in the x axis because of the factor $0.5$.

\newcommand{\rota}[1]{\rotatebox{0}{\footnotesize\hspace{-0cm}#1}}
\begin{table*}[t]
\centering
\caption{
    Watermark robustness on different tasks and image transformations applied before decoding.
    We report the bit accuracy, averaged over $10\times1$k images generated with $10$ different keys.
    The combined transformation is a combination Crop $50\%$, Brightness $1.5$ and JPEG $80$.
    More detail on the evaluation is available in the supplement~\ref{app:evaluation}. \\
    }\label{tab:supp-robustness}
\resizebox{1.0\linewidth}{!}{
    \footnotesize
    \begin{tabular}{ *{2}{l} *{11}{p{1.0cm}} }
    \toprule
    \multirow{2}{*}{Task}                       & \multirow{2}{*}{}       & \multicolumn{9}{c}{Image transformation} \\ \cmidrule{3-12}
                                                &                       &  \rota{None} & \rota{Crop $0.1$} & \rota{JPEG $50$} & \rota{Resi. $0.7$} & \rota{Bright. $2.0$} & \rota{Cont. $2.0$} & \rota{Sat. $2.0$} & \rota{Sharp. $2.0$} & \rota{Text over.} & \rota{Comb.}\\ \midrule
     Text-to-Image          & LDM~\cite{rombach2022ldm}                         & $0.99$ & $0.95$ & $0.88$ & $0.91$ & $0.97$ & $0.98$ & $0.99$ & $0.99$ & $0.99$ & $0.92$ \\ \midrule
     Image Edition          & DiffEdit~\cite{couairon2022diffedit} & $0.99$ & $0.95$ & $0.90$ & $0.91$ & $0.98$ & $0.98$ & $0.99$ & $0.99$ & $0.99$ & $0.94$\\ \midrule
     Inpainting - Full      & \multirow{2}{*}{Glide~\cite{nichol2021glide}}     & $0.99$ & $0.97$ & $0.88$ & $0.90$ & $0.98$ & $0.99$ & $0.99$ & $1.00$ & $0.99$ & $0.93$ \\ 
    {\color{white}Inpa} - Mask only   &                                         & $0.89$ & $0.76$ & $0.73$ & $0.77$ & $0.84$ & $0.86$ & $0.89$ & $0.91$ & $0.89$ & $0.78$ \\  \midrule
     Super-Resolution & LDM~\cite{rombach2022ldm} & $0.98$ & $0.93$ & $0.86$ & $0.85$ & $0.96$ & $0.96$ & $0.97$ & $0.98$ & $0.98$ & $0.92$\\ 
    \bottomrule
    \end{tabular}
}
\end{table*}

\begin{figure*}[b]
    \begin{minipage}{0.67\textwidth}
        \centering
        \scriptsize
        \newcommand{\imheight}{0.33\textwidth}
        \setlength{\tabcolsep}{0pt}
        \begin{tabular}{lll}
            Before whitening: & After whitening: & Bernoulli simulation: \\
            \includegraphics[height=\imheight, trim=2cm 0.5cm 3.7cm 1cm, clip]{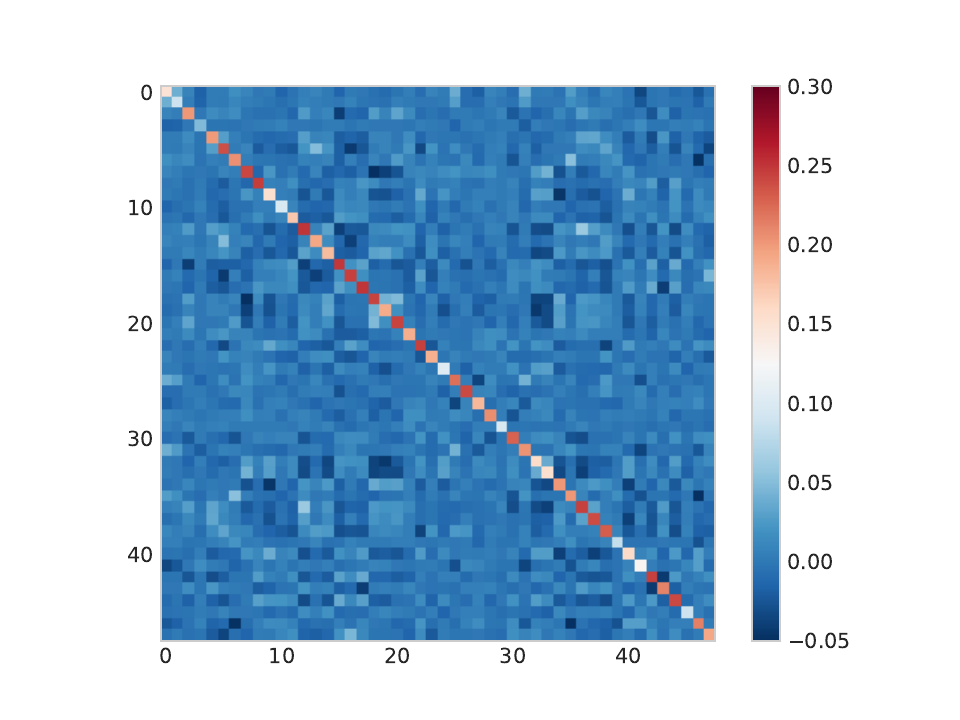} &
            \includegraphics[height=\imheight, trim=2cm 0.5cm 3.7cm 1cm, clip]{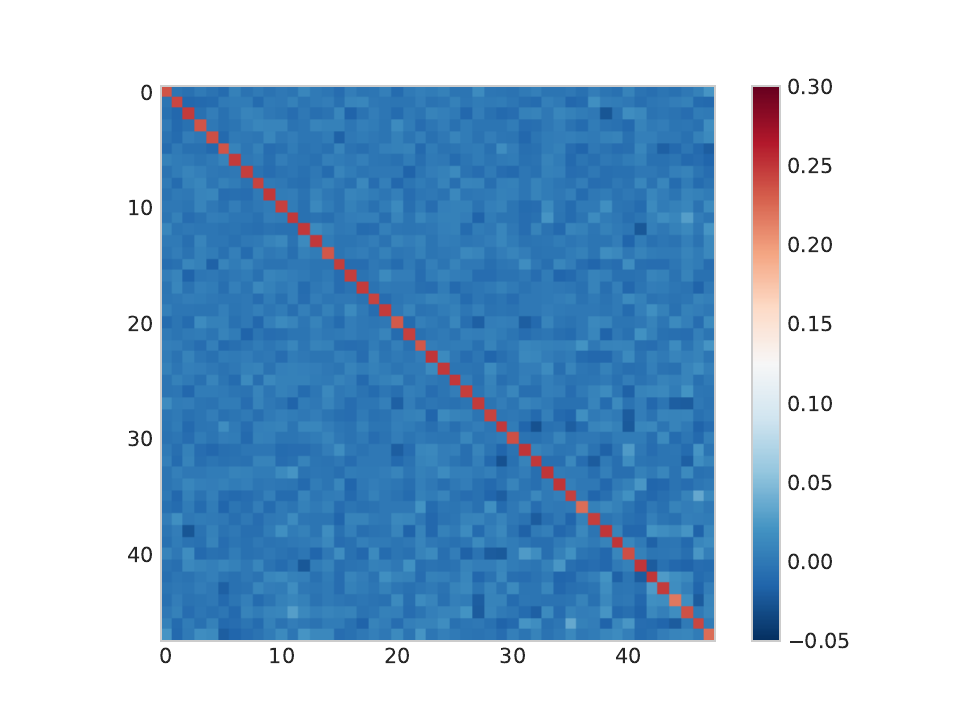} &
            \includegraphics[height=\imheight, trim=2cm 0.5cm 1.5cm 1cm, clip]{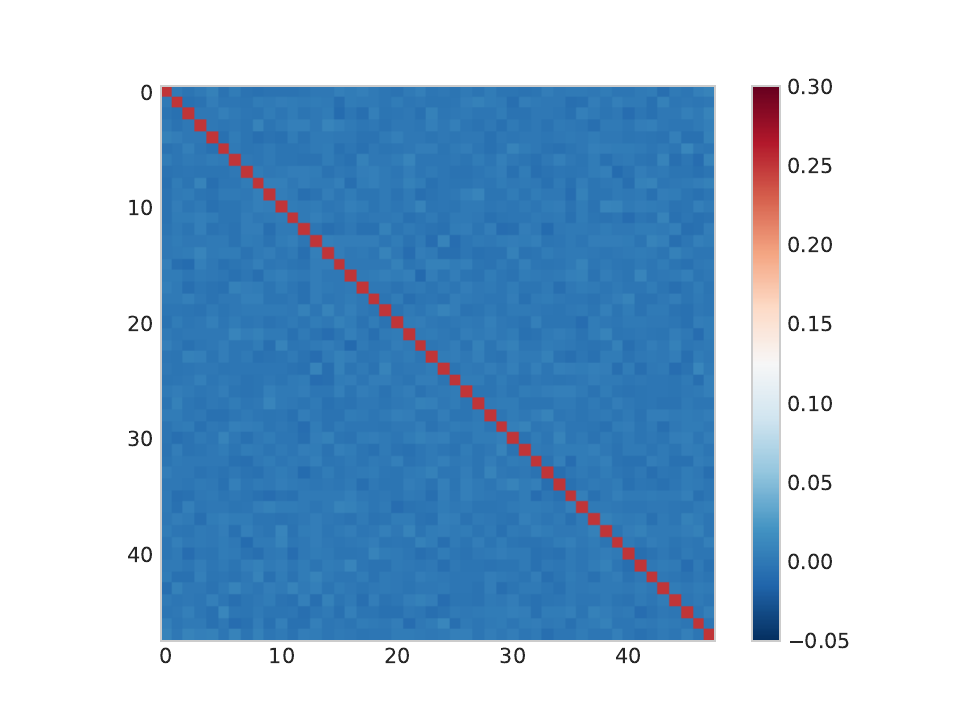} \\
        \end{tabular}
        \caption{Covariance matrices of the bits output by the watermark decoder $\mathcal{W}$ before and after whitening.}
        \label{fig:supp-assumption}
    \end{minipage}\hfill
    \begin{minipage}{0.27\textwidth}
        \centering
        \includegraphics[width=0.99\textwidth, trim=0 0 0 0, clip]{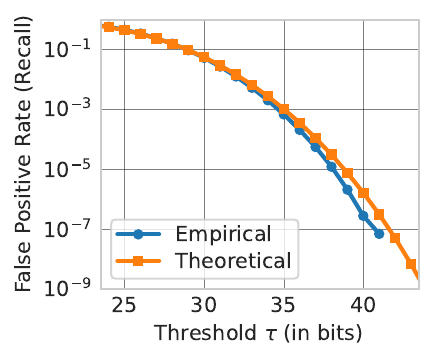}
        \caption{FPR Empirical check.}
        \label{fig:supp-fpr-check}
    \end{minipage}
\end{figure*}

\section{Additional Experiments}

\subsection{Perceptual loss}\label{sec:supp-percep-loss}
The perceptual loss of \eqref{eq:loss2} affects the image quality.
\autoref{fig:supp-lossi} shows how the parameter $\lambda_i$ affects the image quality.
For high values, the image quality is very good. %
For low values, artifacts mainly appear in textured area of the image. 
It is interesting to note that this begins to be problematic only for low PSNR values (around 25 dB).

\autoref{fig:supp-lossi} shows an example of a watermarked image for different perceptual losses: Watson-VGG~\cite{czolbe2020loss}, Watson-DFT~\cite{czolbe2020loss}, LPIPS~\cite{zhang2018unreasonable}, MSE, and LPIPS+MSE.
We set the weight $\lambda_i$ of the perceptual loss so that the watermark performance is approximately the same for all types of loss, and such that the degradation of the image quality is strong enough to be seen.
Overall, we observe that the Watson-VGG loss gave the most eye-pleasing results, closely followed by the LPIPS.
When using the MSE, images are blurry and artifacts appear more easily, even though the PSNR is higher.

\subsection{Additional results on watermarks robustness}\label{sec:supp-robustness}

In \autoref{tab:supp-robustness}, we report the same table as in \autoref{tab:quality-watermarking} that evaluates the watermark robustness in bit accuracy on different tasks, with additional image transformations.
They are detailed and illustrated in App.~\ref{app:evaluation}.
As a reminder, the watermark is a $48$-bit binary key.
It is robust to a wide range transformations, and most often yields above $0.9$ bit accuracy.
The resize and JPEG $50$ transformations seems to be the most challenging ones, and sometimes get bellow $0.9$.
Note that the crop location is not important but the visual content of the crop is, \eg there is no way to decode the watermark on crops of blue sky (this is the reason we only show center crop).

\subsection{Additional network level attacks}\label{sec:supp-network-level}

Tab.~\ref{tab:supp-network-level} reports robustness of the watermarks to different quantization and pruning levels for the LDM decoder. 
Quantization is performed naively, by rounding the weights to the closest quantized value in the min-max range of every weight matrix.
Pruning is done using PyTorch~\cite{paszke2019pytorch} pruning API, with the L1 norm as criterion.
We observe that the network generation quality degrades faster than WM robustness. 
To reduce bit accuracy lower than 98\%, quantization degrades the PSNR $<$25dB, and pruning $<$20dB.

\begin{table}[t]
    \caption{
        Bit accuracy after network attacks, observed over 10$\times$1k images generated from text prompts.
    }\label{tab:supp-network-level}
    \centering
    \setlength{\tabcolsep}{6pt}
    \vspace*{0.2cm}
    \resizebox{0.95\linewidth}{!}{
        \begin{tabular}{ll|ll}
            \toprule
            Quantization (8-bits) & 0.99 & Pruning L1 ($30\%$) & 0.99 \\
            Quantization (4-bits)  & 0.99 & Pruning L1 ($60\%$) & 0.95 \\
            \bottomrule
        \end{tabular}
    }
    \vspace*{-0.3cm}
\end{table}

\subsection{Scaling factor at pre-training.} 
The watermark encoder does not need to be perceptually good and it is beneficial to degrade image quality during pre-training.
In the following, ablations are conducted on a shorter schedule of $50$ epochs, on $128\times 128$ images and $16$-bits messages.
In \autoref{tab:encoder-quality}, we train watermark encoders/extractors for different scaling factor $\alpha$ (see Sec.~\ref{subsec:pre-training}), and observe that $\alpha$ strongly affects the bit accuracy of the method.
When it is too high, the LDM needs to generate low quality images for the same performance because the distortions seen at pre-training by the extractor are too strong.
When it is too low, they are not strong enough for the watermarks to be robust: the LDM will learn how to generate watermarked images, but the extractor won't be able to extract them on edited images.

\begin{table}[t]
    \centering
    \caption{
        Influence of the (discarded) watermark encoder perceptual quality. P$_{1,2}$ stands for Phase 1 or 2.
    }\label{tab:encoder-quality}\vspace{0.2cm}
    \resizebox{0.95\linewidth}{!}{
    \begin{tabular}{l *{5}{l}}
        \toprule
        Scaling factor $\alpha$ & $0.8$ & $0.4$ & $0.2$ & $0.1$ & $0.05$ \\ 
        \midrule
        (P$_{1}$) - PSNR $\uparrow$  & $16.1$ & $21.8$ & $27.2$ & $33.5$ & $39.3$ \\
        (P$_{2}$) - PSNR $\uparrow$  & $27.9$ & $30.5$ & \textbf{30.8} & $28.8$ & $27.8$ \\
        \midrule
        (P$_{1}$) - Bit acc. $\uparrow$ on `none'& $1.00$ & $1.00$ & $0.86$ & $0.72$ & $0.62$ \\
        (P$_{2}$) - Bit acc. $\uparrow$ on `none'& $0.98$ & \textbf{0.98} & $0.91$ & $0.90$ & $0.96$ \\
        (P$_{2}$) - Bit acc.  $\uparrow$ on `comb.'& \textbf{0.86} & $0.73$ & $0.82$ & $0.81$ & $0.69$ \\
        \bottomrule 
    \end{tabular}
    }
\end{table}

\subsection{Are the decoded bits i.i.d. Bernoulli random variables?}
\label{app:assumption}
The FPR and the $p$-value \eqref{eq:p-value} are computed with the assumption that, for vanilla images (not watermarked),
the bits output by the watermark decoder $\mathcal{W}$ are independent and identically distributed (i.i.d.) Bernoulli random variables with parameter $0.5$.
This assumption is not true in practice, even when we tried using regularizing losses in the training at phase one~\cite{bardes2022vicreg, sablayrolles2018catalyser}.
This is why we whiten the output at the end of the pre-training.

\autoref{fig:supp-assumption} shows the covariance matrix of the hard bits output by $\mathcal{W}$ before and after whitening. 
They are computed over $5$k vanilla images, generated with our LDM at resolution $512\times512$ (as a reminder the whitening is performed on $1$k vanilla images from COCO at $256\times256$).
We compare them to the covariance matrix of a Bernoulli simulation, where we simulate $5k$ random messages of $48$ Bernoulli variables.
We observe the strong influence of the whitening on the covariance matrix, although it still differs a little from the Bernoulli simulation.
We also compute the bit-wise mean and observe that for un-whitened output bits, some bits are very biased.  
For instance, before whitening, one bit had an average value of $0.95$ (meaning that it almost always outputs $1$).
After whitening, the maximum average value of a bit is $0.68$.
For the sake of comparison, the maximum average value of a bit in the Bernoulli simulation was $0.52$.
It seems to indicate that the distribution of the generated images are different than the one of vanilla images, and that it impacts the output bits.
Therefore, the bits are not perfectly i.i.d. Bernoulli random variables. 
We however found they are close enough for the theoretical FPR computation to match the empirical one (see next section) -- which was what we wanted to achieve.

\subsection{Empirical check of the FPR}\label{app:fpr-check}
In \autoref{fig:tpr-fpr}, we plotted the TPR against a theoretical value for the FPR, with the i.i.d. Bernoulli assumption.
The FPR was computed theoretically with \eqref{eq:p-value}.
Here, we empirically check on smaller values of the FPR (up to $10^{-7}$) that the empirical FPR matches the theoretical one (higher values would be too computationally costly).
To do so, we use the $1.4$ million vanilla images from the training set of ImageNet resized and cropped to $512\times512$, and perform the watermark extraction with $\mathcal{W}$.
We then fix $10$ random $48$-bits key $m^{(1)},\cdots, m^{(10)}$, and, for each image, we compute the number of matching bits $d(m', m^{(i)})$ between the extracted message $m'$ and the key $m^{(i)}$, and flag the image if $d(m', m^{(i)})\geq \tau$.

\autoref{fig:supp-fpr-check} plots the FPR averaged over the $10$ keys, as a function of the threshold $\tau$.
We compare it to the theoretical one obtained with \eqref{eq:p-value}.
As it can be seen, they match almost perfectly for high FPR values. 
For lower ones ($<10^{-6}$), the theoretical FPR is slightly higher than the empirical one.
This is a good thing since it means that if we fixed the FPR at a certain value, we would observe a lower one in practice.

\section{Additional Qualitative Results}\label{app:qualitative}

\begin{figure*}
\centering
    \scriptsize
    \newcommand{\imwidth}{0.165\textwidth}
        \setlength{\tabcolsep}{0pt}
        \begin{tabular}{c@{\hskip 2pt}ccccc}
        \toprule
        Original & Stable Signature & Dct-Dwt & SSL Watermark & FNNS & HiDDeN \\
        \midrule
        \includegraphics[width=\imwidth]{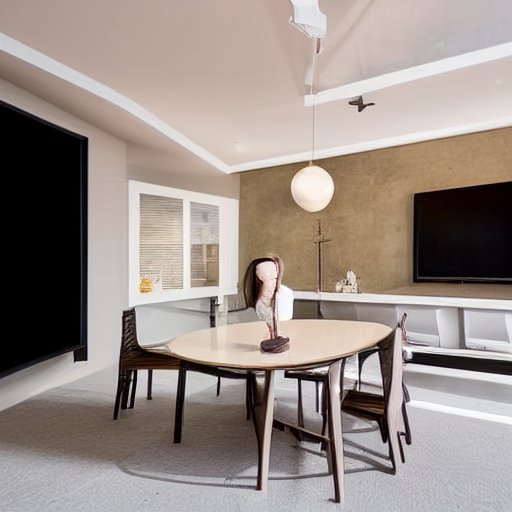} &
        \includegraphics[width=\imwidth]{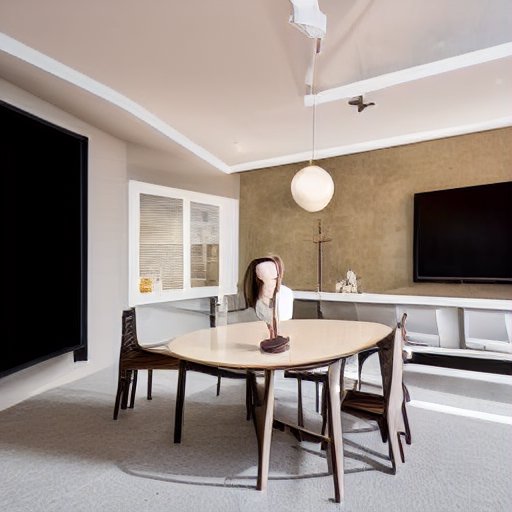} &
        \includegraphics[width=\imwidth]{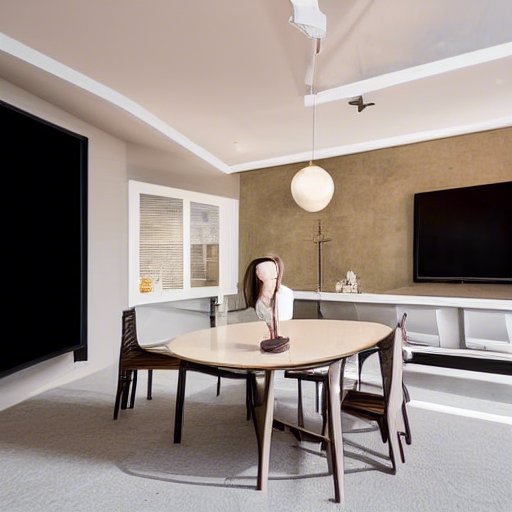} &
        \includegraphics[width=\imwidth]{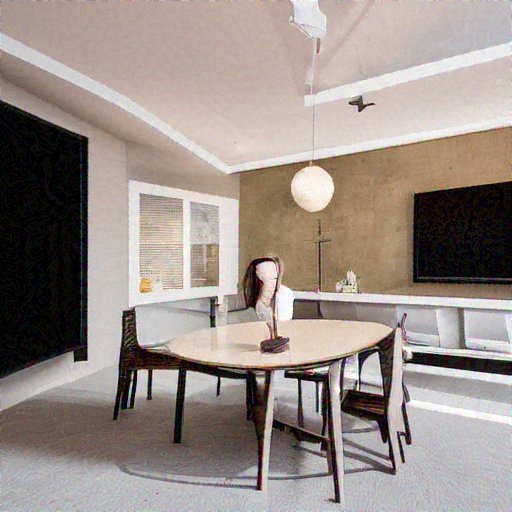} &
        \includegraphics[width=\imwidth]{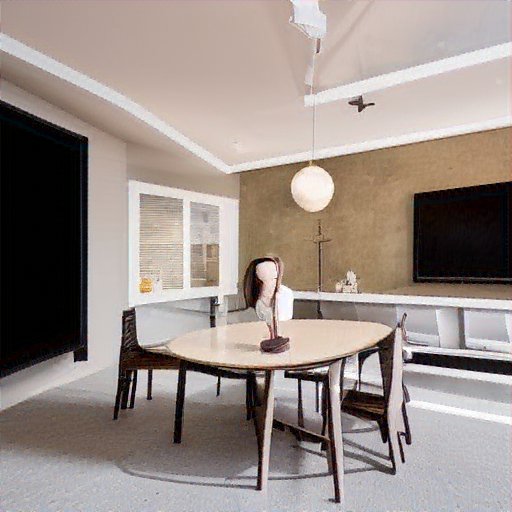} &
        \includegraphics[width=\imwidth]{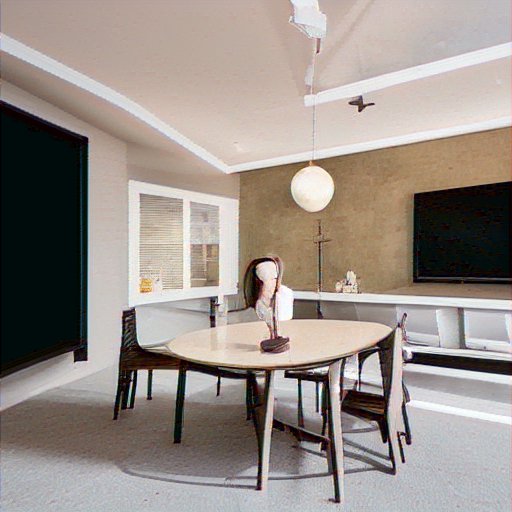} \\

         &
        \includegraphics[width=\imwidth]{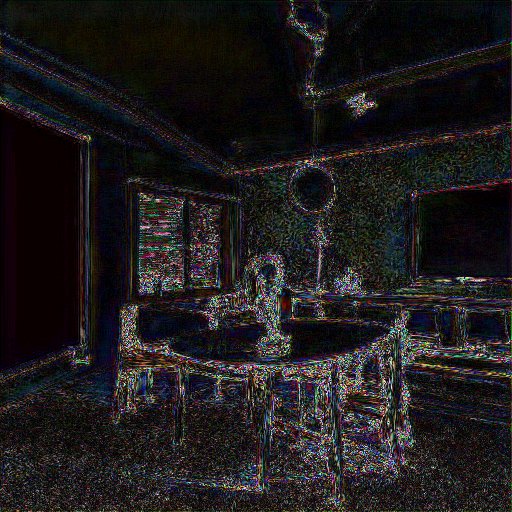} &
        \includegraphics[width=\imwidth]{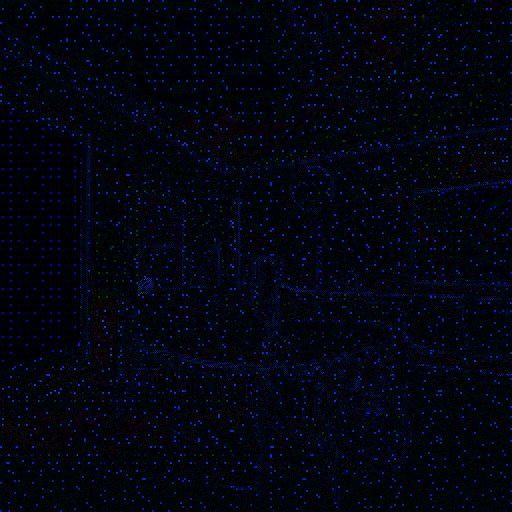} &
        \includegraphics[width=\imwidth]{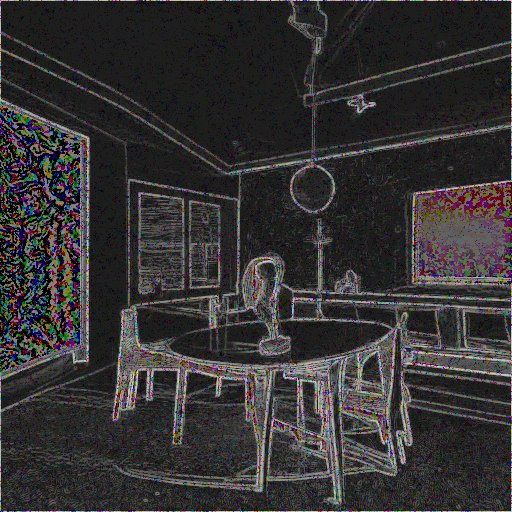} &
        \includegraphics[width=\imwidth]{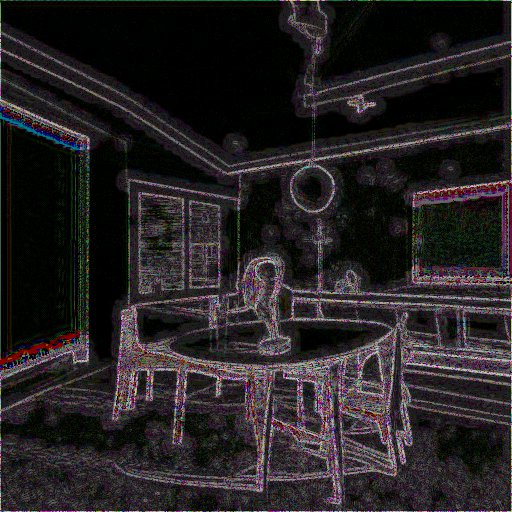} &
        \includegraphics[width=\imwidth]{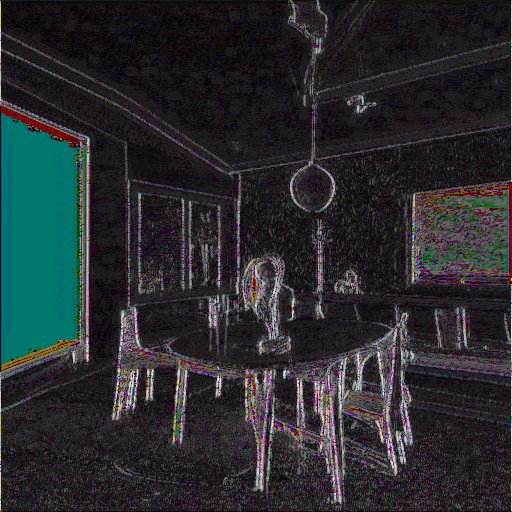} \\
        \rule{0pt}{6ex}%

        \includegraphics[width=\imwidth]{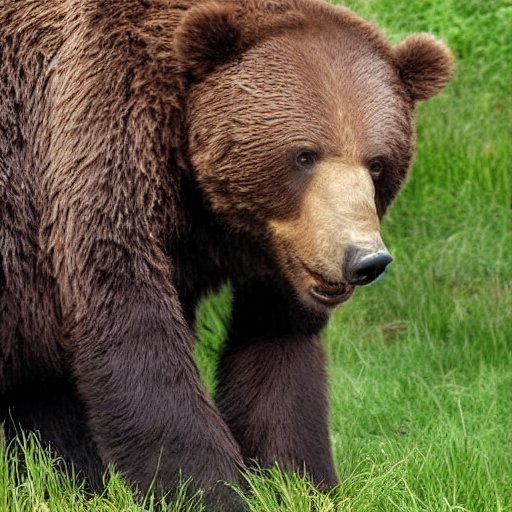} &
        \includegraphics[width=\imwidth]{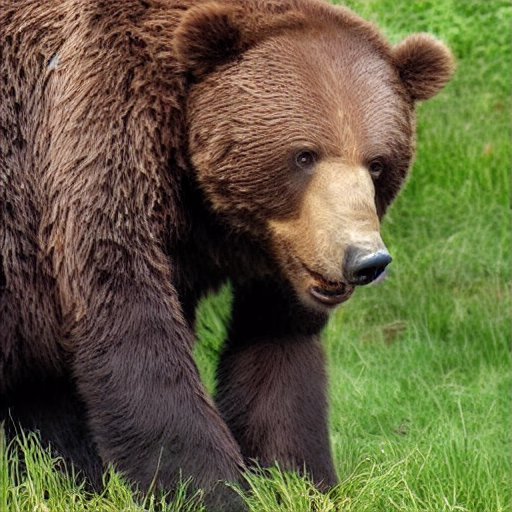} &
        \includegraphics[width=\imwidth]{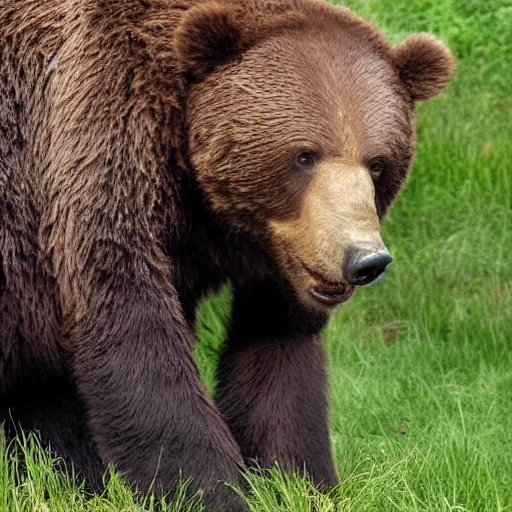} &
        \includegraphics[width=\imwidth]{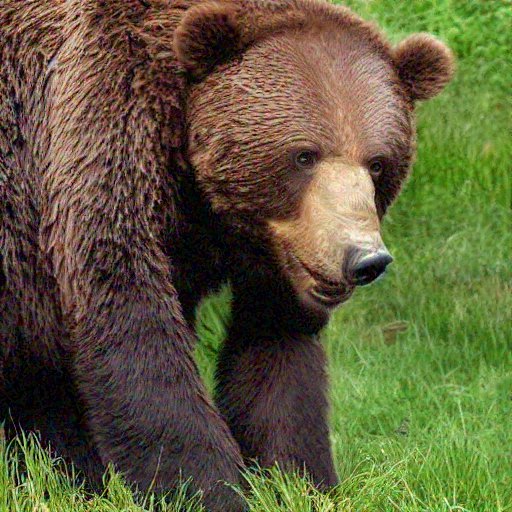} &
        \includegraphics[width=\imwidth]{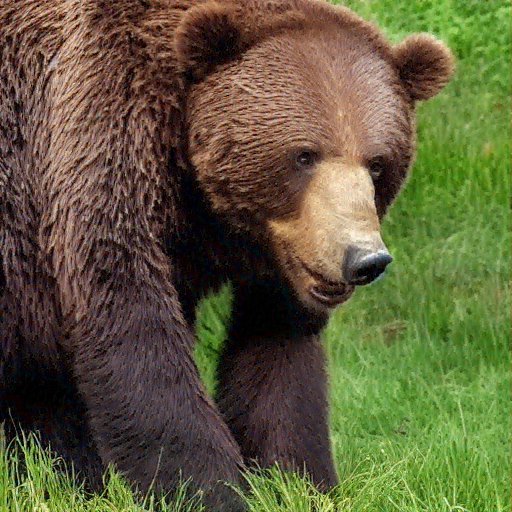} &
        \includegraphics[width=\imwidth]{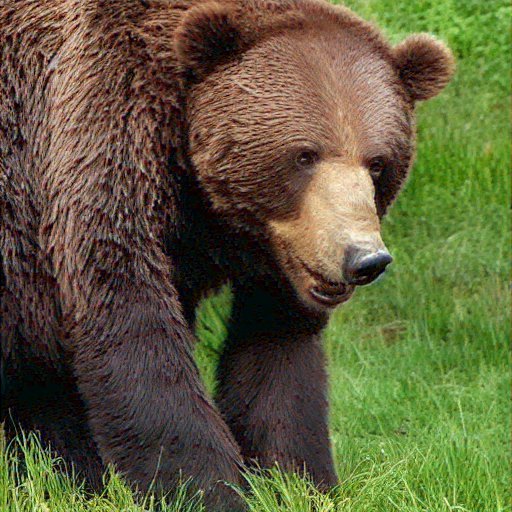} \\

         &
        \includegraphics[width=\imwidth]{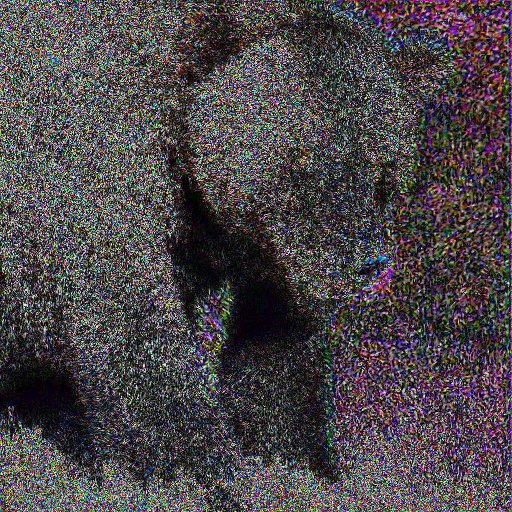} &
        \includegraphics[width=\imwidth]{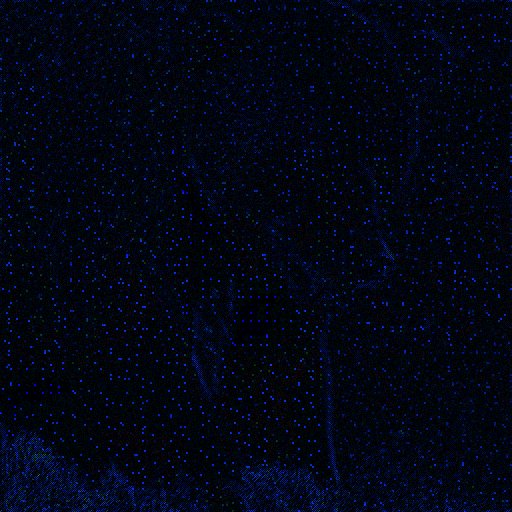} &
        \includegraphics[width=\imwidth]{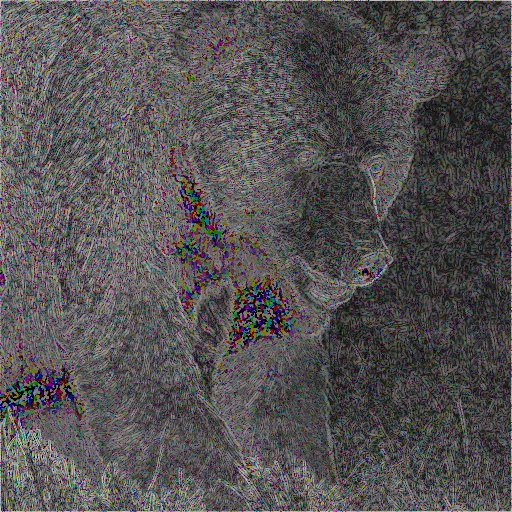} &
        \includegraphics[width=\imwidth]{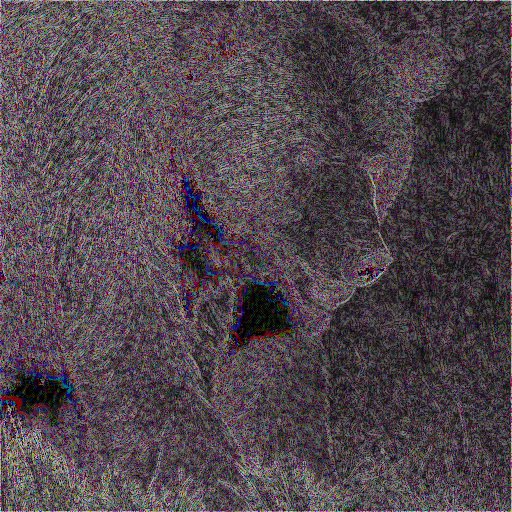} &
        \includegraphics[width=\imwidth]{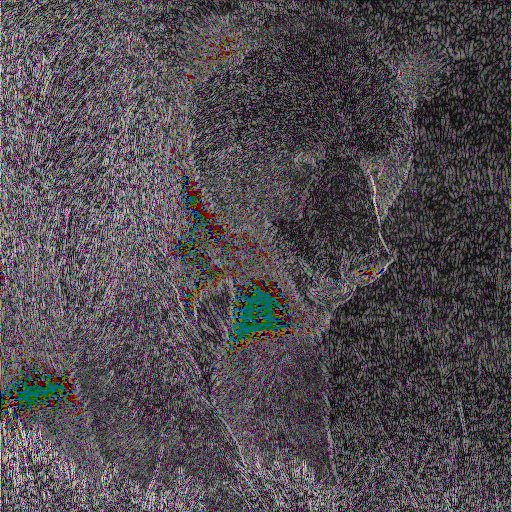} \\
    \\
    \end{tabular}
\caption{\label{fig:supp-watermark} Qualitative results for different watermarking methods on generated images at resolution $512$.}
\end{figure*}

\begin{figure*}
\centering
    \scriptsize
    \newcommand{\imwidth}{0.165\textwidth}
        \setlength{\tabcolsep}{0pt}
        \begin{tabular}{ccc@{\hskip 2pt}ccc}
        \toprule
        Original & Watermarked & Difference & Original & Watermarked & Difference \\
        \midrule
        \includegraphics[width=\imwidth]{figs/supp/txt2img/00_nw.jpg} &
       \includegraphics[width=\imwidth]{figs/supp/txt2img/00_w.jpg} &
       \includegraphics[width=\imwidth]{figs/supp/txt2img/00_diff.jpg} &
       \includegraphics[width=\imwidth]{figs/supp/txt2img/01_nw.jpg} &
       \includegraphics[width=\imwidth]{figs/supp/txt2img/01_w.jpg} &
       \includegraphics[width=\imwidth]{figs/supp/txt2img/01_diff.jpg} \\
       \rule{0pt}{6ex}%

       \includegraphics[width=\imwidth]{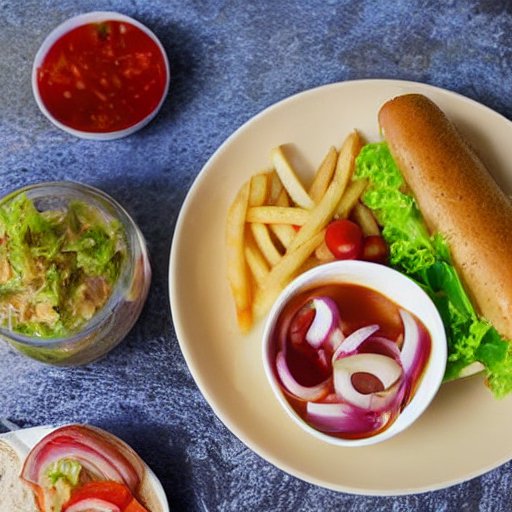} &
       \includegraphics[width=\imwidth]{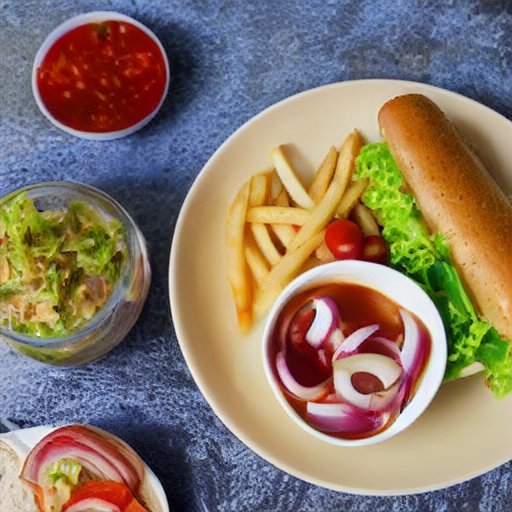} &
       \includegraphics[width=\imwidth]{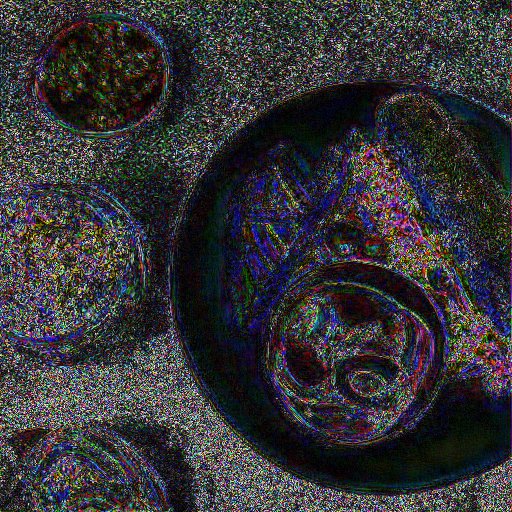} &
       \includegraphics[width=\imwidth]{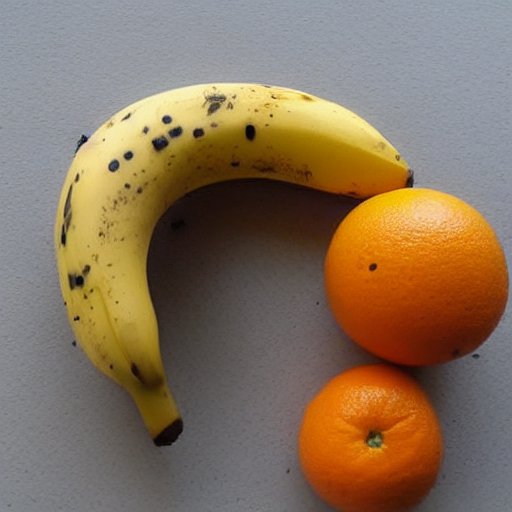} &
       \includegraphics[width=\imwidth]{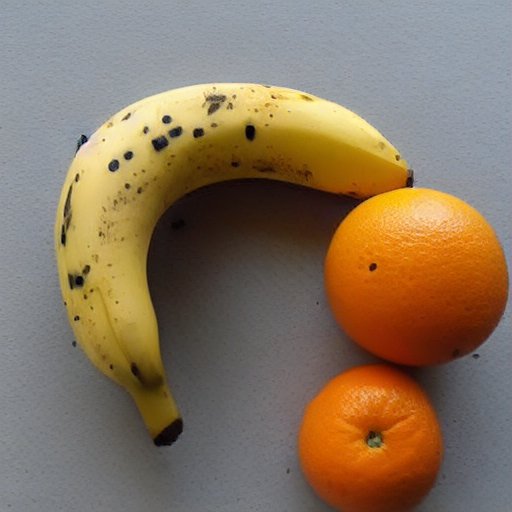} &
       \includegraphics[width=\imwidth]{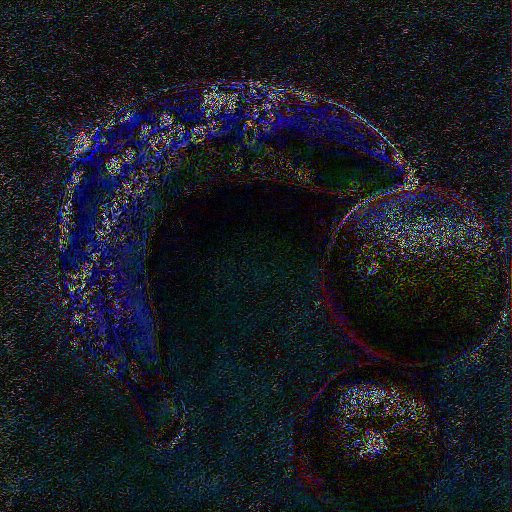} \\
       \rule{0pt}{6ex}%

       \includegraphics[width=\imwidth]{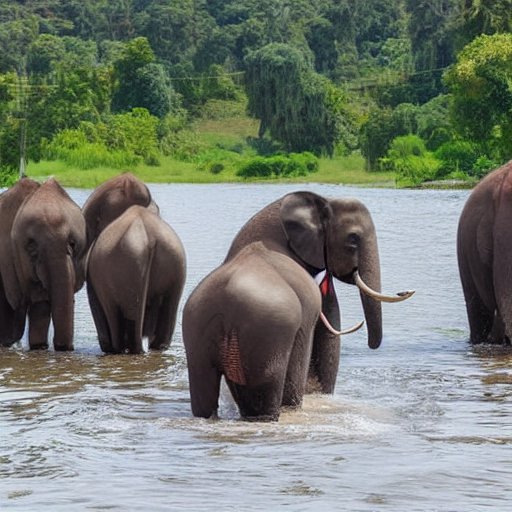} &
       \includegraphics[width=\imwidth]{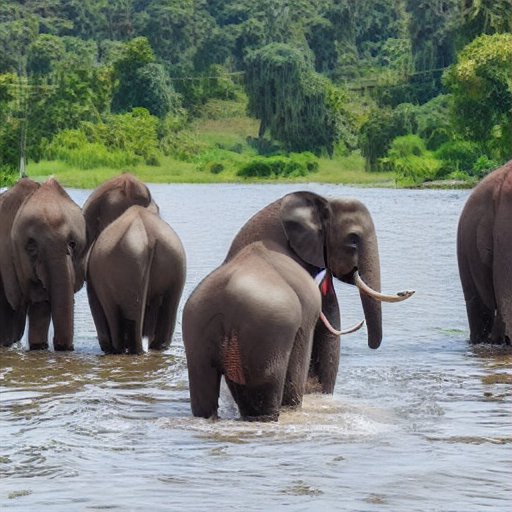} &
       \includegraphics[width=\imwidth]{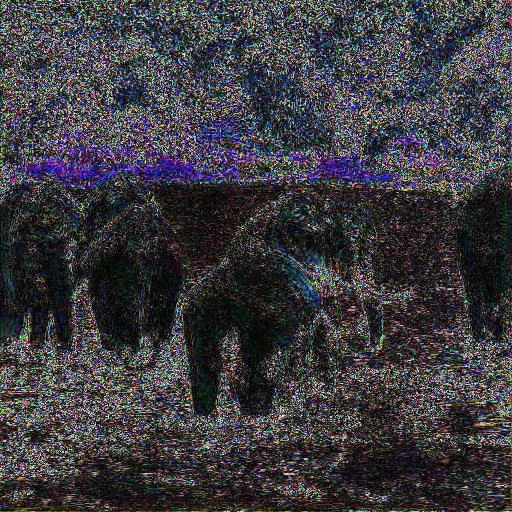} &
       \includegraphics[width=\imwidth]{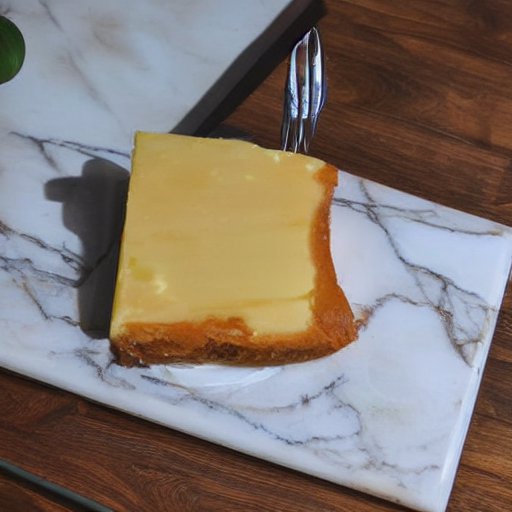} &
       \includegraphics[width=\imwidth]{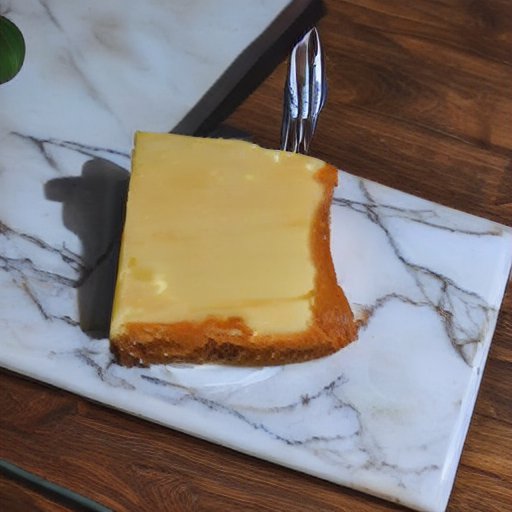} &
       \includegraphics[width=\imwidth]{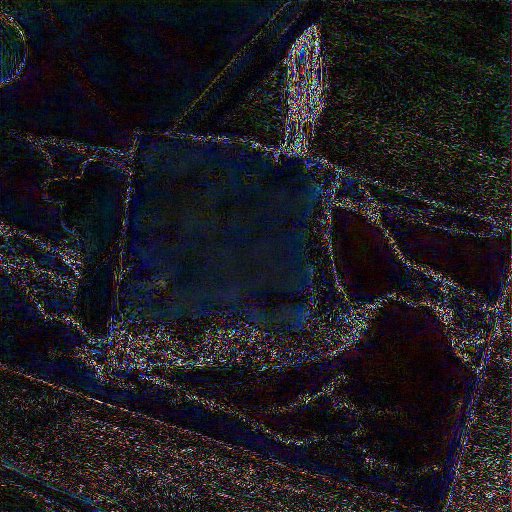} \\
       \rule{0pt}{6ex}%

        \includegraphics[width=\imwidth]{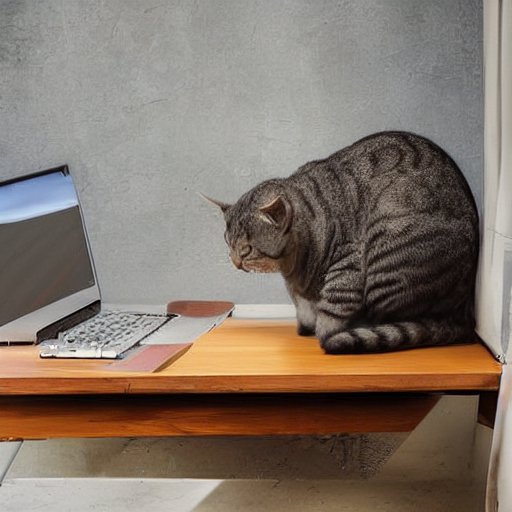} &
        \includegraphics[width=\imwidth]{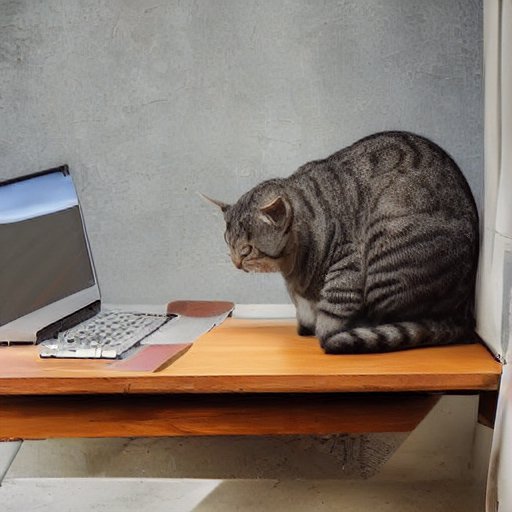} &
        \includegraphics[width=\imwidth]{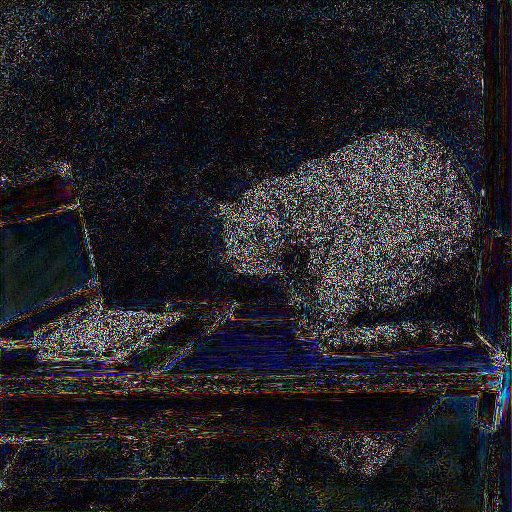} &
        \includegraphics[width=\imwidth]{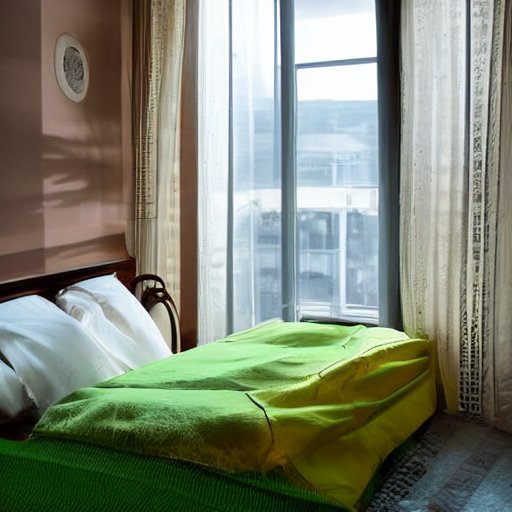} &
        \includegraphics[width=\imwidth]{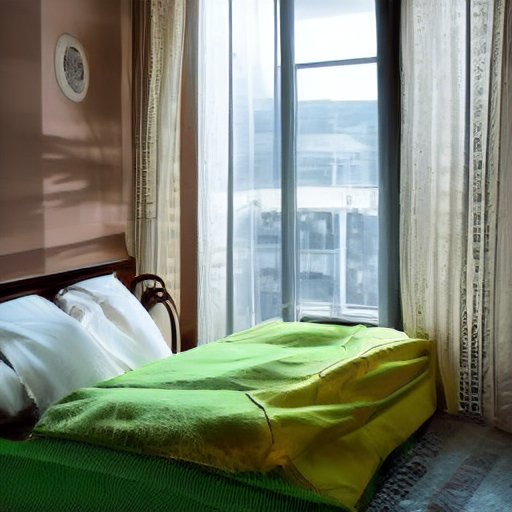} &
        \includegraphics[width=\imwidth]{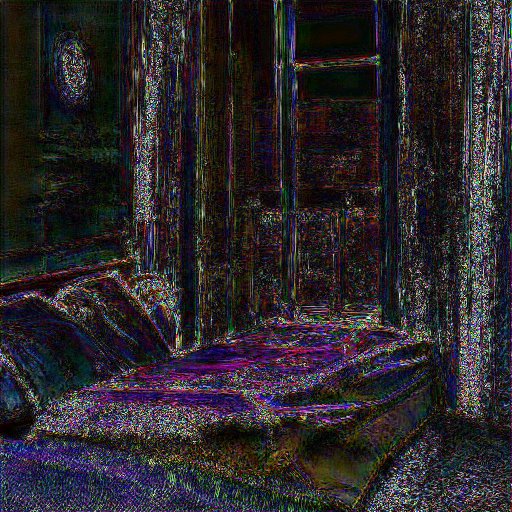} \\
        \rule{0pt}{6ex}%
        
        \includegraphics[width=\imwidth]{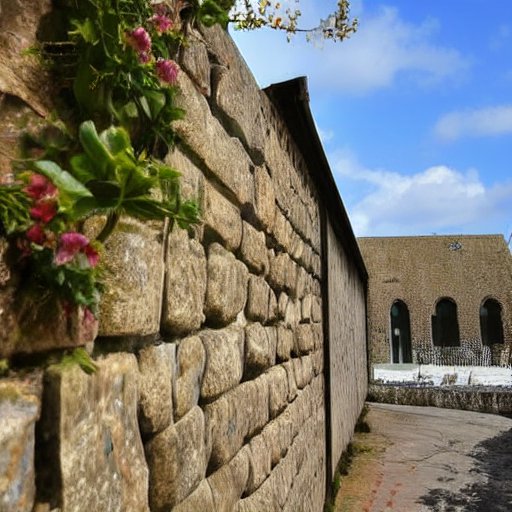} &
        \includegraphics[width=\imwidth]{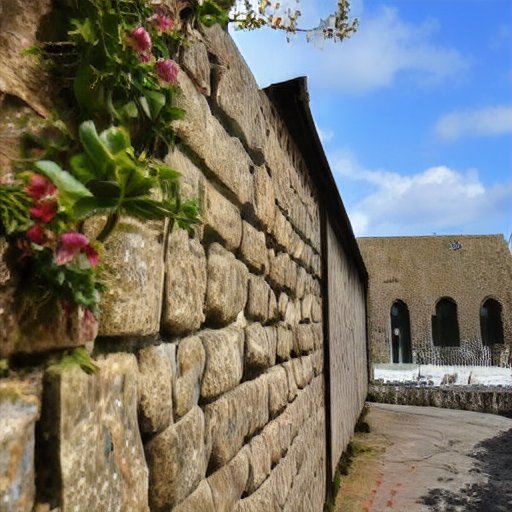} &
        \includegraphics[width=\imwidth]{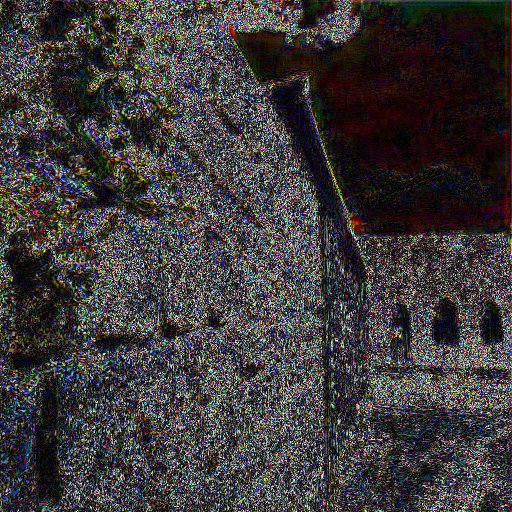} &
        \includegraphics[width=\imwidth]{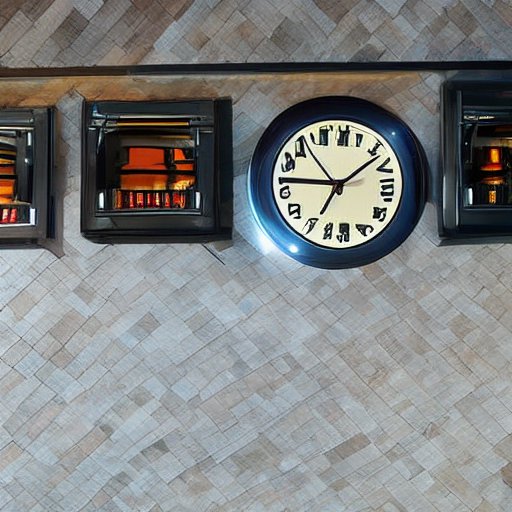} &
        \includegraphics[width=\imwidth]{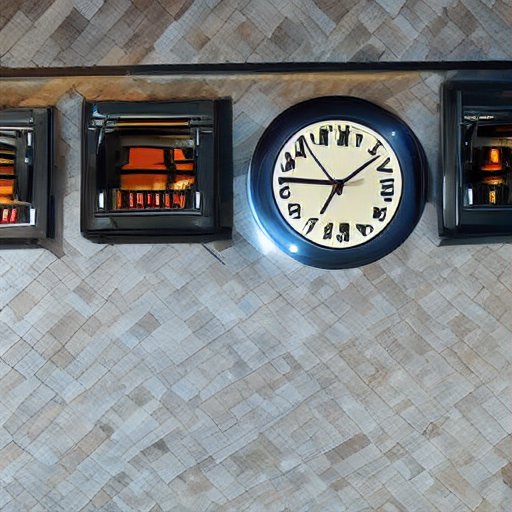} &
        \includegraphics[width=\imwidth]{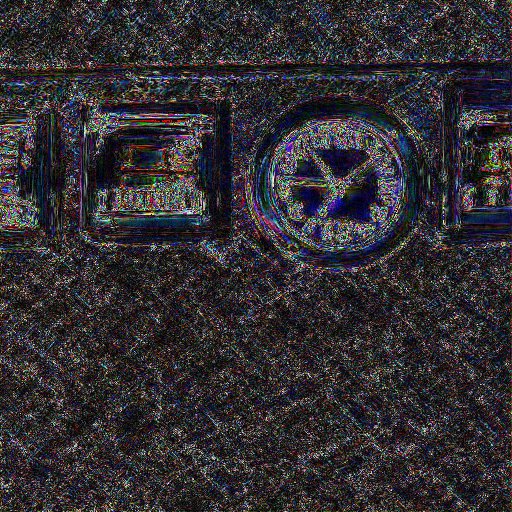} \\
        \rule{0pt}{6ex}%

        \includegraphics[width=\imwidth]{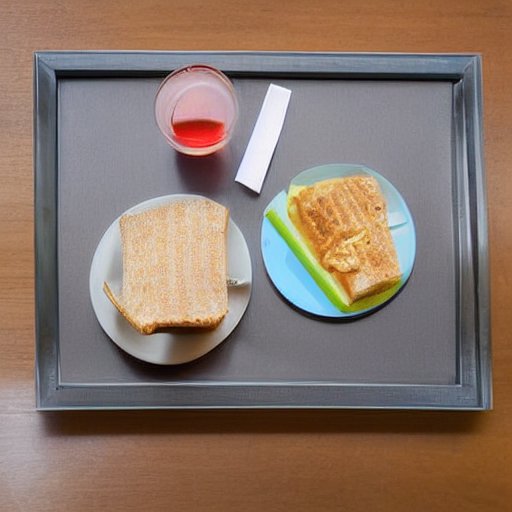} &
        \includegraphics[width=\imwidth]{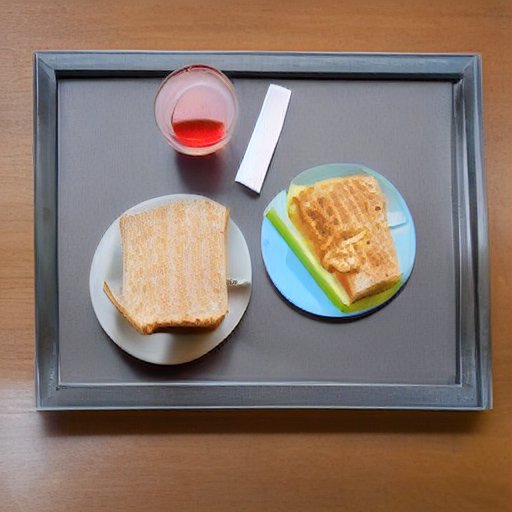} &
        \includegraphics[width=\imwidth]{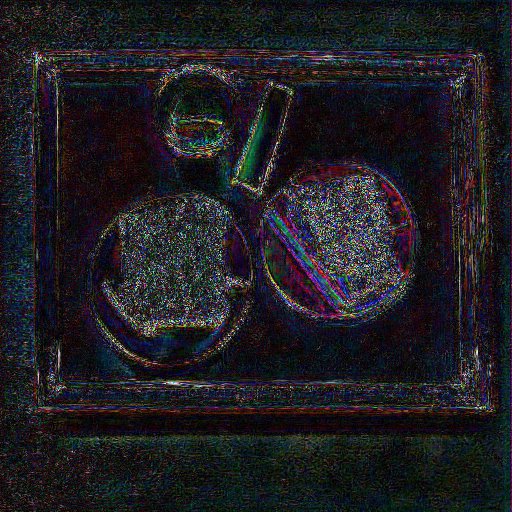} &
        \includegraphics[width=\imwidth]{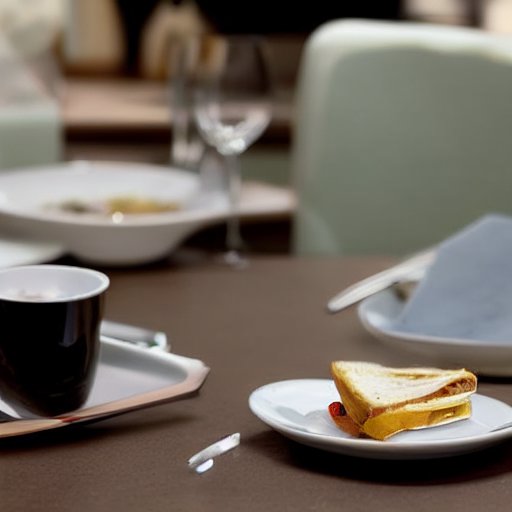} &
        \includegraphics[width=\imwidth]{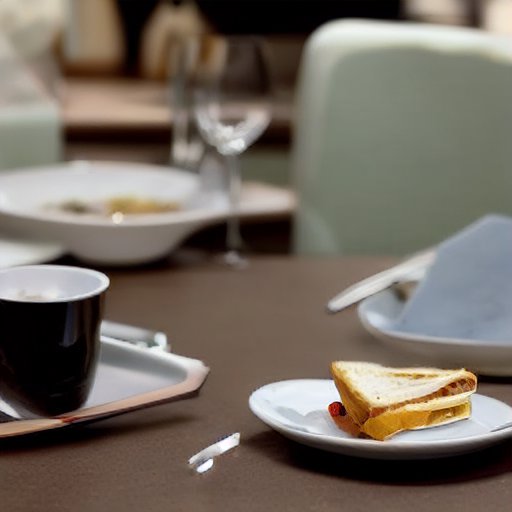} &
        \includegraphics[width=\imwidth]{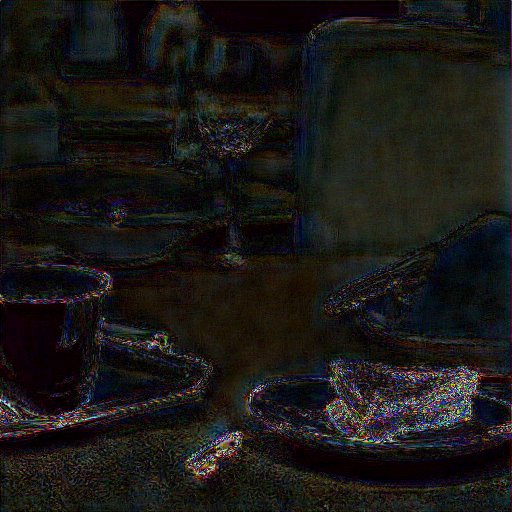} \\
    \bottomrule \\
    \end{tabular}
\caption{\label{fig:supp-txt2img} Qualitative results on prompts of the validation set of MS-COCO, at resolution $512$ and for a $48$-bits signature.
Images are generated from the same latents, with original or watermarked generative models.}
\end{figure*}

\begin{figure*}
    \centering
    \scriptsize
    \newcommand{\imwidth}{0.124\textwidth}
    \setlength{\tabcolsep}{0pt}
    \begin{tabular}{cc@{\hskip 2pt}ccc@{\hskip 2pt}ccc}
        \toprule
        Image to inpaint & Mask & Original & Watermarked & Difference & Original & Watermarked & Difference \\
        \midrule
        \hspace{0pt}
        \includegraphics[width=\imwidth]{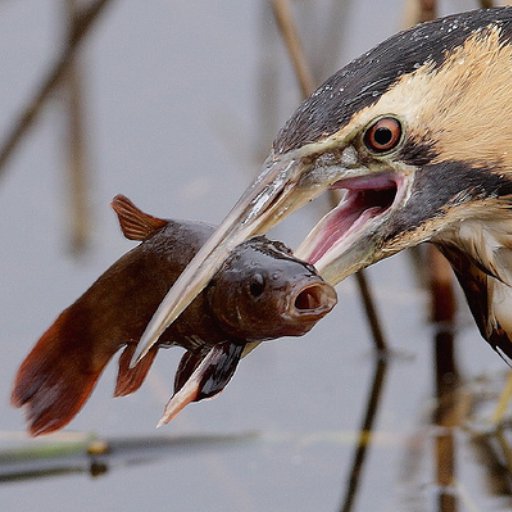} &
        \includegraphics[width=\imwidth]{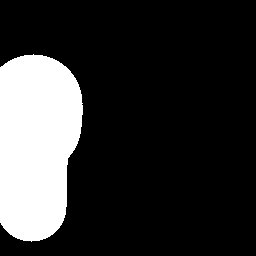} &
        \includegraphics[width=\imwidth]{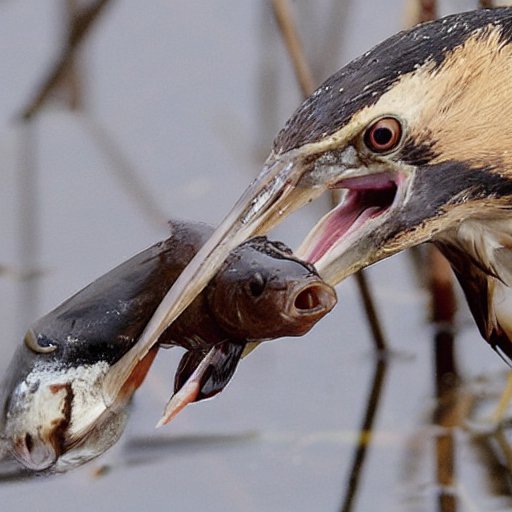} &
        \includegraphics[width=\imwidth]{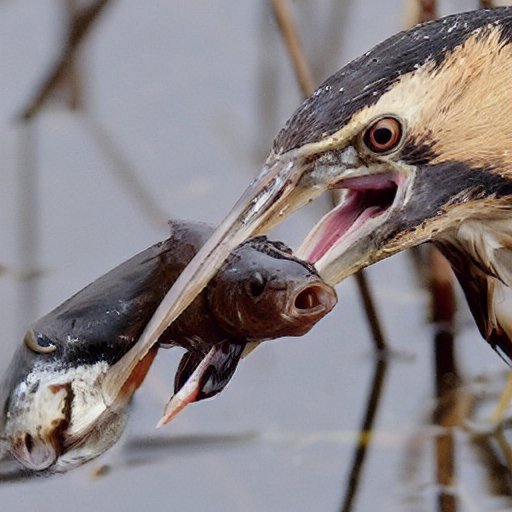} &
        \includegraphics[width=\imwidth]{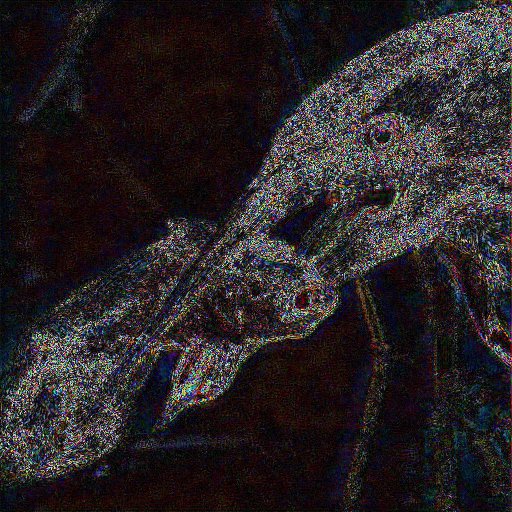} &
        \includegraphics[width=\imwidth]{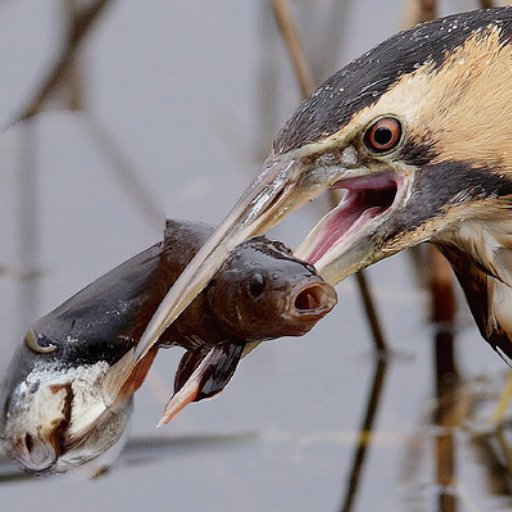} &
        \includegraphics[width=\imwidth]{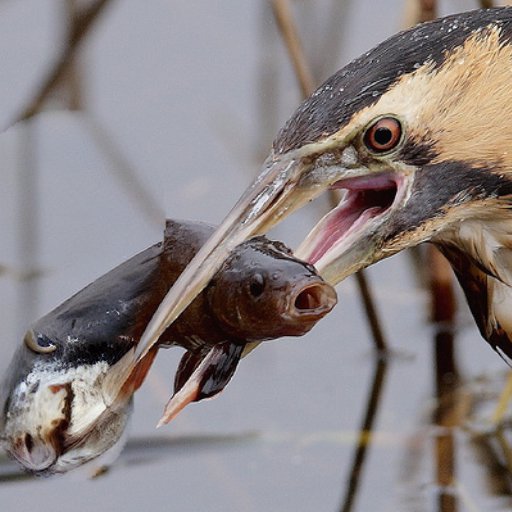} &
        \includegraphics[width=\imwidth]{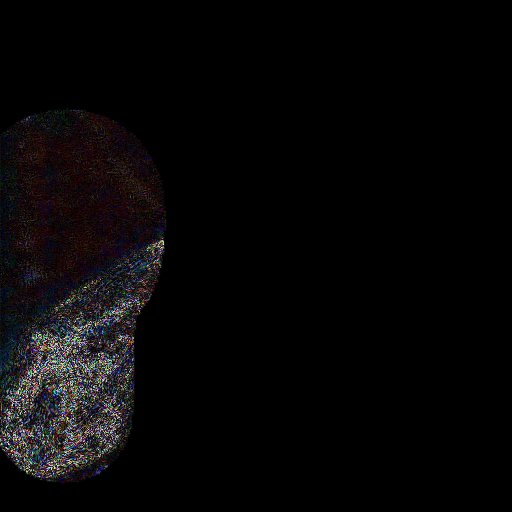} \\
        \rule{0pt}{6ex}

        \includegraphics[width=\imwidth]{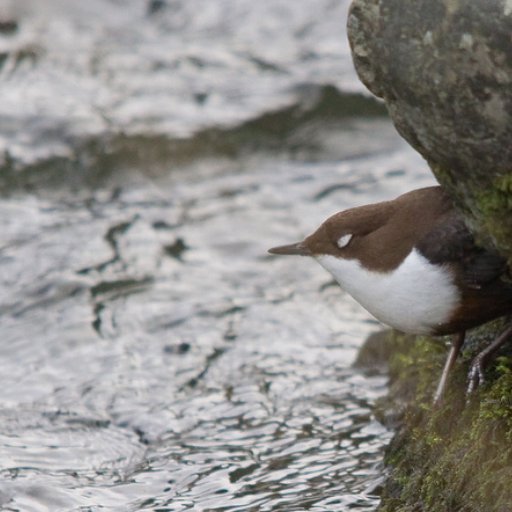} &
        \includegraphics[width=\imwidth]{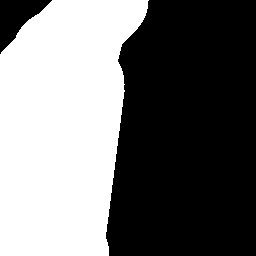} &
        \includegraphics[width=\imwidth]{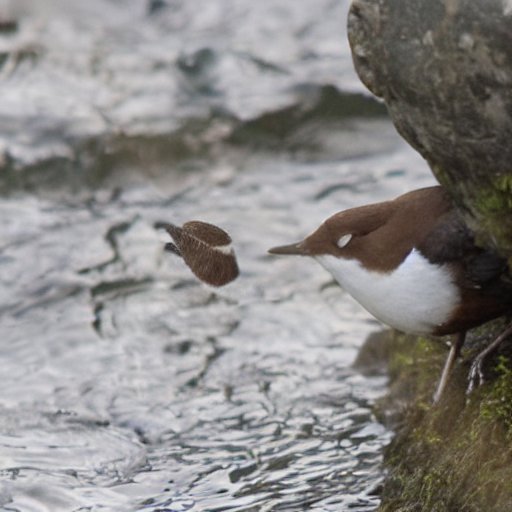} &
        \includegraphics[width=\imwidth]{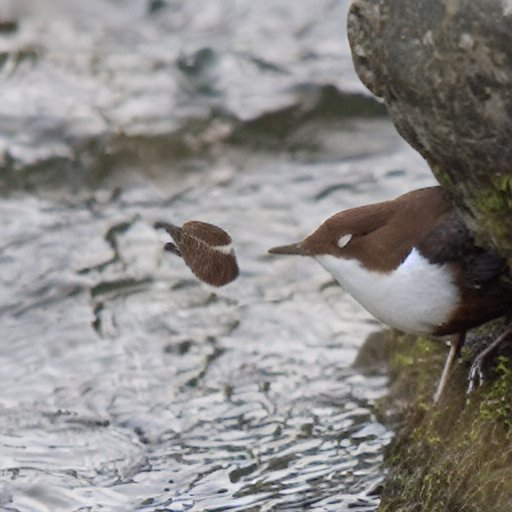} &
        \includegraphics[width=\imwidth]{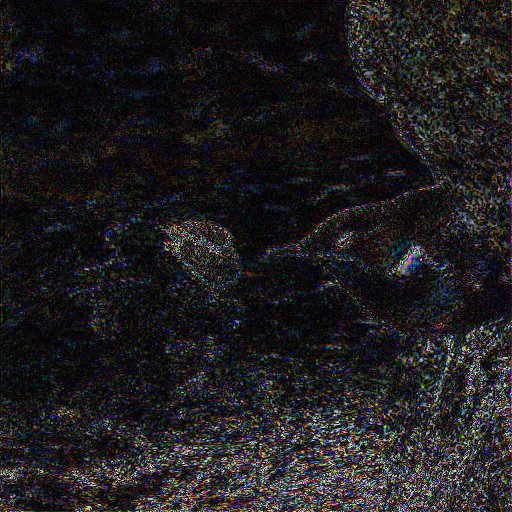} &
        \includegraphics[width=\imwidth]{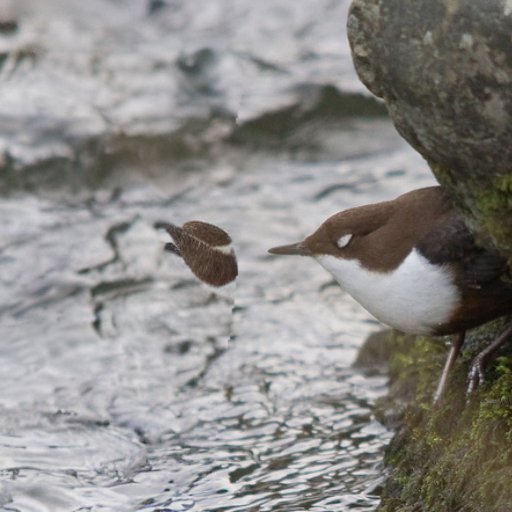} &
        \includegraphics[width=\imwidth]{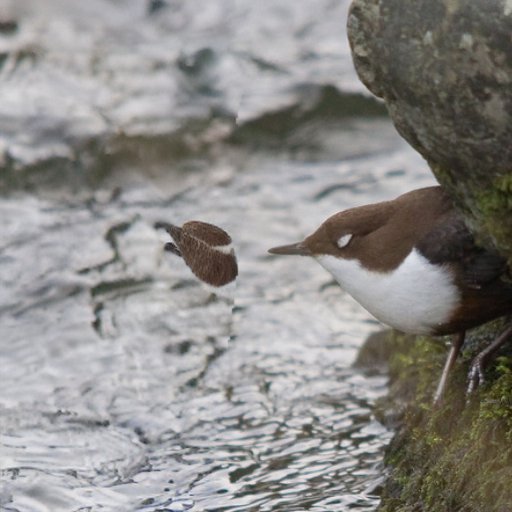} &
        \includegraphics[width=\imwidth]{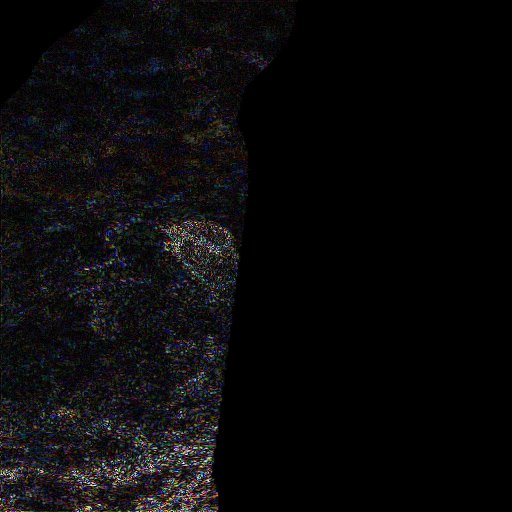} \\
        \rule{0pt}{6ex}

        \includegraphics[width=\imwidth]{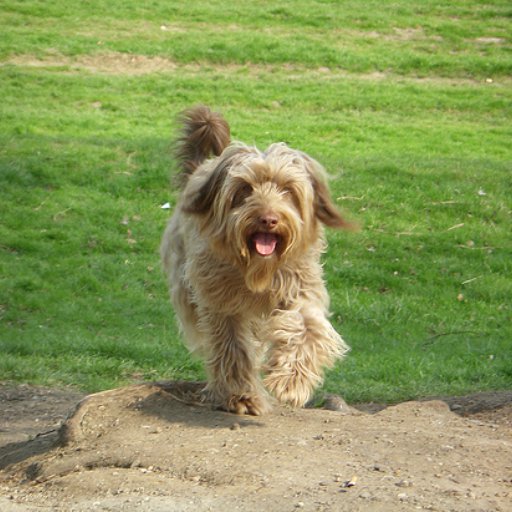} &
        \includegraphics[width=\imwidth]{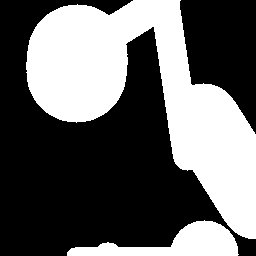} &
        \includegraphics[width=\imwidth]{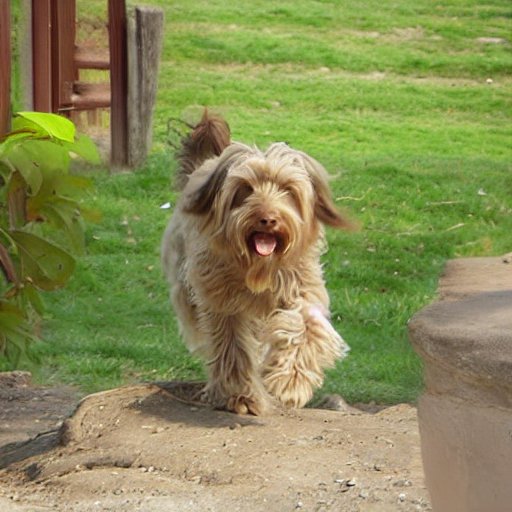} &
        \includegraphics[width=\imwidth]{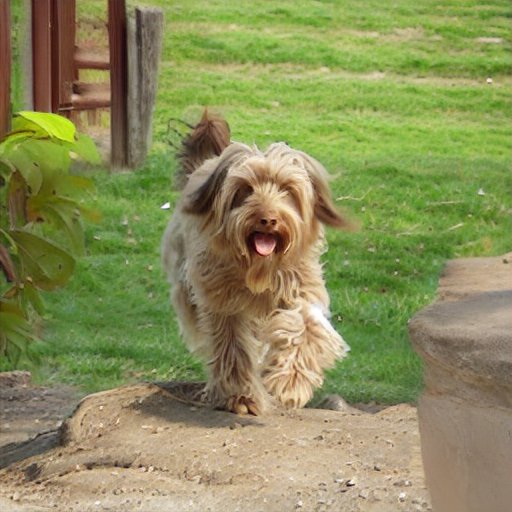} &
        \includegraphics[width=\imwidth]{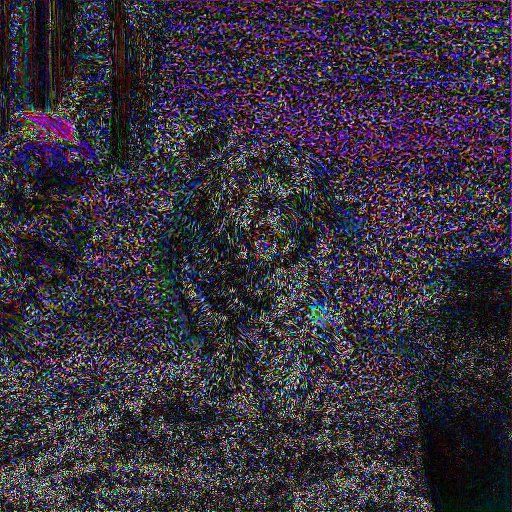} &
        \includegraphics[width=\imwidth]{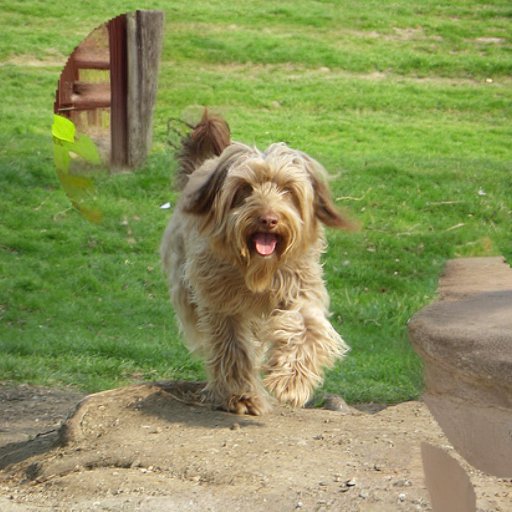} &
        \includegraphics[width=\imwidth]{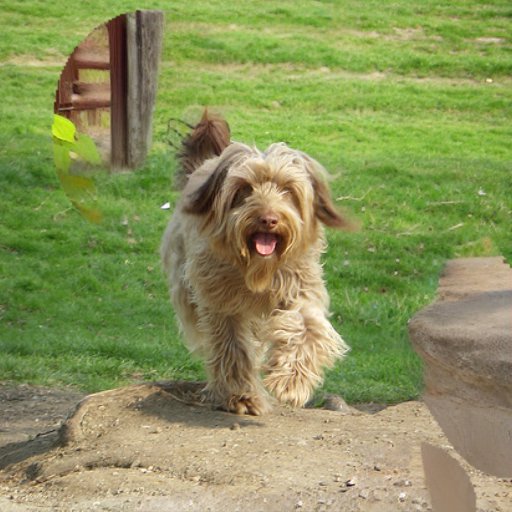} &
        \includegraphics[width=\imwidth]{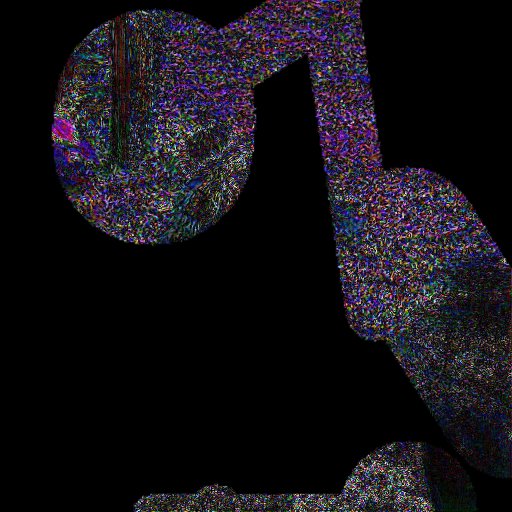} \\
        \rule{0pt}{6ex}

        \includegraphics[width=\imwidth]{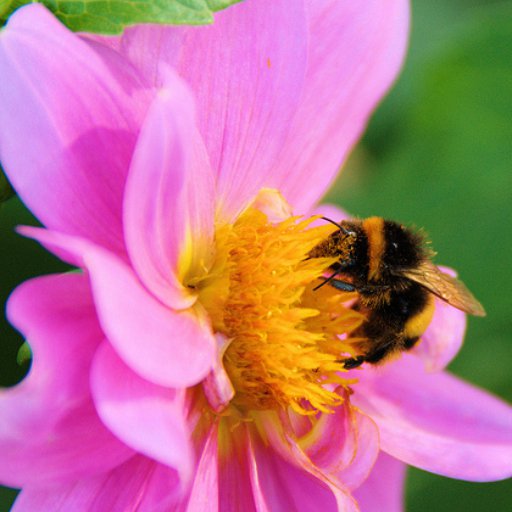} &
        \includegraphics[width=\imwidth]{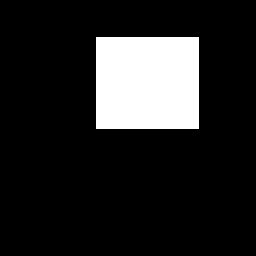} &
        \includegraphics[width=\imwidth]{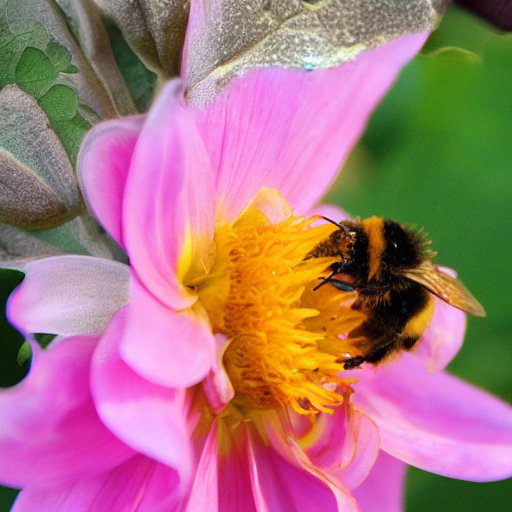} &
        \includegraphics[width=\imwidth]{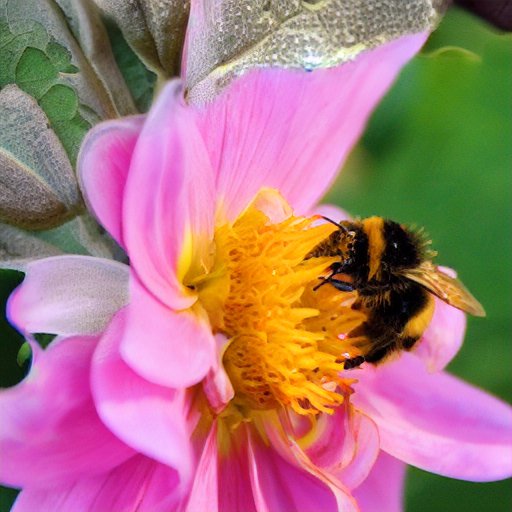} &
        \includegraphics[width=\imwidth]{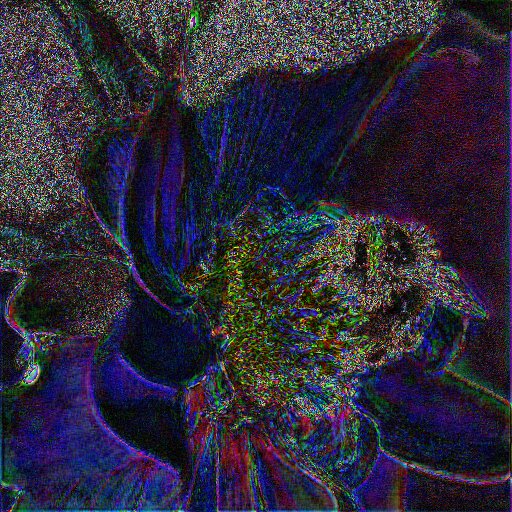} &
        \includegraphics[width=\imwidth]{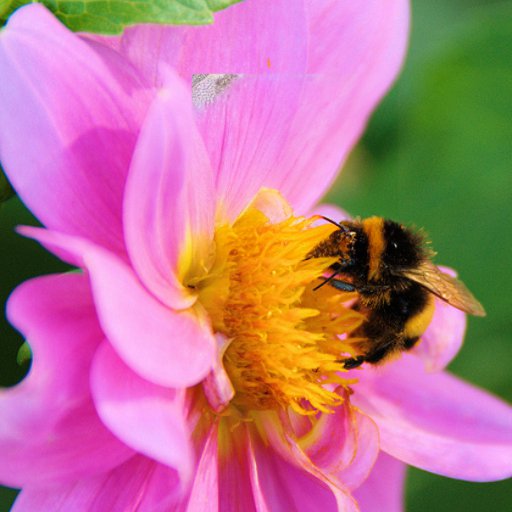} &
        \includegraphics[width=\imwidth]{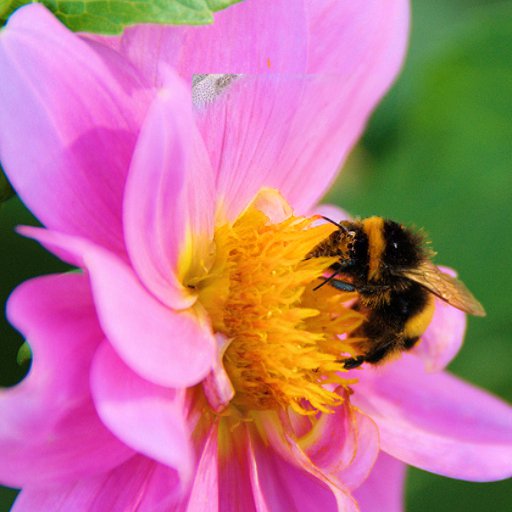} &
        \includegraphics[width=\imwidth]{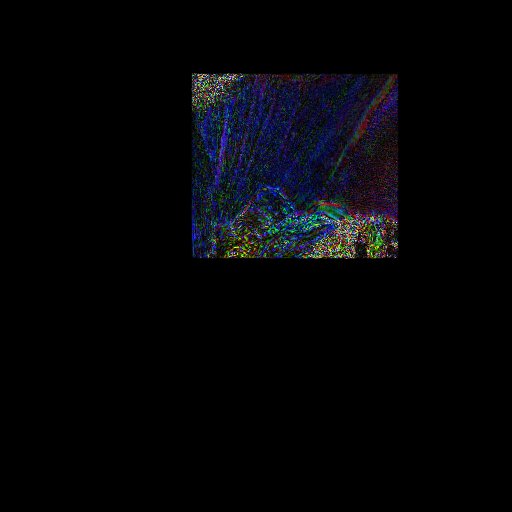} \\
        \rule{0pt}{6ex}

        \includegraphics[width=\imwidth]{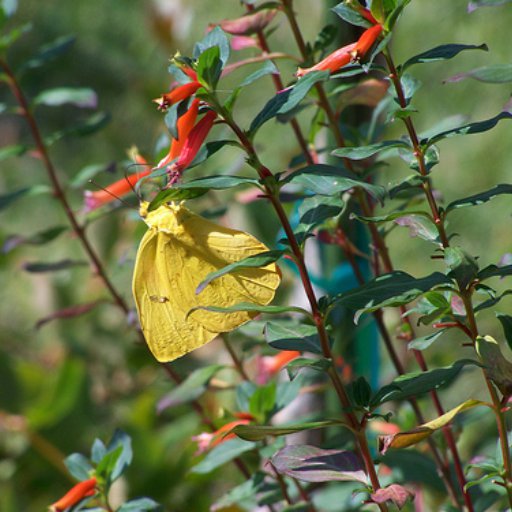} &
        \includegraphics[width=\imwidth]{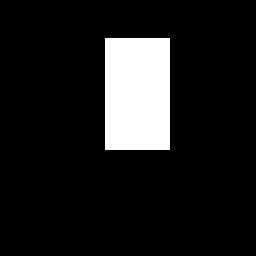} &
        \includegraphics[width=\imwidth]{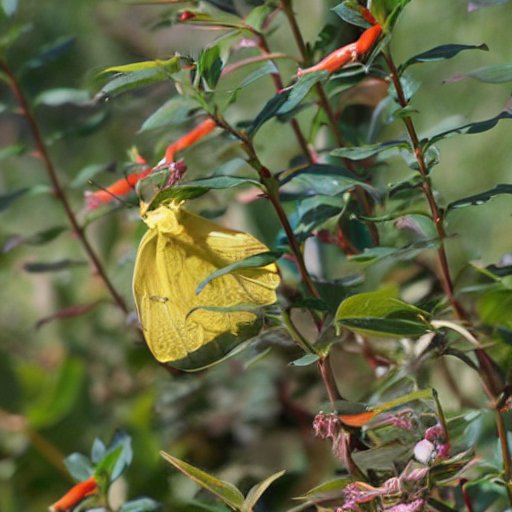} &
        \includegraphics[width=\imwidth]{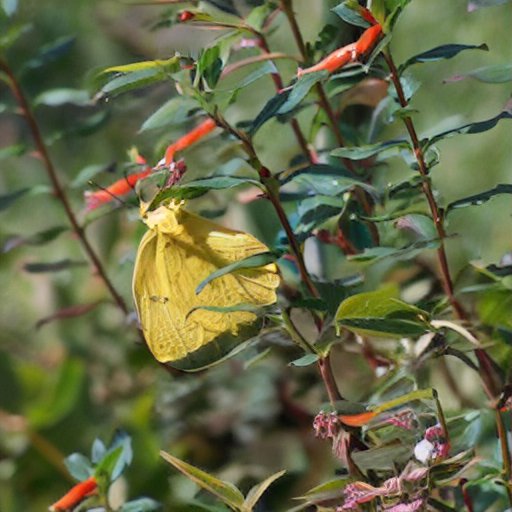} &
        \includegraphics[width=\imwidth]{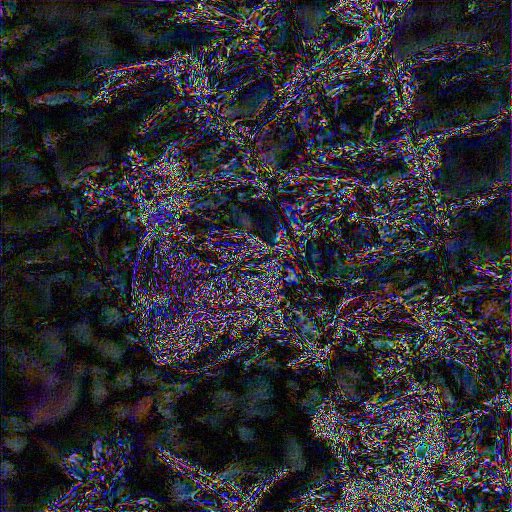} &
        \includegraphics[width=\imwidth]{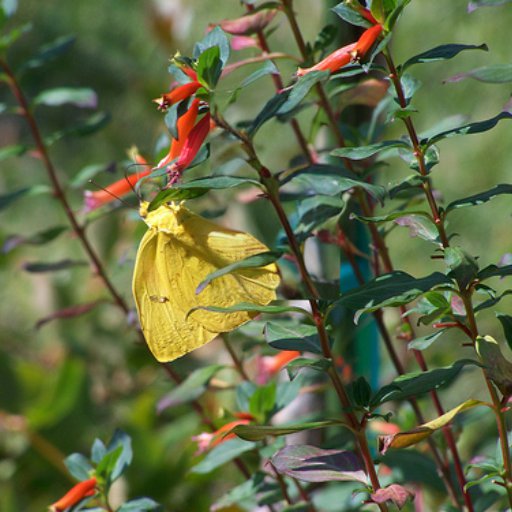} &
        \includegraphics[width=\imwidth]{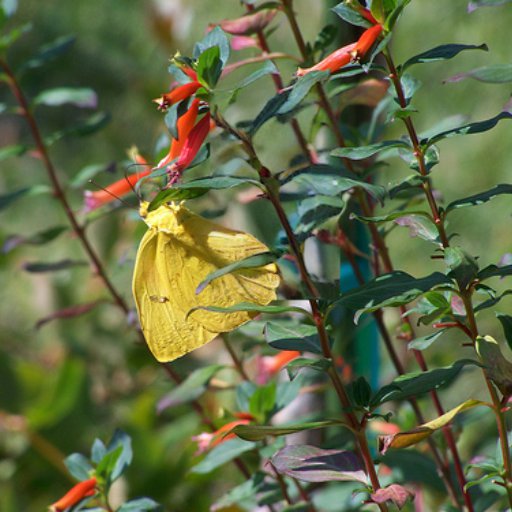} &
        \includegraphics[width=\imwidth]{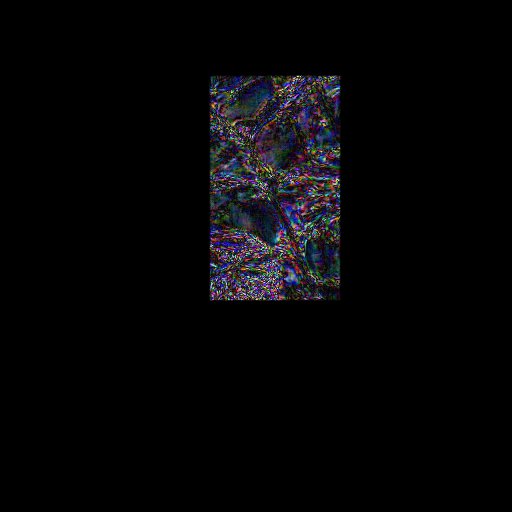} \\
        \rule{0pt}{6ex}

        \includegraphics[width=\imwidth]{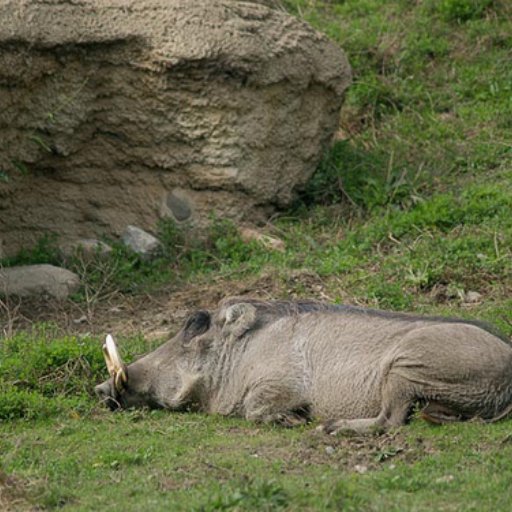} &
        \includegraphics[width=\imwidth]{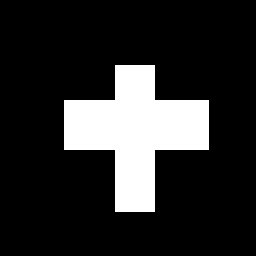} &
        \includegraphics[width=\imwidth]{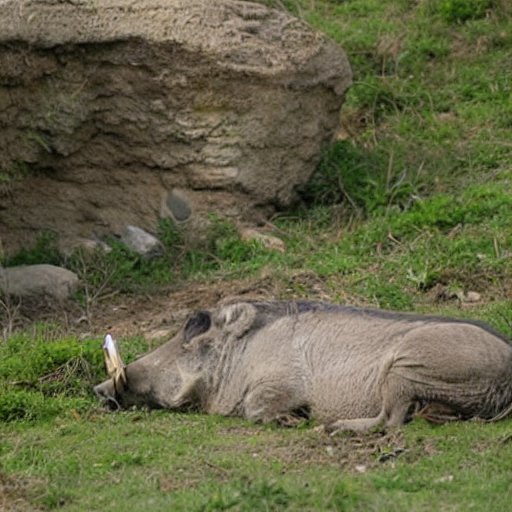} &
        \includegraphics[width=\imwidth]{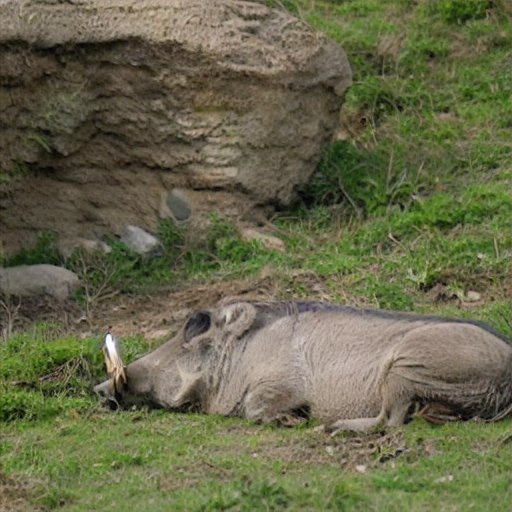} &
        \includegraphics[width=\imwidth]{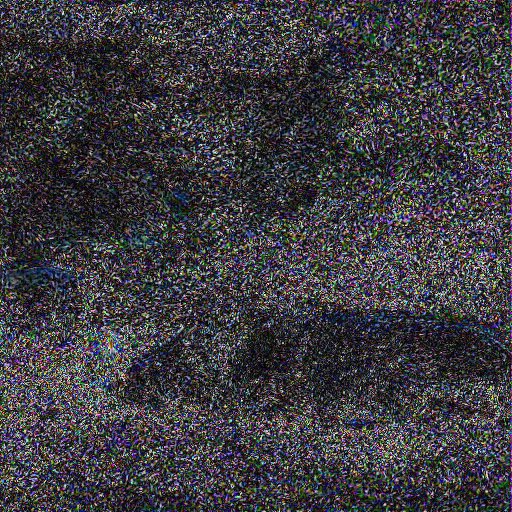} &
        \includegraphics[width=\imwidth]{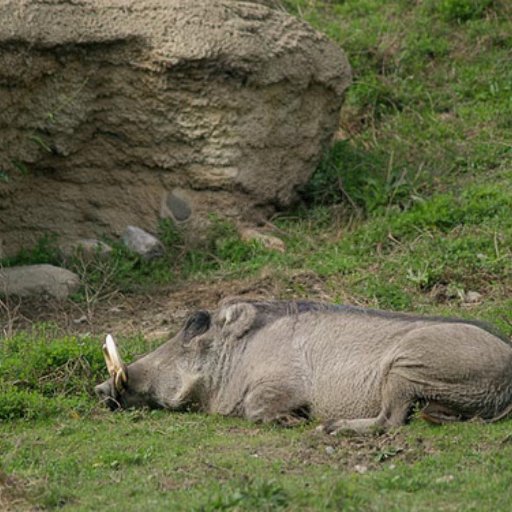} &
        \includegraphics[width=\imwidth]{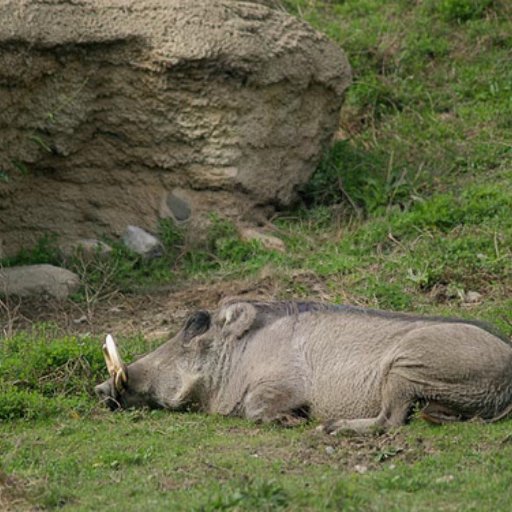} &
        \includegraphics[width=\imwidth]{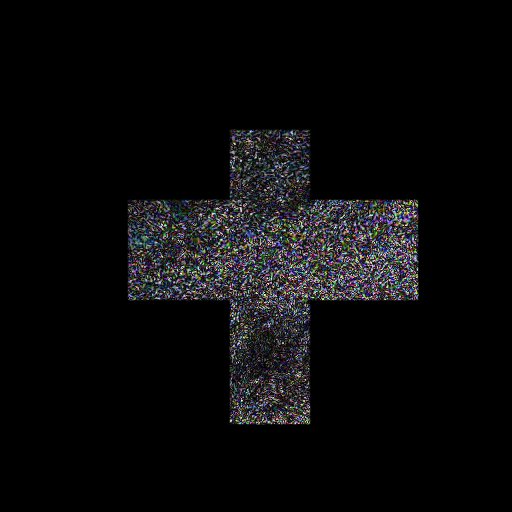} \\
        \bottomrule \\
    \end{tabular}
    \caption{
        \label{fig:supp-inpaint} Qualitative results for inpainting on ImageNet, with masks created from LaMa protocol~\cite{suvorov2022resolution}, with original or watermarked generative models.
        We consider 2 scenarios: 
        (middle) the full image is modified to fill the masked area, 
        (rigtht) only the masked area is filled.
        Since our model is not fine-tuned for inpainting, the last scenario introduces copy-paste artifacts.
        From a watermarking point of view, it is also the more interesting, since the watermark signal is only present in the masked area (and erased wherever the image to inpaint is copied).
        Even in this case, the watermark extractor achieves bit accuracy significantly higher than random.
    }
\end{figure*}

\begin{figure*}
    \centering
    \scriptsize
    \newcommand{\imwidth}{0.124\textwidth}
    \setlength{\tabcolsep}{0pt}
    \begin{tabular}{cccc@{\hskip 2pt}cccc}
        \toprule
        Low resolution & Original & Watermarked & Difference & Low resolution & Original & Watermarked & Difference \\
        \midrule
        \hspace{0pt}
        \includegraphics[width=\imwidth]{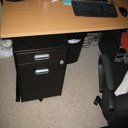} &
        \includegraphics[width=\imwidth]{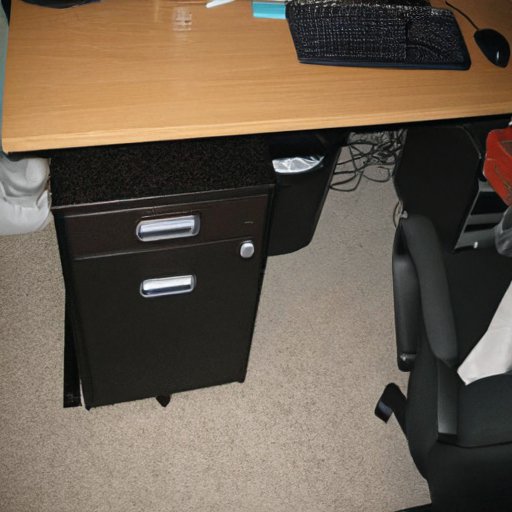} &
        \includegraphics[width=\imwidth]{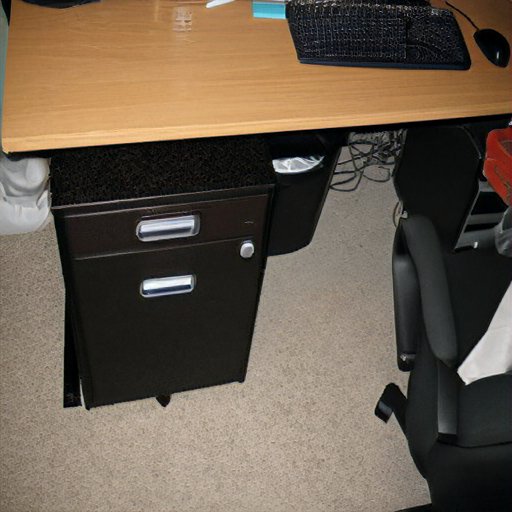} &
        \includegraphics[width=\imwidth]{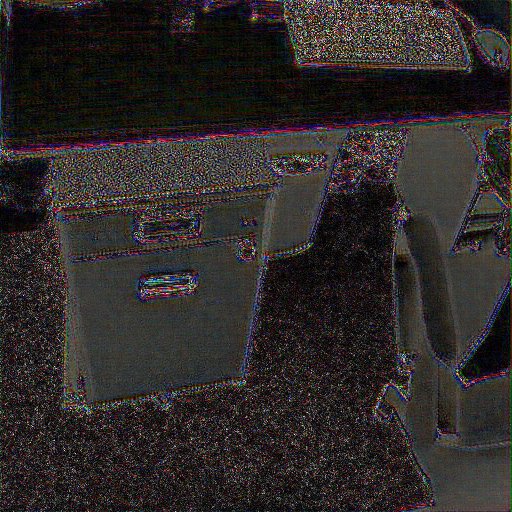} &
        \includegraphics[width=\imwidth]{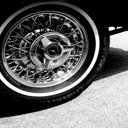} &
        \includegraphics[width=\imwidth]{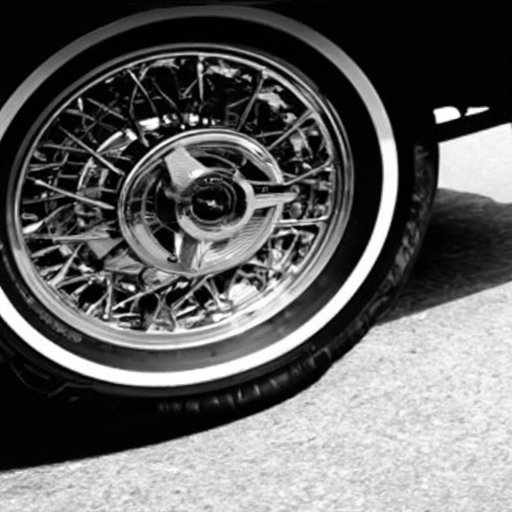} &
        \includegraphics[width=\imwidth]{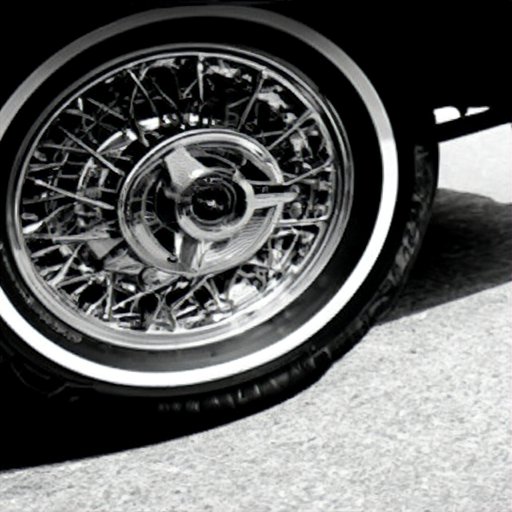} &
        \includegraphics[width=\imwidth]{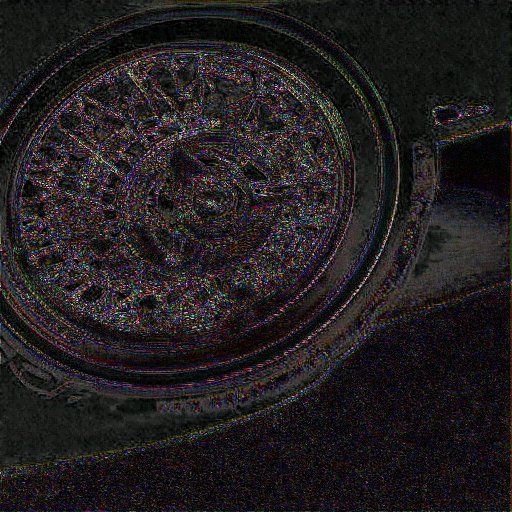} \\
        \rule{0pt}{6ex}

        \includegraphics[width=\imwidth]{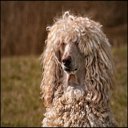} &
        \includegraphics[width=\imwidth]{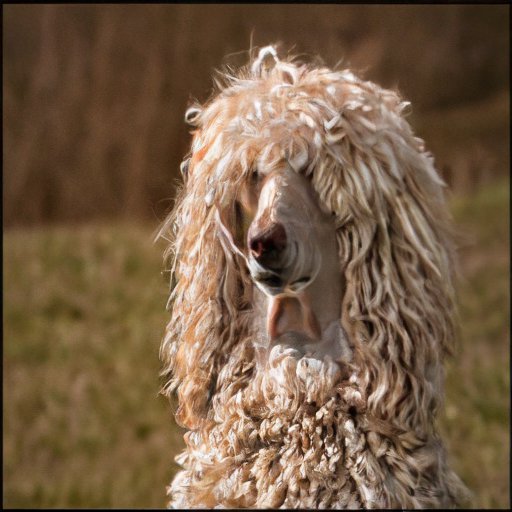} &
        \includegraphics[width=\imwidth]{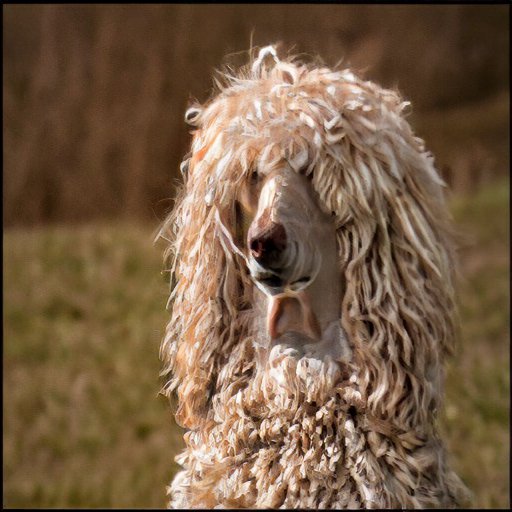} &
        \includegraphics[width=\imwidth]{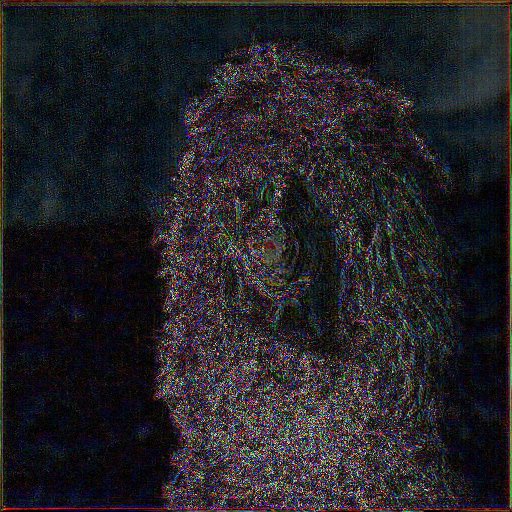} &
        \includegraphics[width=\imwidth]{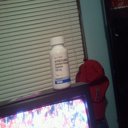} &
        \includegraphics[width=\imwidth]{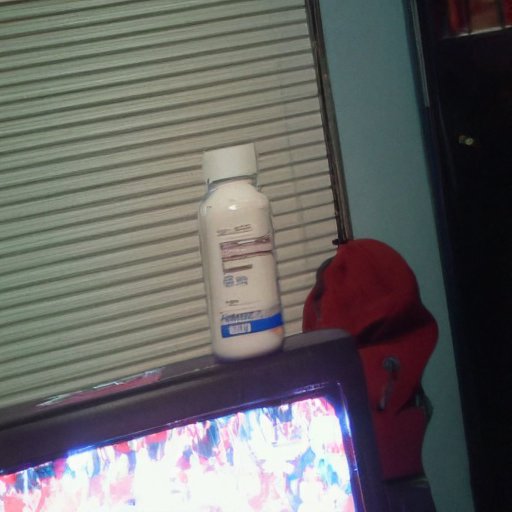} &
        \includegraphics[width=\imwidth]{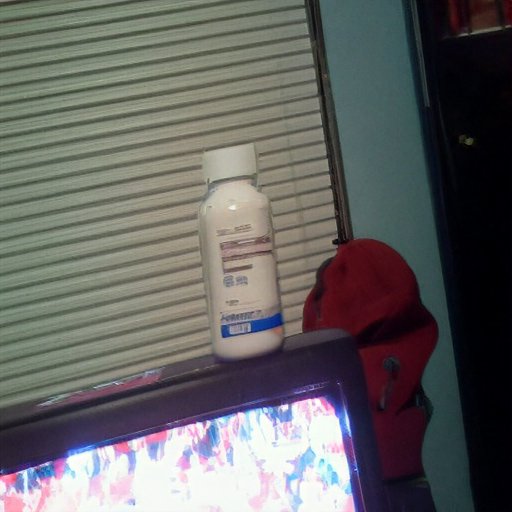} &
        \includegraphics[width=\imwidth]{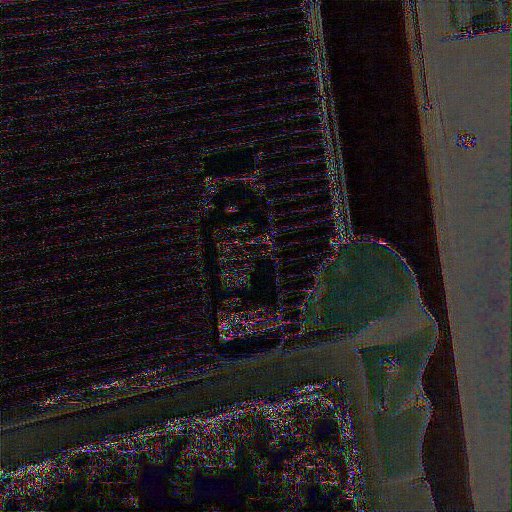} \\
        \rule{0pt}{6ex}

        \includegraphics[width=\imwidth]{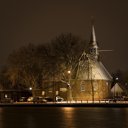} &
        \includegraphics[width=\imwidth]{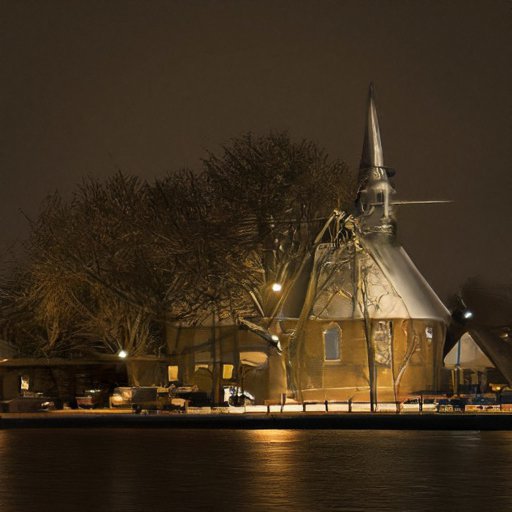} &
        \includegraphics[width=\imwidth]{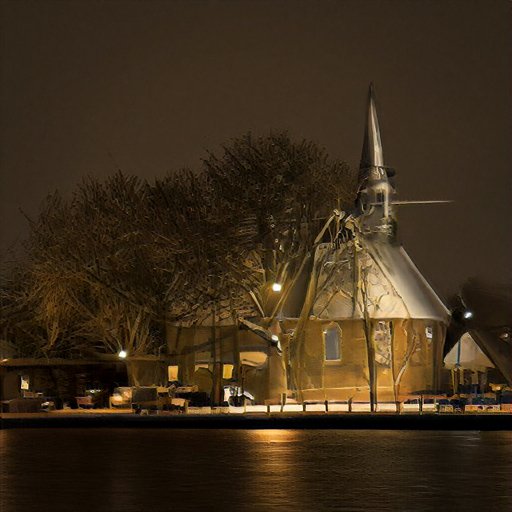} &
        \includegraphics[width=\imwidth]{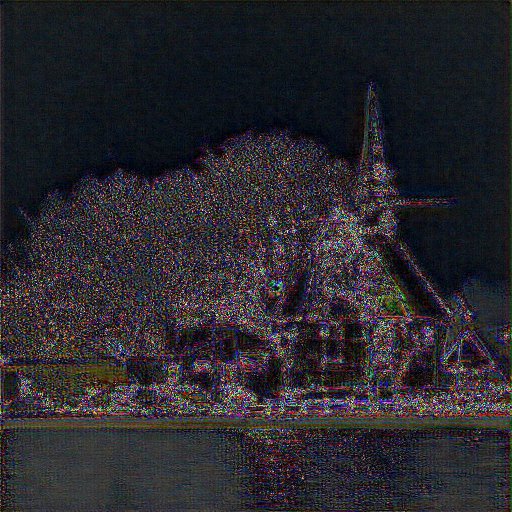} &
        \includegraphics[width=\imwidth]{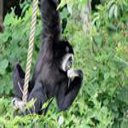} &
        \includegraphics[width=\imwidth]{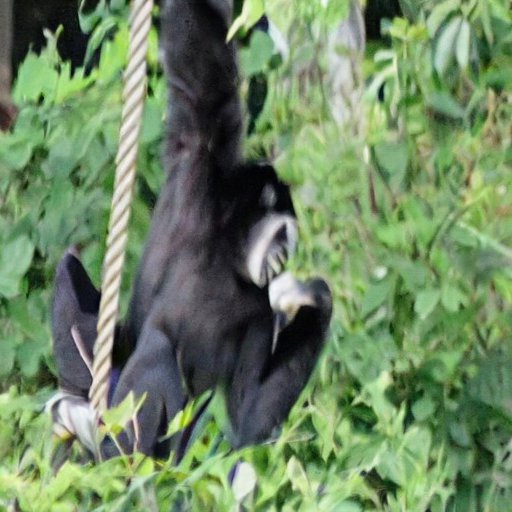} &
        \includegraphics[width=\imwidth]{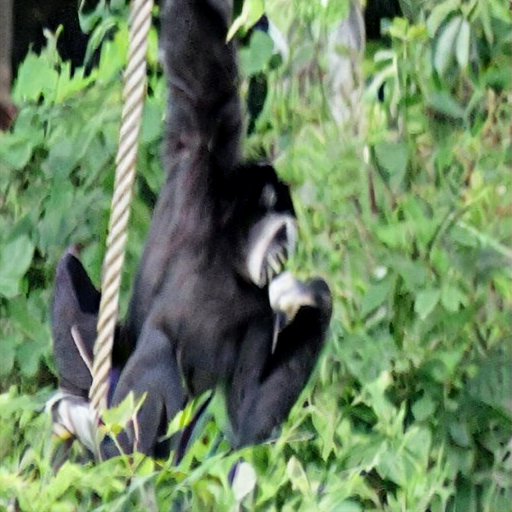} &
        \includegraphics[width=\imwidth]{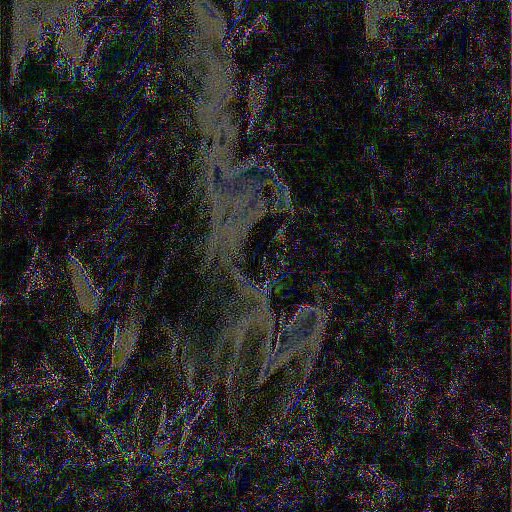} \\
        \rule{0pt}{6ex}

        \includegraphics[width=\imwidth]{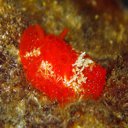} &
        \includegraphics[width=\imwidth]{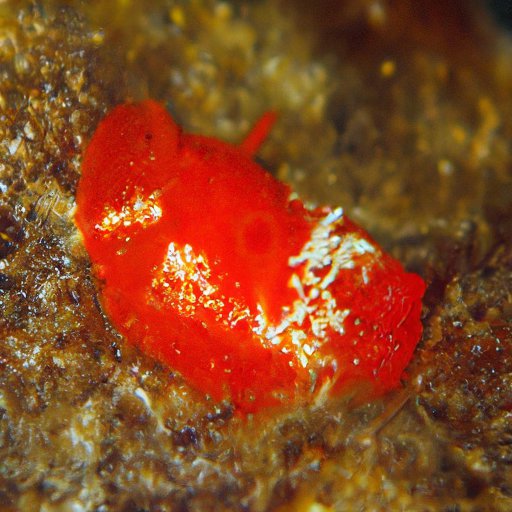} &
        \includegraphics[width=\imwidth]{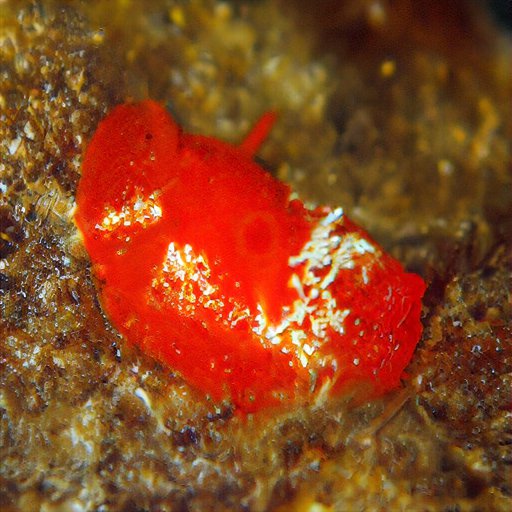} &
        \includegraphics[width=\imwidth]{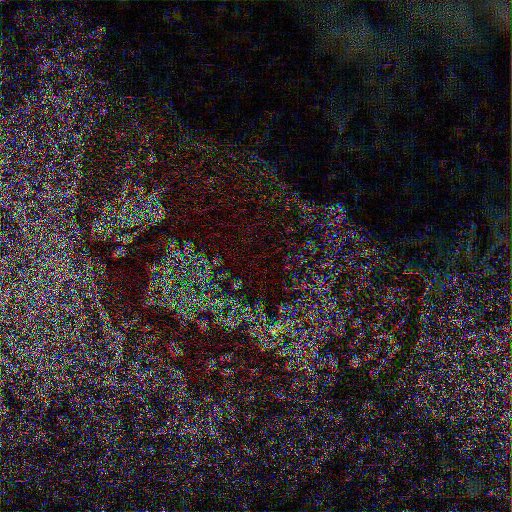} &
        \includegraphics[width=\imwidth]{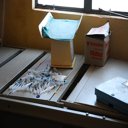} &
        \includegraphics[width=\imwidth]{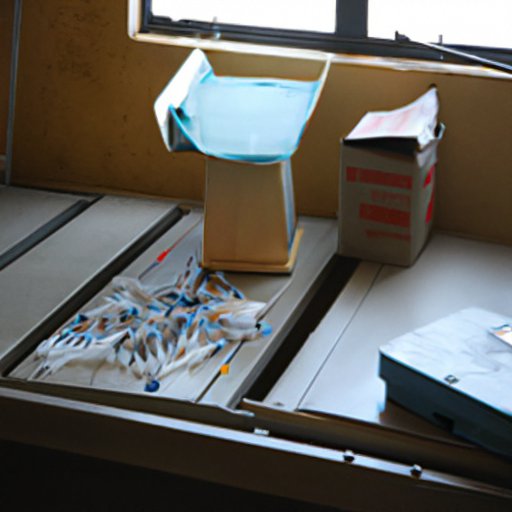} &
        \includegraphics[width=\imwidth]{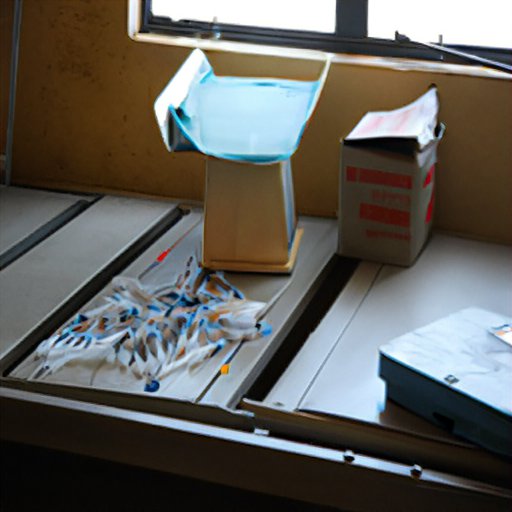} &
        \includegraphics[width=\imwidth]{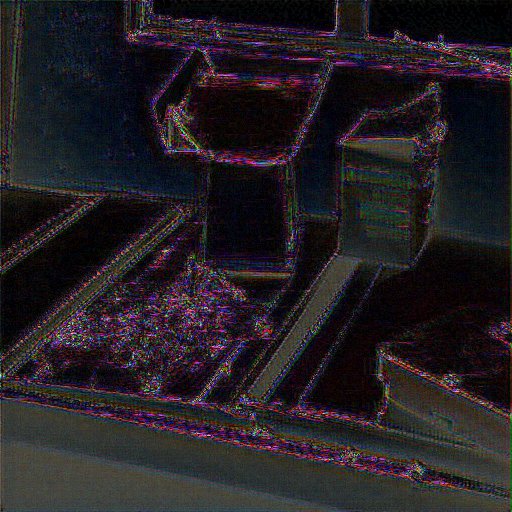} \\
        \bottomrule \\
    \end{tabular}
    \caption{
        \label{fig:supp-sr} Qualitative results for super-resolution on ImageNet, with original and watermarked generative models.
        Low resolution images are $128\times128$, and upscaled to $512\times512$ with an upscaling factor $f=4$.
    }
\end{figure*}

\end{document}